%% file: main.tex
\documentclass[10pt,twocolumn,letterpaper]{article}
\usepackage[OT1]{fontenc}
\usepackage[pagenumbers]{cvpr}
\input{preamble}
\input{supplementary_preamble}
\definecolor{cvprblue}{rgb}{0.21,0.49,0.74}
\usepackage[pagebackref,breaklinks,colorlinks,allcolors=cvprblue]{hyperref}
\hypersetup{
  pdftitle={GameRecreationBench: Can Multimodal Coding Agents Recreate Games through Interactive Play?},
  pdfauthor={Lidong Lu, Lidong Zou, Tong Lu, Guo Chen},
  pdfsubject={Interactive game recreation by multimodal coding agents},
  pdfkeywords={GameRecreationBench, coding agents, multimodal agents, game recreation, benchmark}
}
\title{\benchmarkname{}: Can Multimodal Coding Agents Recreate Games through Interactive Play?}
\input{authors}

\date{}

\makeatletter
\g@addto@macro\@maketitle{\input{figures/benchmark_teaser}}
\makeatother

\begin{document}
\maketitle

\input{sec/0_abstract}
\input{sec/1_intro}
\input{sec/2_related}
\input{sec/3_benchmark}
\input{sec/4_experiments}
\input{sec/6_conclusion}

\section*{LLM Usage Statement}
We used Codex to assist with research workflows, code development, and manuscript preparation. All research decisions, experimental results, and submitted content were reviewed and verified by the authors, who take full responsibility for the work.
{
    \small
    \bibliographystyle{ieeenat_fullname}
    \bibliography{main}
}
\input{supplementary}

\end{document}

%% file: preamble.tex
\usepackage{colortbl}
\definecolor{p2rtable}{HTML}{F1EDF7}
\definecolor{p2rpurple}{HTML}{7950B5}
\definecolor{p2rmuted}{HTML}{B4ADBE}
\definecolor{p2rbest}{HTML}{DCCDF0}
\definecolor{p2rsecond}{HTML}{F1EBF8}
\usepackage{pifont}
\newcommand{\cmark}{\textcolor{p2rpurple}{\ding{51}}}
\newcommand{\xmark}{\textcolor{p2rmuted}{\ding{55}}}

\newcommand{\benchmarkname}{GameRecreationBench}

\usepackage{placeins}
\newcommand{\plainimg}[2]{{\setlength{\fboxsep}{0pt}\setlength{\fboxrule}{1.4pt}\fcolorbox{white}{white}{\includegraphics[width=\dimexpr#1-2.8pt\relax]{#2}}}}
\newcommand{\hlbest}[1]{{\setlength{\fboxsep}{1.2pt}\colorbox{p2rbest}{\strut #1}}}
\newcommand{\hlsecond}[1]{{\setlength{\fboxsep}{1.2pt}\colorbox{p2rsecond}{\strut #1}}}

%% file: supplementary_preamble.tex
\usepackage{mdframed}
\newmdenv[topline=true,bottomline=true,leftline=false,rightline=false,linewidth=0.8pt,linecolor=p2rpurple,
  backgroundcolor=p2rtable,innerleftmargin=6pt,innerrightmargin=6pt,innertopmargin=5pt,innerbottommargin=5pt,
  skipabove=8pt,skipbelow=8pt]{promptframe}
\newenvironment{promptbox}[1]{\begin{figure*}[t]\begin{promptframe}\fontsize{7.6}{9}\selectfont\noindent{\color{p2rpurple}\textbf{#1}}\par\smallskip\setlength{\parskip}{2.5pt}}{\end{promptframe}\end{figure*}}

\definecolor{sfull}{HTML}{3E8E5E}\definecolor{spart}{HTML}{E0A030}\definecolor{szero}{HTML}{C8473A}
\newcommand{\scoreimg}[3]{{\setlength{\fboxsep}{0pt}\setlength{\fboxrule}{1.4pt}\fcolorbox{#1}{white}{\includegraphics[width=\dimexpr#2-2.8pt\relax]{#3}}}}
\newcommand{\noshot}[3]{{\setlength{\fboxsep}{0pt}\setlength{\fboxrule}{1.4pt}\fcolorbox{#1}{black!6}{\parbox[b][\dimexpr#3-2.8pt\relax][c]{\dimexpr#2-2.8pt\relax}{\centering\color{black!50}no footage}}}}
\newcommand{\scorelegend}{{\setlength{\fboxsep}{0pt}\setlength{\fboxrule}{1.4pt}%
  \fcolorbox{sfull}{white}{\rule{0pt}{5pt}\hspace{9pt}}~full points\quad
  \fcolorbox{spart}{white}{\rule{0pt}{5pt}\hspace{9pt}}~partial points\quad
  \fcolorbox{szero}{white}{\rule{0pt}{5pt}\hspace{9pt}}~zero points\quad
  \fcolorbox{szero}{black!6}{\rule{0pt}{5pt}\hspace{9pt}}~no footage of this moment}}

%% file: authors.tex
\author{
Lidong Lu\textsuperscript{1}\quad
Lidong Zou\textsuperscript{1}\quad
Tong Lu\textsuperscript{1}\thanks{Corresponding author.}\quad
Guo Chen\textsuperscript{2}\\[0.5em]
\textsuperscript{1}Nanjing University\qquad
\textsuperscript{2}NVIDIA\\[0.5em]
{\small \url{https://GameRecreationBench.github.io}}
}

%% file: figures/benchmark_teaser.tex
\noindent\begin{minipage}{\textwidth}
    \centering
    \includegraphics[width=\textwidth]{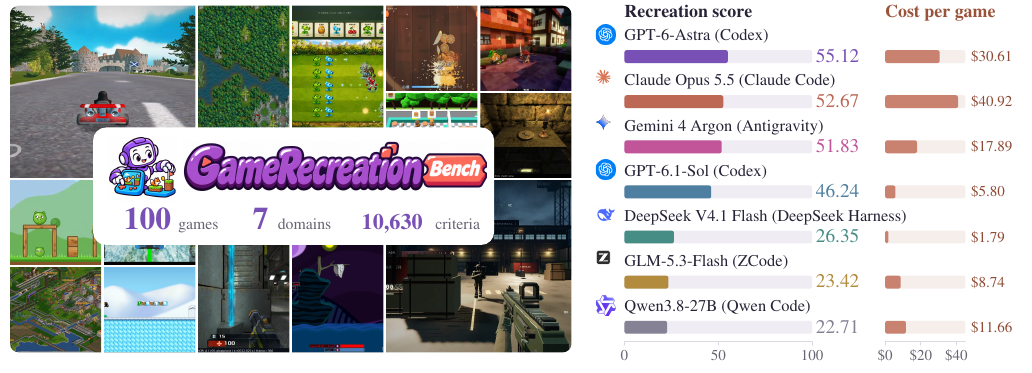}
    \captionof{figure}{\textbf{Overview of \benchmarkname{}.} Left, the benchmark spans 100 games across 7 domains, scored against 10,630 hidden criteria. Right, each system's mean recreation score and API-equivalent cost per game, where the best system reaches 55.12 of 100.}
    \label{fig:benchmark-teaser}
\end{minipage}
\par\vspace{8pt}

%% file: sec/0_abstract.tex
\begin{abstract}
Recreating a game through play requires a coding agent to discover how the game responds to actions and reproduce those behaviors in executable software.
Existing game benchmarks mostly evaluate either playing games or building them from written specifications, and recreation through play has so far been studied only on small sets of 2D games.
We introduce \benchmarkname{}, a benchmark of 100 open-source games, 48 of them in 3D, spanning diverse controls, multiplayer modes, and multi-level progression.
Agents play each game through screen frames and keyboard and mouse input and must deliver a standalone Godot recreation, without access to the source code or the evaluation rubric.
To score a recreation, we play the same scenarios in it and in the original game, and a video judge compares the two recordings to score mechanics, visual fidelity, game flow, interaction, and runtime stability against a rubric written for each game.
Across seven multimodal coding-agent systems, the best scores only 55.12 out of 100, and most lost points come from behavior that never appears in the recreation.
This missing behavior lies mainly beyond each game's opening. The best system earns 69.9\% of the points in the opening scenario but only 45.7\% in scenarios that require further play.
Separating playing from building shows that the leading agents rebuild about 90\% of the gameplay they reach in play, but their play covers at most a quarter of it.
Current agents thus rebuild what a game reveals early, while discovering its later levels, modes, and branches through play remains the central challenge.
\end{abstract}

%% file: sec/1_intro.tex
\section{Introduction}
\label{sec:intro}

A drag launches a projectile, a collision topples a structure, and clearing the stage opens the next level with new structures and goals.
These events give a player a concrete experience of how a game works: how movement responds to input, what a collision changes, and what it takes to advance from one level to the next.
For a developer, they also specify what a recreation must do.
An agent able to turn this experience into software could prototype gameplay from a playable example or rebuild an existing game in a new engine, returning to the original whenever its understanding or implementation falls short.
This prospect motivates our question: \emph{Can multimodal coding agents recreate what they play?}

Both halves of this ability have advanced separately.
Agents trained or prompted to play now operate games from pixels and keyboard or mouse input~\cite{mnih2015dqn,vinyals2019alphastar,baker2022vpt,tan2024cradle,sima2024scaling}, and coding agents resolve repository-level issues by editing, executing, and testing code~\cite{jimenez2024swebench,yang2024sweagent,wang2025openhands}.
Recent benchmarks ask agents to build complete games~\cite{luo2026gamecraft,chen2026swegame,zhang2025vgamegym}, but they specify the target through text requirements or source code, so the agent never has to find out how the game behaves.
Recreation through play joins the two.
Varying a launch gesture helps the agent learn how motion depends on input and how collisions change the scene. It must then express these observations as a program that reproduces the motion, state changes, and visual feedback.
Testing that program may expose a discrepancy that calls for another experiment in the original, and later stages introduce new goals that must be added without breaking earlier behavior.
The task thus links visual understanding, active exploration, and iterative software development through one evolving implementation, and it extends work on learning models of an environment through interaction~\cite{ha2018worldmodels,tsividis2021theory,tang2024worldcoder} to complete, executable games.

We introduce \benchmarkname{}, a benchmark of \textbf{100 existing open-source games}: 52 in 2D and 48 in 3D, spanning \textbf{seven gameplay domains} from puzzles and action to strategy, role-playing, and simulation.
The collection includes games played in independent rounds, games that unfold over multiple levels, and simulations whose worlds keep evolving.
As shown in Figure~\ref{fig:benchmark-protocol}, agents access each original only through a unified screen-and-action interface, with its source code and scoring criteria kept private.
They freely alternate between reference play, implementation, and testing to build an independently launchable Godot~\cite{godot} project that reproduces the original's appearance, gameplay, and game flow.

We judge a recreation the way a player experiences it, by whether it looks and behaves like the original as play unfolds.
The evidence is the recreation's own behavior during play, not its code or a description of its rules~\cite{qiao2026gamereplica}.
Every game has a structured rubric over five dimensions: mechanics, visual fidelity, game flow, interaction, and runtime stability.
Each criterion describes a property observable during play, such as the consequence of an action, a transition between stages, or the lighting of a scene, and reference recordings show the expected behavior and presentation.
To collect evidence, we freeze the submitted project and run a separate demonstration phase guided by game-specific scenarios that probe core mechanics, level transitions, and later-stage content.
A video-based judge applies the rubric to the recordings and assigns full, partial, or zero points to each criterion.
We also measure how much of each rubric the agent's own play covers and how much of what it played it rebuilds, which separates failures of playing from failures of building.

We evaluate seven multimodal coding-agent systems on all 100 games, and the best reaches a mean score of \textbf{55.12 out of 100}.
For every system, 64--75\% of the lost points come from behavior missing from the recreation.
These omissions grow beyond the opening: the strongest system earns 69.9\% of the points in each game's entry scenario but 45.7\% beyond it, and the games that are hard for every system, such as 3D, real-time, and multiplayer games, are those whose openings reveal the least.
Agents rebuild about 90\% of the gameplay they play, but their play covers at most a quarter of it, and the weaker systems fall behind mainly in building what they play.
Holding the model fixed, the coding harness alone moves the score by up to 5.3 points, mostly through content beyond the opening.
Current agents thus rebuild mainly what a game shows early, and exploring later levels, modes, and branches remains the central open challenge.

%% file: sec/2_related.tex
\section{Related Work}
\label{sec:related}

\input{tables/benchmark_comparison}

\paragraph{Coding agents.}
Code generation benchmarks have moved from self-contained functions and competition problems~\cite{chen2021codex,austin2021mbpp,hendrycks2021apps,li2022alphacode,jain2025livecodebench,zhuo2025bigcodebench} to repository-level issue resolution~\cite{jimenez2024swebench} and long-horizon projects such as library implementation, machine-learning engineering, and research replication~\cite{zhao2025commit0,chan2025mlebench,starace2025paperbench}.
Agents for these settings reason and act through tools, refine programs with execution feedback and self-reflection~\cite{yao2023react,shinn2023reflexion,chen2024selfdebug,wang2024codeact}, and are organized as single-agent scaffolds~\cite{yang2024sweagent,wang2025openhands,xia2024agentless} or role-based teams~\cite{qian2024chatdev,hong2024metagpt}.
Multimodal benchmarks add visual specifications, asking models to generate webpages from screenshots, sketches, or interaction traces~\cite{si2025design2code,laurencon2024websight,yun2024web2code,li2024sketch2code,xiao2024interaction2code}, reproduce plots and charts~\cite{wu2024plot2code,yang2025chartmimic}, solve visual programming problems~\cite{li2024mmcode}, fix user-facing issues~\cite{yang2025swebenchmultimodal}, and build interactive artifacts~\cite{zhang2025artifactsbench}.
In all these tasks the target is given up front as text, code, or images, whereas in \benchmarkname{} it must be obtained by playing, and a mismatch found in testing sends the agent back to the original.

\paragraph{Game agents and world models.}
Multimodal agents operate websites, desktops, and phones from screenshots~\cite{deng2023mind2web,zhou2024webarena,koh2024visualwebarena,xie2024osworld,rawles2025androidworld}, aided by visual grounding~\cite{yang2023setofmark,zheng2024seeact} and GUI-specialized models~\cite{hong2024cogagent,qin2025uitars}.
Games have long served as testbeds for such agents, from Atari and real-time strategy~\cite{bellemare2013ale,mnih2015dqn,vinyals2019alphastar,berner2019openaifive} to open-ended worlds~\cite{kuttler2020nethack,hafner2022crafter,fan2022minedojo,baker2022vpt,wang2023voyager,sima2024scaling}.
Recent benchmarks measure language and vision-language models on text-based and visual game suites~\cite{wu2024smartplay,liu2024agentbench,paglieri2024balrog,hu2025lmgame} and on commercial or browser video games~\cite{zhang2025videogamebench,park2025orak,ouyang2026gameworld}, and Cradle controls games through screen observations and keyboard--mouse actions~\cite{tan2024cradle}.
These benchmarks, the game-playing group in Table~\ref{tab:benchmark-comparison}, score progress within the game being played.
A second line learns how an environment works from interaction. Neural world models simulate games from observations and actions~\cite{ha2018worldmodels,kim2020gamegan,alonso2024diamond,bruce2024genie,valevski2025gamengen}, and Video2Game builds interactive environments from a single video~\cite{xia2024video2game}.
Program-based approaches infer theories or executable world models as code~\cite{lake2017building,tsividis2021theory,tang2024worldcoder,dainese2024codeworld}, abstraction benchmarks probe rule induction~\cite{chollet2019arc}, and exploration methods seek states that reveal new behavior~\cite{pathak2017curiosity,ecoffet2021goexplore}.
\benchmarkname{} joins the two lines at the scale of complete games: agents play to build, and the model they learn must be an engine project whose look and play match the original.

\input{figures/benchmark_protocol}

\paragraph{Game creation.}
Beyond procedural generation of levels and assets~\cite{summerville2018pcgml}, recent benchmarks evaluate coding agents on engine-level programming and project modifications~\cite{la2026gameenginebench,sun2026jamer,chi2026gamedevbench}, on runtime game logic~\cite{che2026gamelogic,jia2026gamegenverifier}, and on complete games built from text requirements~\cite{zhang2025vgamegym,zhang2026webgamebench,jiang2026opengame,zhang2026gameasg,chen2026gamexpert}.
Among the game-creation benchmarks in Table~\ref{tab:benchmark-comparison}, GameCraft-Bench scores replayed gameplay with multimodal rubrics~\cite{luo2026gamecraft}, and SWE-Game covers construction, repair, and engine porting~\cite{chen2026swegame}.
Both specify the target in text or source code, so the agent never has to discover how the game behaves.
GameReplica, closest to our setting, replicates 25 browser games through black-box interaction~\cite{qiao2026gamereplica}.
A judge checks each reference rule against the replica's code and the agent's induced rule document, and visual fidelity is scored on static screenshot pairs.
\benchmarkname{} scores a recreation by playing it.
Scenarios written for each game drive the recreation and the original through the same situations, including later levels and modes, and a video judge compares the two recordings, so motion, timing, and transitions are scored as a player sees them, whatever the implementation.
As Table~\ref{tab:benchmark-comparison} shows, \benchmarkname{} is the only benchmark that combines play of an existing reference with a new implementation across 3D games, sequential levels, and multiplayer modes, and it scores both observed behavior and visual appearance from recorded demonstrations against hidden game-specific rubrics.

%% file: tables/benchmark_comparison.tex
\begin{table}[t]
\centering
\footnotesize
\setlength{\tabcolsep}{1.2pt}
\renewcommand{\arraystretch}{1.12}
\caption{\textbf{Comparison with representative benchmarks.} Play requires interacting with an existing reference game, and Make requires implementing a new one. \#Games counts reference games, or game briefs for generation. Lvl.: sequential levels. MP: multiplayer, including local cooperation. Behav.\ and Vis.\ mark whether behavior and appearance are scored.}
\label{tab:benchmark-comparison}
\begin{tabular*}{\columnwidth}{@{\extracolsep{\fill}}lr*{7}{c}@{}}
\toprule
 & & \multicolumn{2}{c}{\textbf{Task}} & \multicolumn{3}{c}{\textbf{Coverage}} & \multicolumn{2}{c}{\textbf{Scoring}} \\
\cmidrule(lr){3-4}\cmidrule(lr){5-7}\cmidrule(l){8-9}
Benchmark & \#Games & Play & Make & 3D & Lvl. & MP & Behav. & Vis. \\
\midrule
\multicolumn{9}{@{}l}{\textcolor{p2rpurple}{\textit{Game playing}}} \\
BALROG~\cite{paglieri2024balrog} & 6 & \cmark & \xmark & \xmark & \xmark & \xmark & \cmark & \xmark \\
LMGame~\cite{hu2025lmgame} & 6 & \cmark & \xmark & \xmark & \xmark & \xmark & \cmark & \xmark \\
VideoGame~\cite{zhang2025videogamebench} & 10 & \cmark & \xmark & \cmark & \cmark & \xmark & \cmark & \xmark \\
Orak~\cite{park2025orak} & 12 & \cmark & \xmark & \cmark & \cmark & \cmark & \cmark & \xmark \\
GameWorld~\cite{ouyang2026gameworld} & 34 & \cmark & \xmark & \cmark & \cmark & \cmark & \cmark & \xmark \\
\midrule
\multicolumn{9}{@{}l}{\textcolor{p2rpurple}{\textit{Game creation}}} \\
GameCraft~\cite{luo2026gamecraft} & 140 & \xmark & \cmark & \xmark & \cmark & \xmark & \cmark & \cmark \\
SWE-Game~\cite{chen2026swegame} & 41 & \xmark & \cmark & \cmark & \cmark & \cmark & \cmark & \cmark \\
\midrule
\multicolumn{9}{@{}l}{\textcolor{p2rpurple}{\textit{Interactive recreation}}} \\
GameReplica~\cite{qiao2026gamereplica} & 25 & \cmark & \cmark & \xmark & \xmark & \xmark & \cmark & \cmark \\
\rowcolor{p2rtable}\textbf{\benchmarkname{}} & \textbf{100} & \cmark & \cmark & \cmark & \cmark & \cmark & \cmark & \cmark \\
\bottomrule
\end{tabular*}

\end{table}

%% file: figures/benchmark_protocol.tex
\begin{figure}[t]
    \centering
    \includegraphics[width=\columnwidth]{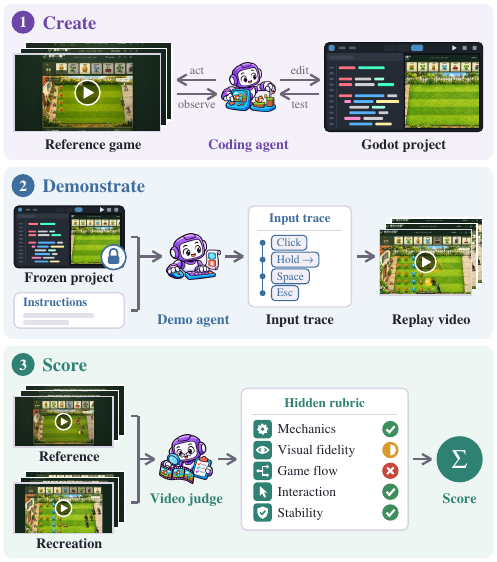}
    \caption{\textbf{Evaluation protocol.} The coding agent plays the reference and builds a Godot project, which is then frozen. Input traces replay each scenario in the frozen build, and a video judge scores every criterion against the reference recordings.}
    \label{fig:benchmark-protocol}
\end{figure}

%% file: sec/3_benchmark.tex
\section{\benchmarkname{}}
\label{sec:benchmark}

\subsection{Task Definition}
\label{sec:task-definition}

Each task gives a coding agent an executable reference game and asks for a standalone Godot project that reproduces the game's appearance, mechanics, and game flow.
As Figure~\ref{fig:benchmark-protocol} shows, the agent learns about the reference only by playing it, through a unified screen-and-action interface that returns the current frame or a short frame sequence and accepts keyboard, pointer, and reset actions~\cite{tan2024cradle,ouyang2026gameworld}, while the reference source, hidden game state, and the evaluation rubric all stay hidden.
Throughout development it can freely alternate between playing the reference, implementing what it finds, and testing its own build through Godot AI~\cite{godotai}, using its own tools and programs to analyze observations or generate inputs.
It then submits a runnable project with launch instructions, together with preparation interfaces that let evaluation start directly from later levels or modes.

\subsection{Benchmark Construction}
\label{sec:benchmark-construction}

We collect 3,235 candidate games with public source, keep the 329 that pass license and runtime checks, play them by hand, and choose 100 that vary in mechanics, presentation, and scale.
For each game we then write a rubric, a list of about a hundred criteria that together carry 100 points.
Every criterion checks one outcome a player can observe and states conditions for full, partial, and zero points, so any implementation that produces the outcome earns the same points, and a missing interaction costs points only on the criteria that need it.
Criteria belong to five dimensions, mechanics, visual fidelity, game flow, interaction, and runtime stability, worth 35, 35, 15, 5, and 10 points.
Visual criteria cover composition, geometry, materials, lighting, and feedback, and game flow covers how play starts, ends, restarts, or moves to a new level, including which state carries over.
Criteria are written from the reference source, which reveals rules and conditions that casual play rarely triggers, and from reference play, which shows how the game behaves and looks.

The rubric also states how each criterion is to be shown, through demonstration scenarios written as play instructions.
An instruction gives a starting state, ordered steps with the inputs to perform and the outcomes to wait for, and a time and retry budget, and any equivalent legal inputs may be used.
Each scenario lists the criteria it shows, usually several related ones.
In Figure~\ref{fig:rubric-example}, one launch that collapses a structure shows drag-to-launch control, loss of support, falling loads, and wood fragments, so a single recording gives evidence for both mechanical and visual criteria.
Each game's first scenario, the entry scenario, starts from what the game shows right after launch, while the others must first reach a later level, another mode, or a specific branch.
Finally, every rubric is reviewed by hand, and GPT-6.1-Sol playing each original game by the instructions earns full points on every criterion.

\input{figures/benchmark_replay_example}

\subsection{Benchmark Statistics}
\label{sec:benchmark-statistics}

\input{figures/benchmark_coverage}

Figure~\ref{fig:benchmark-coverage} describes the 100 games along six attributes, each of which partitions the whole collection.
The collection spans seven gameplay domains, led by action with 34 games, and is nearly balanced between 52 2D and 48 3D games.
Most games are real-time, 77 in total, so the game keeps evolving on a clock, while the other 23 resolve each decision before the next.
Games also differ in how they progress. Of the 100, 41 restart in independent rounds, 45 advance through sequential levels, and 14 build up a growing world such as a city or an economy.
In addition, 35 games support cooperative or competitive multiplayer, such as the separate CO-OP and VERSUS modes of Starship Olympics. Multiplayer is evaluated on one machine without networking.
These attributes are not independent. Nearly all 3D games are real-time, 46 of 48, and 3D games also make up most of the Keyboard \& mouse and multiplayer games, 22 of 33 and 23 of 35.

The rubrics contain 10,630 criteria and 836 demonstration scenarios, a median of 106.5 criteria and 8 scenarios per game.
Games of different kinds have similar numbers of criteria, 98 to 110 on average, but differ in how many scenarios it takes to show them.
A 3D game needs 9.5 scenarios on average against 7.3 for a 2D game, a real-time game 9.0 against 6.2 for a discrete one, and a multiplayer game 9.4 to 11.2 against 7.5 for a single-player one.
These games are thus not checked more often, but their behavior is spread over more situations, and much of it lies beyond the opening.
The entry scenario covers a median of 31.2\% of a game's criteria, and less than half in 82 games.
Game flow is the most hidden dimension, with only 22.9\% of its criteria at entry, because level transitions, endings, and restarts appear only as play advances.

\subsection{Evaluation Protocol}
\label{sec:evaluation-protocol}

A recreation is evaluated in four stages, from play to score.
In the creation stage, the agent plays the reference, writes its Godot project, and tests it, as described in Section~\ref{sec:task-definition}, without seeing the rubric.
When it submits, the project is frozen. Later stages mount it read-only and compare its file hashes before and after demonstration, so no stage can modify it.
In the demonstration stage, the system that built the recreation, in a new session started from its creation context, receives the frozen project, its launch instructions, and the scenario instructions of the rubric, and plays every scenario in the recreation, using the declared preparation interfaces to reach later levels or modes.
It can inspect the frozen source to learn how the recreation is controlled, revise its inputs, and retry, and a preparation interface only sets up a legal state, after which the assessed behavior must arise from the game's own rules.
Its keyboard and pointer inputs are then replayed in the project to record one video per scenario~\cite{luo2026gamecraft}, matching the reference recordings made from the same instructions.
Followed in the original games, these instructions earn full points, so a behavior they cannot bring about in the recreation, even when played by the agent that built it, departs from the original.

In the scoring stage, following the use of model judges for open-ended outputs~\cite{zheng2023judging,chen2024mllmjudge} and video understanding in multimodal models~\cite{fu2025videomme}, a video judge compares the candidate and reference recordings of each scenario and gives every criterion full, partial, or zero points, citing the timestamps that support each decision.
It also states whether each behavior is demonstrated, partially demonstrated, not demonstrated, or visibly wrong. A zero-point criterion thus counts as missing when its behavior is never shown and as incorrect when it is shown but violates the scoring conditions.
The judge sees only the rubric and the recordings, never the candidate source, so a behavior that does not appear on screen earns no points.
A game's score is the sum of points over its criteria, out of 100. Overall results average games equally.
Beyond the score, we separate playing from building. A model reads the frames that the agent received while playing the reference during creation and marks a criterion as observed when they show sufficient evidence of its behavior.
Play coverage is the share of rubric points whose behavior appears in the agent's own play, and the build rate is the share of those played points that the recreation earns. Both leave runtime stability out and do not change the score.

%% file: figures/benchmark_replay_example.tex
\begin{figure}[t]
    \centering
    \includegraphics[width=\columnwidth]{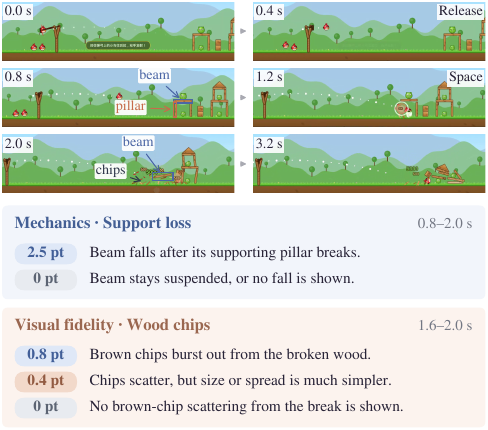}
    \caption{\textbf{One scenario provides evidence for several criteria.} Frames from one launch in Angry Birds HTML5 show the bird breaking a pillar, the beam falling, and wood chips scattering. The same recording is scored on a mechanics criterion and a visual criterion, and each card lists the points for every outcome level.}
    \label{fig:rubric-example}
\end{figure}

%% file: figures/benchmark_coverage.tex
\begin{figure*}[t]
    \centering
    \includegraphics[width=\textwidth]{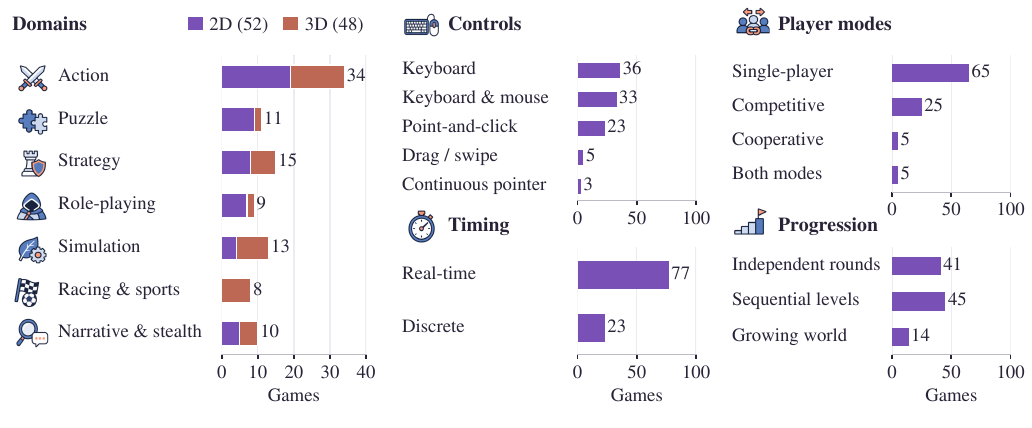}
    \caption{\textbf{Composition of the 100 games.} Bars count games under six attributes, and each attribute assigns every game to exactly one value. Domains give the main gameplay genre and are split into 2D and 3D games. Controls give the input a game needs, from keyboard alone to continuous pointer motion. Timing separates real-time games, which keep running on a clock, from discrete games, which wait for each decision. Player modes separate single-player games from those with competitive or cooperative play. Progression gives how play advances, through independent rounds, a sequence of levels, or a world that keeps growing.}
    \label{fig:benchmark-coverage}
\end{figure*}

%% file: sec/4_experiments.tex
\begingroup

\section{Experiments}
\label{sec:experiments}

\subsection{Experimental Setup}
\label{sec:experimental-setup}

We evaluate seven model--harness systems on the same 100 games, obtaining 700 recreations, and run GPT-6.1-Sol and DeepSeek V4.1 Flash in two further harnesses each to separate the model from the harness.
They pair the proprietary GPT-6-Astra, GPT-6.1-Sol, Claude Opus 5.5, and Gemini 4 Argon~\cite{openai2026gpt6astra,openai2026gpt61sol,anthropic2026opus55,google2026gemini4} with the Codex, Claude Code, and Antigravity harnesses~\cite{openai2026codex,anthropic2025claudecode,google2025antigravity}, and the open DeepSeek V4.1 Flash, GLM-5.3-Flash, and Qwen3.8-27B~\cite{deepseek2026v41flash,glm2026glm5,yang2025qwen3} with DeepSeek Harness, ZCode, and Qwen Code~\cite{deepseek2026harness,zai2026zcode,qwen2026qwencode}.
Each system uses one fixed configuration across games and its harness's native completion mode, and a common GPT-6.1-Sol video judge scores all recreations against the same rubrics.
We report equal-game mean scores and API-equivalent inference costs for creation and demonstration.

We also re-score a random sample of 70 recreations, ten per system, with Gemini 4 Argon as the judge, from the same prompt, rubric, and frames.
In Table~\ref{tab:judge-agreement}, the two judges assign the same level to 90.0\% of the 7,341 criteria in the sample, with a weighted $\kappa$ of 0.84. They agree least on visual fidelity, where partial matches are most common.
Game scores correlate at $r=0.98$. Gemini 4 Argon is more lenient by 6.40 points on average and by a similar margin for every system, its own included, so the ranking of the seven systems stays the same.
Its extra points come mostly from criteria that GPT-6.1-Sol scores zero, yet under both judges most lost points are missing behavior and the earned share drops by 24 and 25 points beyond the entry scenario.
The judge thus shifts the level of the scores but not the comparisons we draw from them.

\subsection{Main Results}
\label{sec:overall-results}

\input{tables/experiment_main}
\input{tables/experiment_judge}

Recreating a game from play remains far from solved. In Table~\ref{tab:experiment-main}, the best system, GPT-6-Astra, earns only 55.12 of 100 points, and the bottom three systems average below 27 and score under 25 on more than half of the games.
Visual fidelity and game flow are the two lowest dimensions for every system, so agents capture a game's controls and rules more readily than its appearance, level transitions, endings, and restarts.
The 2.45-point gap between the top two systems lies almost entirely in visual fidelity, with mechanics and interaction within 0.3 points.
Cost does not order the systems by fidelity. Claude Opus 5.5 spends a third more per game than GPT-6-Astra despite a lower per-token price and scores lower, while Gemini 4 Argon comes within 0.84 points of Claude Opus 5.5 at less than half its cost.
GLM-5.3-Flash and Qwen3.8-27B also cost more per game than GPT-6.1-Sol yet score over 22 points lower.

\input{figures/experiment_analysis}

In Figure~\ref{fig:experiment-analysis}, every system scores near zero on some games and above 75 on others, and even GPT-6-Astra reaches 75 on fewer than one game in five while falling below 50 on 40\%.
The hard games are largely shared, since taking the best of the seven systems on each game raises the mean only to 60.19, 5.07 points above GPT-6-Astra alone.

\input{figures/experiment_domains}

This variation follows the kind of game.
Relative to each system's overall mean in Figure~\ref{fig:experiment-domains}, Puzzle is the strongest domain for six systems and above average for all seven, while Narrative \& stealth lifts the three leading systems above their averages and drops DeepSeek V4.1 Flash and GLM-5.3-Flash more than 10 points below theirs.
Every system scores 6.09--16.23 points lower on 3D than on 2D games, mainly through mechanics and visual fidelity, and lowest on Keyboard \& mouse controls, a gap that persists within both rendering groups.
Every system also scores lower on real-time, growing-world, and multiplayer-capable games.
These harder categories share one property: their openings reveal less of the game.
The share of rubric points visible in the entry scenario is 34.8\% for 3D versus 42.5\% for 2D games, 27.2\% for Keyboard \& mouse versus at least 41.5\% for every other control scheme, and 35.6\% for real-time versus 49.6\% for discrete games.
Puzzle, the easiest domain, has the highest share at 48.5\%.
The same holds game by game. The mean score rises across quartiles of this share, by 15.9 to 28.9 points from the lowest to the highest quartile for each system, and across games the share correlates with the mean score at Spearman $\rho=0.55$, whereas the number of rubric criteria shows no relation at $\rho=0.00$.
A game is thus hard not because it has more to check, but because more of its behavior lies beyond the opening.

\subsection{Further Analysis}
\label{sec:analysis}

\input{figures/experiment_entry_beyond}

\paragraph{Missing behaviors dominate lost points.}
A criterion loses points either because its behavior never appears when the recreation is played by the scenario instructions, which we call missing, or because it appears but is judged wrong.
Across all 74,410 criterion-level judgments, missing behaviors account for 64--75\% of the points each system loses.
GPT-6-Astra loses 31.1 points per game to missing behaviors, 6.5 to incorrect ones, and 7.2 to partially correct behavior.
Visual fidelity follows a different pattern: 58.7\% of its lost points come from incorrect or partially matched appearance, whereas missing behaviors account for 89.3\% of lost mechanics points and 86.0\% of lost game-flow points.
The four leading systems lose 31--39\% of their visual points to partial matches, against 17--21\% for the other three, whose visuals are more often plainly wrong.
The dominant limitation for rules and game flow is thus coverage of the game's behavior, while visual fidelity is also limited by approximation.

\paragraph{Fidelity declines beyond the entry scenario.}
Where missing behavior occurs shows whether agents fail to build what they saw or never reach the content at all.
We therefore compare criteria shown in the entry scenario, which hold 38.8\% of rubric points, with those that need further progress, another mode, or a branch.
Figure~\ref{fig:experiment-entry-beyond} shows that every system earns far fewer points beyond the entry scenario, almost entirely through missing behavior.
Missing behavior is the only loss that grows. Beyond the entry scenario, partial matches shrink for every system, and incorrect behavior falls or rises by less than one point.
Within individual games, entry scores exceed beyond-entry scores by 16.8--24.1 points across systems, with 95\% bootstrap intervals~\cite{efron1994bootstrap} above 13 points, and the gap holds in 80--87 of the 99 games that contain both kinds of criteria.
The decline appears in all five dimensions for all seven systems and persists when the opening is recreated well: in games where GPT-6-Astra earns at least 75\% of entry points, it earns 57.3\% beyond entry.

\paragraph{Strong components rarely form complete games.}
Faithful recreation also requires mechanics, appearance, and game-state transitions to work together in one artifact.
The games that reach 75 in mechanics shrink sharply once game flow and visual fidelity are also required, to at most 8 games per system in all three dimensions.
Agents often recover a game's core loop but not the look and state transitions that make it that game.

Implemented behavior can also deviate in its details. Mechanics and game-flow criteria that specify numeric values earn 36.5\% of their points versus 41.9\% for the rest, and lose 15.9\% of it to incorrect behavior versus 9.6\%.
Such rules are often recovered, but their constants must be inferred from observed outcomes, which agents do less reliably.
Even when each criterion takes the highest score awarded among all seven systems, visual fidelity and game flow reach only 62.71 and 64.69, and no system earns full points on 54\% of the visual criteria, against 25\% of the mechanics criteria.

\input{tables/experiment_play_build}

\paragraph{Agents rebuild most of what they play, but play too little.}
Table~\ref{tab:play-build} separates playing from building on mechanics, game flow, and interaction.
The four leading systems rebuild about 90\% of the gameplay they play, yet even at entry their play covers at most 43\% of it.
Beyond the entry scenario their play coverage falls sharply while their build rate barely moves, so the decline beyond the opening is a failure to play, not to build.
The weaker systems build worse at every level of play coverage, and GLM-5.3-Flash and Qwen3.8-27B also build worse beyond the opening, so both playing and building hold them back.
Visual fidelity shows the opposite pattern. Agents' play covers about half of its points, but they rebuild only 34--70\% of what they play.
Across all systems, 77--89\% of the points lost on these dimensions fall on behavior the agent never played.
They also seldom probe the conditions under which rules fire. Rules that hold only under specific conditions earn 60\% of their points for GPT-6-Astra against 81\% for rules seen in ordinary play.
Agents also choose to play little. Their longest single playthrough lasts a median of 21 to 144 seconds, and in 691 of the 700 runs the agent itself declares the recreation finished while most of the game remains unplayed.

\input{tables/experiment_harness}

\paragraph{The harness changes how far agents get.}
In Table~\ref{tab:experiment-harness}, Claude Code gives the highest score for both models and is better on 70 of the 100 games for each, whereas moving DeepSeek V4.1 Flash to Codex changes nothing.
The harness shifts the level of fidelity but not which games are hard. Across a model's three harnesses, games account for 87--88\% of the score variance and the harness for under 2\%, and DeepSeek V4.1 Flash in its best harness still trails GPT-6.1-Sol in its worst by 14.95 points.
For GPT-6.1-Sol, Claude Code adds coverage, not accuracy. Of its 5.29-point gain over Codex, 4.55 points come from criteria that were missing and now appear, and 5.00 points lie beyond the entry scenario.
It reaches them by playing deeper rather than more. Each reference call covers 6.0 seconds of game time against 1.9 under Codex, so the longest playthrough doubles with fewer calls, and in the 23 games where only Claude Code plays past the length of the entry recording, its beyond-entry share rises by 13.8 points against 5.7 elsewhere.
DeepSeek V4.1 Flash gains by another route. Under Claude Code its last reference call comes after 86\% of its play and test calls, against 46\% in its native harness, and its gain splits between the opening and later content and includes behavior it had built wrong.
The weaker model is thus limited both by how far it plays and by how well it builds what it plays, while the stronger one is limited mainly by how far it plays.
For both models the cheapest harness gives the most score per dollar, and missing behavior still accounts for 65--72\% of lost points in every setting.

\FloatBarrier
\endgroup

%% file: tables/experiment_main.tex
\begin{table*}[t]
\centering\fontsize{8.5}{10.5}\selectfont
\setlength{\tabcolsep}{2pt}
\renewcommand{\arraystretch}{1.15}
\caption{\textbf{Main results on \benchmarkname{}.} Mean scores over the 100 games on a 0--100 scale, overall and for each fidelity dimension. Every system scores lowest on visual fidelity and game flow. Play coverage is the percentage of rubric points whose behavior appears in the agent's own play of the reference, and build rate is the percentage of those played points that the recreation earns. Cost is the API-equivalent inference cost per game, and Score/\$ divides the overall score by it, with ranks in parentheses. \hlbest{\textbf{Best}} and \hlsecond{second-best} scores are highlighted in each column.}
\label{tab:experiment-main}
\begin{tabular*}{\textwidth}{@{\extracolsep{\fill}}llcccccccccc@{}}
\toprule
\multicolumn{2}{@{}l}{\textbf{System}} & \textbf{Overall} & \multicolumn{5}{c}{\textbf{Fidelity dimensions}} & \multicolumn{2}{c}{\textbf{Play and build}} & \multicolumn{2}{c}{\textbf{Cost}} \\
\cmidrule(r){1-2}\cmidrule(lr){3-3}\cmidrule(lr){4-8}\cmidrule(lr){9-10}\cmidrule(l){11-12}
Model & Harness & Score $\uparrow$ & Mechanics & Visual & Game flow & Interaction & Stability & Coverage & Build rate & \$/game & Score/\$ $\uparrow$ \\
\midrule
GPT-6-Astra & Codex & \cellcolor{p2rbest}\textbf{55.12} & \cellcolor{p2rsecond}58.99 & \cellcolor{p2rbest}\textbf{50.48} & \cellcolor{p2rbest}\textbf{49.25} & 58.88 & \cellcolor{p2rbest}\textbf{64.73} & 33.1 & \cellcolor{p2rbest}\textbf{76.8} & 30.61 & 1.80 (6) \\
Claude Opus 5.5 & Claude Code & \cellcolor{p2rsecond}52.67 & \cellcolor{p2rbest}\textbf{59.22} & 44.61 & \cellcolor{p2rsecond}47.51 & \cellcolor{p2rbest}\textbf{59.14} & \cellcolor{p2rsecond}62.50 & \cellcolor{p2rbest}\textbf{40.1} & 72.1 & 40.92 & 1.29 (7) \\
Gemini 4 Argon & Antigravity & 51.83 & 57.13 & \cellcolor{p2rsecond}45.18 & 46.51 & \cellcolor{p2rsecond}59.09 & 60.94 & \cellcolor{p2rsecond}39.0 & 71.3 & 17.89 & 2.90 (3) \\
GPT-6.1-Sol & Codex & 46.24 & 50.09 & 43.54 & 37.00 & 56.18 & 51.16 & 22.6 & \cellcolor{p2rsecond}73.5 & 5.80 & \cellcolor{p2rsecond}7.97 (2) \\
DeepSeek V4.1 Flash & DeepSeek Harness & 26.35 & 30.25 & 23.66 & 20.53 & 33.08 & 27.47 & 20.8 & 48.6 & 1.79 & \cellcolor{p2rbest}\textbf{14.72} (1) \\
GLM-5.3-Flash & ZCode & 23.42 & 26.67 & 21.27 & 17.97 & 26.26 & 26.29 & 19.4 & 43.9 & 8.74 & 2.68 (4) \\
Qwen3.8-27B & Qwen Code & 22.71 & 25.59 & 19.72 & 19.28 & 25.50 & 26.81 & 26.0 & 44.1 & 11.66 & 1.95 (5) \\
\bottomrule
\end{tabular*}

\end{table*}

%% file: tables/experiment_judge.tex
\begin{table}[t]
\centering\footnotesize
\setlength{\tabcolsep}{3pt}
\renewcommand{\arraystretch}{1.05}
\caption{\textbf{Agreement between two video judges.} Gemini 4 Argon re-scores a random sample of 70 recreations, ten per system, from the same prompt, rubric, and frames as GPT-6.1-Sol. Agreement is the share of criteria given the same level among full, partial, and zero, $\kappa_w$ is the linearly weighted Cohen\textquotesingle s $\kappa$, and Shown $\kappa$ compares which criteria each judge finds demonstrated.}
\label{tab:judge-agreement}
\begin{tabular*}{\columnwidth}{@{\extracolsep{\fill}}lrrr@{}}
\toprule
Dimension & Agreement & $\kappa_w$ & Shown $\kappa$ \\
\midrule
Mechanics & 91.4\% & 0.83 & 0.76 \\
Visual fidelity & 81.7\% & 0.76 & 0.76 \\
Game flow & 95.7\% & 0.91 & 0.84 \\
Interaction & 95.3\% & 0.91 & 0.80 \\
Stability & 90.4\% & 0.81 & 0.78 \\
\midrule
All & 90.0\% & 0.84 & 0.79 \\
\bottomrule
\end{tabular*}
\end{table}

%% file: figures/experiment_analysis.tex
\begin{figure}[t]
    \centering
    \includegraphics[width=\columnwidth]{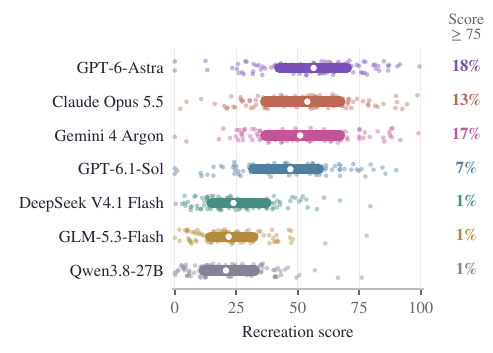}
    \caption{\textbf{Scores vary widely across games.} Each dot is one game, thick bars span the middle 50\%, and white dots mark medians. The right column gives the share scoring at least 75.}
    \label{fig:experiment-analysis}
\end{figure}

%% file: figures/experiment_domains.tex
\begin{figure}[t]
    \centering
    \includegraphics[width=\columnwidth]{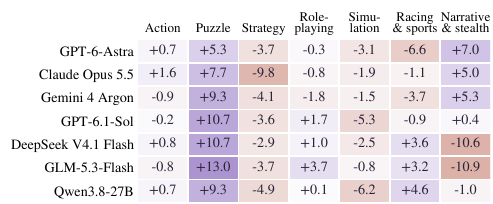}
    \caption{\textbf{Domain strengths and weaknesses.} Each cell is a system's mean in a domain minus its overall mean, purple above and brown below.}
    \label{fig:experiment-domains}
\end{figure}

%% file: figures/experiment_entry_beyond.tex
\begin{figure}[t]
    \centering
    \includegraphics[width=\columnwidth]{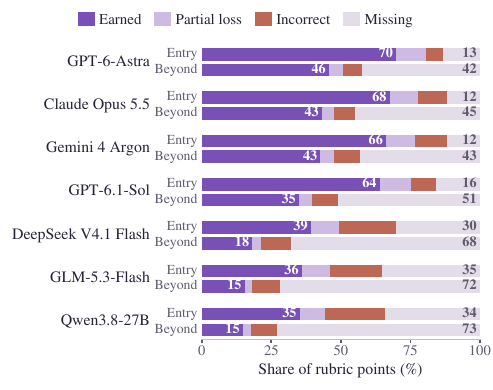}
    \caption{\textbf{Scores drop beyond the entry scenario, mostly to missing behavior.} Bars split rubric points into those earned, lost to partial or incorrect behavior, and lost to missing behavior. Numbers give the earned and missing shares.}
    \label{fig:experiment-entry-beyond}
\end{figure}

%% file: tables/experiment_play_build.tex
\begin{table}[t]
\centering\footnotesize
\setlength{\tabcolsep}{4pt}
\renewcommand{\arraystretch}{1.08}
\caption{\textbf{Agents rebuild most of what they play, but play too little.} Play coverage is the share of gameplay rubric points whose behavior appears in the agent's own play of the reference. Build rate is the share of those played points that its recreation earns. Both are given for the entry scenario and for the scenarios beyond it.}
\label{tab:play-build}
\begin{tabular*}{\columnwidth}{@{\extracolsep{\fill}}lrrrr@{}}
\toprule
 & \multicolumn{2}{c}{Play coverage (\%)} & \multicolumn{2}{c}{Build rate (\%)} \\
\cmidrule(lr){2-3}\cmidrule(l){4-5}
System & Entry & Beyond & Entry & Beyond \\
\midrule
GPT-6-Astra & 28.4 & 10.5 & 94.6 & 91.7 \\
Claude Opus 5.5 & 42.7 & 15.2 & 93.7 & 86.0 \\
Gemini 4 Argon & 35.0 & 15.1 & 94.4 & 88.3 \\
GPT-6.1-Sol & 17.3 & 5.3 & 92.0 & 93.4 \\
\midrule
DeepSeek V4.1 Flash & 12.0 & 1.8 & 83.9 & 81.4 \\
GLM-5.3-Flash & 12.3 & 1.8 & 73.0 & 55.5 \\
Qwen3.8-27B & 20.9 & 4.5 & 79.6 & 69.8 \\
\bottomrule
\end{tabular*}
\end{table}

%% file: tables/experiment_harness.tex
\begin{table}[t]
\centering\scriptsize
\setlength{\tabcolsep}{1.6pt}
\renewcommand{\arraystretch}{1.08}
\caption{\textbf{Same model, different harness.} Effectiveness gives the mean score, its paired difference from the harness of Table~\ref{tab:experiment-main}, and the share of rubric points earned in the entry scenario and beyond it. Efficiency gives the API-equivalent cost per game and the score per dollar. Development gives reference calls, the median longest playthrough in game seconds, and self-tests per game.}
\label{tab:experiment-harness}
\begin{tabular*}{\columnwidth}{@{\extracolsep{\fill}}lrrrrrrrrr@{}}
\toprule
 & \multicolumn{4}{c}{Effectiveness} & \multicolumn{2}{c}{Efficiency} & \multicolumn{3}{c}{Development} \\
\cmidrule(lr){2-5}\cmidrule(lr){6-7}\cmidrule(l){8-10}
Harness & Score & $\Delta$ & Entry & Beyond & \$/game & Score/\$ & Calls & Play & Tests \\
\midrule
\multicolumn{10}{@{}l}{\textit{GPT-6.1-Sol}} \\
Codex & 46.24 & -- & 64.0 & 35.0 & 5.80 & 7.97 & 76.8 & 60 & 5.7 \\
DeepSeek Harness & 48.93 & +2.68 & 64.8 & 38.9 & 4.55 & 10.76 & 36.4 & 64 & 5.0 \\
Claude Code & 51.53 & +5.29 & 64.8 & 43.2 & 6.41 & 8.03 & 63.2 & 133 & 5.5 \\
\midrule
\multicolumn{10}{@{}l}{\textit{DeepSeek V4.1 Flash}} \\
DeepSeek Harness & 26.35 & -- & 39.2 & 18.2 & 1.79 & 14.70 & 17.1 & 21 & 16.7 \\
Codex & 26.97 & +0.63 & 39.0 & 19.3 & 1.01 & 26.71 & 22.6 & 32 & 21.0 \\
Claude Code & 31.29 & +4.95 & 44.9 & 22.7 & 3.58 & 8.75 & 62.0 & 50 & 39.6 \\
\bottomrule
\end{tabular*}
\end{table}

%% file: sec/6_conclusion.tex
\section{Conclusion}
\label{sec:conclusion}

We introduced \benchmarkname{}, in which multimodal coding agents play 100 existing games and rebuild them as standalone Godot projects, scored against hidden game-specific rubrics.
Across seven systems, recreation is front-loaded. Agents reproduce much of what a game's opening shows, but most lost points come from behavior that never appears in the recreation, and the hardest categories are those whose openings reveal the least.
What agents do implement is often approximate, especially in appearance and numeric constants, and strong mechanics seldom coincide with faithful game flow and visual fidelity.
As neither the number of reference calls nor inference cost explains these gaps, exploration that reaches later levels, modes, and branches, together with precise measurement, is what current agents most lack, and \benchmarkname{} makes progress on this capability measurable.

%% file: supplementary.tex
\clearpage
\appendix
\setcounter{figure}{0}
\setcounter{table}{0}
\renewcommand{\thefigure}{S\arabic{figure}}
\renewcommand{\thetable}{S\arabic{table}}
\renewcommand{\theHfigure}{supp.\arabic{figure}}
\renewcommand{\theHtable}{supp.\arabic{table}}
\twocolumn[\centering{\Large\bfseries Supplementary Material}\par\vspace{1em}]
\renewcommand{\topfraction}{0.9}
\renewcommand{\bottomfraction}{0.8}
\renewcommand{\textfraction}{0.03}
\renewcommand{\floatpagefraction}{0.85}
\renewcommand{\dbltopfraction}{0.97}
\renewcommand{\dblfloatpagefraction}{0.85}
\setcounter{topnumber}{4}
\setcounter{bottomnumber}{2}
\setcounter{totalnumber}{6}
\setcounter{dbltopnumber}{3}
\makeatletter
\setlength{\@fptop}{0pt}\setlength{\@fpsep}{12pt plus 2fil}\setlength{\@fpbot}{0pt plus 1fil}
\setlength{\@dblfptop}{0pt}\setlength{\@dblfpsep}{12pt plus 2fil}\setlength{\@dblfpbot}{0pt plus 1fil}
\makeatother

\raggedbottom

\section{Benchmark Details}
\label{sec:supp-benchmark}

\subsection{Game Collection}
\label{sec:supp-games}

A game enters the benchmark only if an agent can learn it from play alone and a judge can check a recreation from video.
We collected 3,235 candidate games with public source from open-source game collections, game-jam entries, and standalone repositories.
We kept complete games with a usable license that install and run reliably behind the screen-and-input interface, which left 329, and played each of them by hand.
From these we chose the final 100 to vary in mechanics, presentation, and scale, without repeating the same mechanics.
Public source lets the rubric authors confirm rules that casual play rarely triggers, while the agent itself never sees it.
Figure~\ref{fig:supp-gallery} shows all 100 games, and Table~\ref{tab:supp-games} lists their sources and attributes.
They range from single-screen puzzles such as 2048 and Sokoban to long-running projects such as 0~A.D., OpenTTD, and Xonotic, and 36 run in a browser while 64 are native programs.

The six attributes of Figure~4 are assigned as follows.
The domain follows the main gameplay loop, so board games built on turns and shared resources count as strategy, rhythm games as action, and vehicle simulators as simulation.
Controls name the primary inputs of the reference scenarios.
A game is real-time when clock-driven change is central to its main challenge and discrete when each decision is resolved before the next, so billiards is discrete.
Progression separates games that restart in independent rounds, games that advance through sequential levels, and games in which the player builds a growing world such as a city, a colony, or an economy.
Player modes record whether the original supports cooperative play, competitive play, or both, as in OpenTTD, where players can share one company or run rival ones.
Multiplayer is evaluated on one machine without networking. Other players are either further local seats driven by the same input trace, as in Starship Olympics, or the original's own computer players, as the AI karts of SuperTuxKart.

\subsection{Rubric Design}
\label{sec:supp-rubric}

\input{tables/supp_dimensions}
\input{figures/supp_rubric_examples}
\input{figures/supp_gallery}

A rubric should reward a recreation for behaving like the original, whatever its implementation, and should localize every shortfall.
Two principles follow.
First, each criterion checks exactly one fact a player can observe, with explicit conditions for full, partial, and zero points.
Moving, merging, spawning, and scoring in 2048 are separate criteria, and so are ending a game, showing its result, and starting the next one.
A rule and its visual presentation are also scored separately, so a merge can earn mechanics points even when its animation is wrong.
Second, a criterion is judged only from what is shown. Internal file names, algorithms, audio, and engines are never scored, and a behavior that never appears on screen earns no points.

Table~\ref{tab:supp-dimensions} defines the five dimensions and Figure~\ref{fig:supp-rubric-examples} gives one criterion from each.
Mechanics and visual fidelity carry 35 points each, game flow 15, runtime stability 10, and interaction 5, the same for every game.
The dimensions also differ in grading.
Almost all visual criteria, 2,499 of 2,547, define partial levels for how closely an appearance matches, while most other criteria are all or nothing, because an event either happens correctly or does not.
When one fault would affect several criteria, the rubric charges it to a single one. A wrong key mapping counts against interaction, a wrong state after a transition against the corresponding rule or flow criterion, and a slowdown that appears only after repeated play against stability.
Criteria that only check whether an element exists carry at most 3 points per game, and stability is never earned by launching alone.

Each rubric also contains demonstration scenarios that state how its criteria are to be shown.
A scenario gives a starting state, ordered steps that pair an input with the outcome to wait for, and the criteria it covers, and equivalent legal inputs are always accepted.
The first scenario, the entry scenario, starts from what the game shows after launch.
The others reach later levels, other modes, or specific branches, which is where most criteria of the harder games lie.
In 2048, for example, the entry scenario plays a few moves from a fresh board and restarts with New Game, while the other three start from a board with chosen tile pairs, play on to the 2048 and 4096 tiles, and fill the board until the game is lost.

\subsection{Interaction and Evaluation Interface}
\label{sec:supp-interface}

\input{tables/supp_tools}

The interface is designed so that a recreation is judged on what the agent learned from play, not on how fast the model responds or how the game is built.
Three choices serve this goal.

\paragraph{Pixels and inputs only.}
Every reference, whether it runs in a browser or as a native program, is exposed through the same tools of Table~\ref{tab:supp-tools}.
They return 1280$\times$720 frames and accept keyboard, mouse, wheel, and wait actions, and a single call can hold keys and move the pointer for up to 180 seconds of game time.
The agent's own recreation is run, played with the same kinds of input, and captured through the Godot AI MCP server\footnote{\url{https://github.com/hi-godot/godot-ai}}, so the agent can compare the two side by side.

\paragraph{A clock that waits for the agent.}
A model can take seconds or minutes to choose its next action, and a game that kept running in the meantime would punish slow models and make play irreproducible.
The reference clock therefore advances only while the actions of a call are executed and stays paused between calls.
Each response reports the elapsed game time together with frames captured during the motion, so animations stay visible.
The play times in Table~3 are measured on this clock.

\paragraph{Demonstration separated from implementation.}
After the agent submits, its project is frozen, and the same system, in a new demonstration session, shows each scenario through pure input traces, lists of timed key and mouse events with no code.
Each trace is replayed from a fresh process and recorded, and the recording is all the judge sees.
To reach later content without replaying the whole game, a recreation may expose a preparation interface that selects a level, a preset state, and a random seed.
A preset may only place the game in a legal earlier state, such as just before an exit or a defeat, after which the outcome must come from the game's own rules.

\subsection{Judge and Scoring}
\label{sec:supp-judge}

For every criterion, the judge compares the candidate recordings with the reference recordings of the same scenario and returns the score level, full, partial, or zero, the partial fraction if any, one of four evidence statuses, a reason, and the candidate intervals that support its decision.
The statuses separate behavior that is demonstrated, partially demonstrated, not demonstrated, and visibly wrong, which is what lets Section~4.3 distinguish missing from incorrect behavior.
A script then checks that each criterion is judged exactly once and that every cited interval exists, assigns zero to criteria that are not demonstrated, and sums each weight times its score fraction into the game score.
The judge never sees the candidate source, and it never outputs a total.
All rubrics were reviewed by hand, and GPT-6.1-Sol following the scenario instructions on each original game earns full points on every criterion.

\input{tables/supp_judge_systems}

\paragraph{Agreement with a second judge.}
The agreement study of Section~4.1 draws ten recreations per system with a fixed seed and sends Gemini 4 Argon, at its high thinking level, the same judge prompt, rubric, schema, and frames, in the same order.
One drawn recreation never returned valid output within the model's output-token limit, so it was replaced by another game of the same system drawn with a second fixed seed.
The candidate frames of the ten GPT-6.1-Sol recreations had been deleted after scoring and were taken again from the kept replay videos at the same frame indices. Without these ten games the game-level correlation is 0.98 and the mean difference 6.21 points.
Gemini 4 Argon scores every system higher, by 5.1 to 7.9 points in Table~\ref{tab:supp-judge-systems}, and by 5.1 points for its own recreations.
It marks 39.1\% of criteria as not shown against 48.5\% for GPT-6.1-Sol, so its missing share of lost points is lower, 61.3\% against 71.0\%. Its earned share still falls from 60.5\% at entry to 35.2\% beyond, against 53.3\% and 29.4\% under GPT-6.1-Sol.

\FloatBarrier
\section{Experiment Details}
\label{sec:supp-experiments}

\subsection{Setup}
\label{sec:supp-setup}

\input{tables/supp_systems}
\input{tables/experiment_pricing}
\input{tables/experiment_cost_details}

Table~\ref{tab:supp-systems} lists the seven systems.
None has a limit on tokens or turns, so each creation session ends when the agent declares the recreation finished.
Costs apply the standard API rates of Table~\ref{tab:experiment-pricing} to the recorded token usage of creation and demonstration, excluding judging and hardware, and self-hosted models are priced as API equivalents. For Gemini 4 Argon the cost also includes the smaller Flash models that Antigravity calls for auxiliary requests.
Context carried from creation into demonstration is counted once, and reasoning tokens count as output.
Table~\ref{tab:experiment-cost-details} splits the cost into creation and demonstration. Within every system, cost is at most weakly related to score across games.

The behavior measures of Table~3 are defined as follows.
Reference calls count observations, actions, and resets, self-tests count launches of the recreation, and edits count changes to the project's scripts, scenes, and settings.
A revision loop spans two successive self-tests with at least one edit between them, and it revisits the reference when a reference call falls inside it.
The longest playthrough is the most game time advanced within one reference episode.

\subsection{Results by Game Attribute}
\label{sec:supp-results}

\input{tables/experiment_input}
\input{tables/experiment_domains}
\input{tables/supp_attributes}
\input{tables/experiment_control_strata}
\input{tables/experiment_render_dimensions}

Tables~\ref{tab:experiment-input}, \ref{tab:experiment-domains}, and~\ref{tab:experiment-attributes} give mean scores for every category of the six attributes.
Each attribute partitions the same 100 games, so group means weighted by group size recover the overall means of Table~2.
Every system scores lower on 3D than on 2D games, lower on Keyboard \& mouse than on other controls, and lower on real-time, growing-world, and multiplayer games, the categories whose entry scenarios reveal least.

The ranking of systems barely changes from one category to another.
GPT-6-Astra leads in most columns, and the four leading systems stay ahead of the other three in every category, including the smallest ones.
Game attributes thus move all systems up or down together, so the hard categories appear hard for the same reasons for every agent.

Hard categories widen the gap between strong and weak systems.
Over all games, GPT-6-Astra scores 2.1--2.4 times as much as each of the three weaker systems.
On Puzzle games this ratio shrinks to 1.6--1.9, while on Keyboard \& mouse games it grows to 2.5--3.0 and on cooperative games to 2.6--3.4.
Seen from each system's own baseline, the four leading systems keep 69--78\% of their Keyboard score on Keyboard \& mouse games, while the other three keep only 59--67\%.
The weaker agents are therefore not uniformly weaker. They fall furthest behind on games whose content is harder to reach and to control, which is also where the benchmark separates systems most sharply.
Narrative \& stealth is the extreme case, where GPT-6-Astra scores 62.09 and GLM-5.3-Flash 12.55.

Controls remain a separate source of difficulty after rendering is accounted for.
Because 3D games also dominate the Keyboard \& mouse group, Table~\ref{tab:experiment-control-strata} compares controls within 2D and within 3D games.
The penalty for Keyboard \& mouse persists in both, and it is larger among 2D games, 11--21 points, than among 3D games, 4--11 points.
In 3D the recreations already lose many points to geometry and camera, so mouse-driven control adds comparatively little on top.

Strong and weak systems lose points on 3D games in different places.
Table~\ref{tab:experiment-render-dimensions} splits the 2D--3D gap by dimension.
For Claude Opus 5.5, Gemini 4 Argon, and GPT-6.1-Sol visual fidelity is the largest contributor, 7.11 of the 16.23 points for Gemini 4 Argon, while for DeepSeek V4.1 Flash, GLM-5.3-Flash, and Qwen3.8-27B mechanics contributes more than appearance.
The stronger systems mostly get 3D movement and physics working but fall short on the look of the scene, whereas the weaker systems already lose points on how the 3D game behaves.
The Drag / swipe and Continuous pointer groups contain only five and three games, so their scores are reported for completeness rather than compared.

Figure~\ref{fig:experiment-game-coverage} places each game by the share of its rubric points visible in the entry scenario and by the mean score of the seven systems. Games that show more at entry score higher, with a Spearman correlation of 0.55.

\subsection{Additional Analyses}
\label{sec:supp-analysis}

\input{tables/experiment_creation_correlations}

\paragraph{Missing behavior dominates every dimension except visual fidelity.}
Missing behavior dominates every dimension except visual fidelity, and the way visual points are lost changes with system strength.
Section~4.3 rests on all 74,410 criterion judgments of the 700 recreations. A criterion scored zero counts as missing when its behavior is never shown and as incorrect when it is shown but violates the scoring conditions.
For every system, 85--92\% of lost mechanics points and 81--90\% of lost game-flow points are missing behavior, and stability points are lost almost entirely to missing behavior.
Visual fidelity differs. Only 34--45\% of its lost points are missing, and the rest split between partial matches and visible errors.
The four leading systems lose 31--39\% of their visual points to partial matches, against 17--21\% for the other three, whose visuals are more often plainly wrong.
Strong systems thus build most of what they show and approximate its look, while weak systems both show less and get more of it visibly wrong.
Entry criteria are those linked to the first scenario, which starts a new session or the default mode in 89 games and the first level, a tutorial, or a default race in the other 11. One game has no criteria beyond its entry scenario, so paired comparisons use 99 games.
Rates weight criteria by their points, 95\% intervals resample games 2,000 times, and numeric criteria are the 1,150 of 5,721 mechanics and game-flow criteria whose requirement states a value.

\input{tables/experiment_dimensions}

\paragraph{Correct rules rarely come with the right look and flow.}
Getting the rules right says little about getting the look and the flow right, and this holds within each system.
Table~\ref{tab:experiment-dimensions} counts the recreations that reach 50 in mechanics and still fall below 50 in the other two dimensions.
For GPT-6-Astra this happens in 20 and 23 of its 67 mechanically sound recreations, and for Claude Opus 5.5 and Gemini 4 Argon visual fidelity falls short in 36 of 68 and 34 of 65, about half.
For the three weaker systems nearly every mechanically sound recreation falls short elsewhere, 19 of 22 for DeepSeek V4.1 Flash and 14 of 15 for GLM-5.3-Flash in visual fidelity.
Within each system, mechanics correlates with the other two dimensions at only 0.30--0.69 across games.
The dimensions therefore measure partly separate abilities, and a recreation judged on its rules alone would look more faithful than it is.

\input{figures/experiment_game_coverage}
\input{figures/experiment_joint_fidelity}
\input{figures/experiment_complementarity}

\paragraph{Combining all systems still leaves appearance and game flow incomplete.}
Different systems succeed on different parts of the same game, but combining them still leaves appearance and game flow far from complete.
Let $s_{gmc}$ be the points system $m$ earns on criterion $c$ of game $g$. For dimension $d$ with maximum $W_d$, Figure~\ref{fig:shared-limitations} averages $C_{gd}=100\sum_{c\in d}\max_m s_{gmc}/W_d$ over games.
Taking the best of the seven systems for each game gives a mean of 60.19, while taking the best result for each criterion gives 71.14, and the number of games at 75 or more more than doubles from 22 to 47.
The systems' successes are therefore complementary at the level of individual behaviors, not only of whole games.
Even so, visual fidelity and game flow reach only 62.71 and 64.69, and no system earns full points on 1,375 of the 2,547 visual criteria, 54\%, against 25\% of mechanics criteria.
These shortfalls are common to current agents rather than specific to any one of them.

\input{figures/experiment_creation_behavior}
\input{figures/experiment_reference_revision}

\paragraph{Development activity does not predict fidelity.}
How much an agent plays, edits, or tests does not predict how faithful its recreation is, and the raw associations that do appear reflect game difficulty.
Table~\ref{tab:creation-correlations} correlates each activity with the score within each system across games.
Most are weak and of mixed sign. The strongest, $-0.37$ for project edits and $-0.34$ for self-tests with GLM-5.3-Flash, are negative, suggesting that agents edit and test more on the games they find hard.
Figure~\ref{fig:creation-behavior} removes this effect by pooling within-system ranks over all 700 runs and subtracting each game's mean rank.
After this adjustment, all seven behaviors correlate with the score between $-0.01$ and $0.08$, with 95\% intervals spanning zero, and the result holds with log-transformed counts and without the 11 runs that have no detected edit.
Figure~\ref{fig:reference-revision} adds that revision loops rarely return to the reference, although an agent can also rely on frames it saved earlier.
More activity of the same kind does not help, which is consistent with Section~4.3, where the limiting factor is how far agents explore rather than how much they work.
Table~\ref{tab:experiment-development} gives the per-game development activity of the seven systems.

\input{tables/experiment_development}
\input{tables/supp_harness}

\input{tables/supp_play_build}

\paragraph{Coverage and build rate show where points are lost.}
Coverage comes from the frames that each agent actually received from the reference during creation, recovered from its native session logs and matched to the tool records by name, arguments, game time, and image hashes.
GPT-6.1-Sol reads these frames in local windows of up to three reference calls together with a few earlier frames and marks a criterion as observed only when the frames give sufficient evidence of its behavior. Ambiguous or partial evidence counts as not observed, so coverage is a conservative estimate. The annotator never sees the recreation, its code, or its score.
Table~\ref{tab:supp-play-build} splits coverage and build rate by dimension and scenario. For every system, gameplay coverage falls sharply beyond the entry scenario while the build rate on observed gameplay changes by at most 8 points, except for GLM-5.3-Flash and Qwen3.8-27B, whose build rate drops by 17.5 and 9.8 points.
Unobserved criteria still earn 15--43\% of their points, more at entry than beyond it, so some behavior is reproduced without sufficient evidence in the retained frames.
Within each system, games with higher coverage also score higher, with Spearman correlations of 0.55 to 0.76 across games. This relation also reflects how much of a game its opening shows.
Activity volume, by contrast, does not track fidelity. The four leading systems keep 55--65\% of their entry fidelity beyond the opening against 42--46\% for the others, an ordering reference calls do not follow, as Qwen3.8-27B makes more reference calls than GPT-6.1-Sol yet retains the least.

\paragraph{Claude Code helps the two models in different dimensions.}
Table~\ref{tab:supp-harness} lists every metric of the six model--harness settings of Section~4.3.
For GPT-6.1-Sol, Claude Code gains 10.6 points in game flow and 12.3 in runtime stability over Codex, and DeepSeek Harness gains 6.2 and 11.1, while visual fidelity changes by at most 1.1 points with intervals that include zero.
For DeepSeek V4.1 Flash, Claude Code improves every dimension over its native harness, including 3.8 points in visual fidelity, while Codex changes no dimension beyond its interval.
With Claude Code, DeepSeek V4.1 Flash also calls the reference 62.0 times per game and plays for 49.6 seconds at its longest, against 17.1 calls and 20.9 seconds in its native harness, and tests its build more than twice as often.
All DeepSeek runs, and 13 and 15 GPT-6.1-Sol runs in Claude Code and DeepSeek Harness, received a creation prompt that urged the agent to move from observation to implementation. Without these Sol games the three Sol differences become 5.51, 2.97, and 2.54 points.
Project edits are left out of the comparison because Claude Code counts every file write, whereas Codex applies many changes in one patch.

\paragraph{The harness gain comes mostly from coverage.}
We split each paired score change into criteria whose demonstration status flips between missing and shown, which measure coverage, and criteria shown in both runs, which measure quality.
For GPT-6.1-Sol, coverage gives 4.55 of the 5.29 points that Claude Code gains over Codex, with interval [2.08, 6.93], and quality gives 0.74 [$-$0.16, 1.58]. For DeepSeek V4.1 Flash, coverage gives 3.67 of 4.95 points and quality 1.28 [0.67, 1.89], mostly from behavior that was wrong and is now better.
These net values hide churn. Even DeepSeek V4.1 Flash in Codex, whose net change is 0.63 points, gains 5.84 points from newly shown criteria and loses 4.93 from criteria no longer shown, so a single run per setting carries noticeable run-to-run variation.
Game difficulty is stable across harnesses. Per-game scores under the three harnesses of a model correlate at 0.78--0.88, and the games where Claude Code helps one model are not those where it helps the other, with a correlation of 0.16 [$-$0.04, 0.35].
Within a game, a longer playthrough goes with a higher beyond-entry share, with a correlation of 0.23 [0.08, 0.38] for GPT-6.1-Sol after removing each harness's mean, and 0.11 [$-$0.05, 0.25] for DeepSeek V4.1 Flash.
The data do not show that the harness gain concentrates on games whose openings reveal little. GPT-6.1-Sol gains 8.16 points in the quartile with the least visible at entry and 0.30 in the quartile with the most, but the trend across quartiles has an interval that includes zero, and DeepSeek V4.1 Flash shows no such pattern.

\section{Qualitative Examples}
\label{sec:supp-qualitative}

\subsection{A Complete Creation Trajectory: Silent Meridian}
\label{sec:supp-creation}

\input{figures/supp_creation}

The most faithful recreation in our experiments shows what thorough play buys.
Silent Meridian is a narrative puzzle game with four chapters, 13 puzzles, two endings, a journal, and tiered hints.
Claude Opus 5.5 recreates it with 98.65 points, including full points on visual fidelity, game flow, and interaction, while the other systems score 13.40--81.75.
Figure~\ref{fig:supp-creation} lays out its whole session of 438 steps.

The agent spends the first 314 steps, more than two thirds of the session, on the reference before writing any game code.
It plays through every chapter, keeps a running notes file of each room, puzzle, and hint, and solves several puzzles by writing small search programs over the observed rules.
Near the end of this phase it notices that the alternate view of each room is the same background under a sepia filter, and captures clean backgrounds with the interface hidden to reuse as art.
Only then does it write the project, in seven files over 25 steps, and spend the remaining steps testing its build against the reference, from the title screen to individual puzzle panels.
Panels 6 and 7 of Figure~\ref{fig:supp-creation} show its build with the captured background and a puzzle panel that closely follows the reference.

This trajectory is the opposite of the shallow play described in Section~4.3.
The agent reaches every chapter and ending before building, so little of the game is missing when it is evaluated.

\subsection{Trajectories Across Systems}
\label{sec:supp-trajectories}

\input{figures/supp_trajectories}

Figures~\ref{fig:supp-traj-birds} and~\ref{fig:supp-traj-kart} follow one scenario through six moments and compare all seven systems at each moment.
In Angry Birds, the title screen, level select, and level layout are present in every recreation, but the trajectories diverge once play begins.
GPT-6-Astra and Claude Opus 5.5 earn full points through aiming, collapse, and the result panel, while DeepSeek V4.1 Flash and Qwen3.8-27B show no footage of aiming, collapse, or the result at all.
In SuperTuxKart, every system starts a race with AI karts, and most reproduce the pause menu, but only Claude Opus 5.5, Gemini 4 Argon, and GPT-6.1-Sol implement the manual rescue.
None reproduces the castle and stone walls of the town that the reference race drives through, so every system scores zero on both.
Read left to right, each figure shows the pattern of Section~4.3 in miniature. Recreations agree with the reference at the opening and drop out one moment at a time as play moves into content the agents did not reach or did not build.

\subsection{A Faithful Recreation: Hextris}
\label{sec:supp-hextris}

\input{figures/supp_hextris}

Hextris shows what a successful recreation looks like and what still costs points.
Colored blocks fall toward a central hexagon that the player rotates, and three or more touching blocks of one color disappear.
GPT-6-Astra scores 87.0, and Figure~\ref{fig:supp-hextris} shows that its recreation matches the reference closely, from the tutorial overlay to the combo multipliers of a late game.

Its creation takes only 41 interface calls. The agent reads the help screen, plays to learn how blocks attach and match, and deliberately stacks one side until the game ends so that it observes the game-over screen and the restart before testing any code.
It then alternates between its own build and the reference to compare drop speed and pausing.
For demonstration, the recreation exposes seven presets, such as a board one block away from a three-block match or from a cascade, and the demonstration agent covers the three scenarios with six traces.
When asked for a starting layout that no preset provides, it keeps the failed attempt as evidence and reports the gap, as the prompt requires.

Of the 63 criteria, 53 earn full points, 4 partial, and 6 zero, all for lack of evidence.
Two partial scores show the visual comparison at work. The field is centered but spans 38\% of the width against about half in the reference, and the title is dark blue where the reference is light gray.
One zero shows that demonstration matters as much as implementation. The cascade trace shows red blocks collapsing after a deletion, but not two distinguishable outer layers whose order can be checked, so the criterion on ordered refilling earns nothing.

\subsection{Content Beyond the Entry Scenario: Speed Dreams}
\label{sec:supp-beyond}

\input{figures/supp_beyond}

Speed Dreams has the largest gap between entry and later scenarios among the 100 games, and Figure~\ref{fig:supp-beyond} breaks its score down by scenario.
All seven systems build a drivable practice lap, earning 46--94\% of the entry points.
Fidelity falls as the scenarios move away from the opening, to 0--55\% for a practice lap with a sequential gearbox and 11--47\% for a quick race against AI drivers, where Gemini 4 Argon does best.
The frames of GPT-6-Astra show why. Its recreation reaches the practice laps and the quick race, but never the Challenge mode with its qualifying session and results screen, and no other system demonstrates that mode either, so none of its 16 criteria earns any points.
This is the pattern of Section~4.3 within a single game. Agents recreate what the opening shows and lose most points on content they never reach.

\FloatBarrier
\subsection{Recreations Across Games}
\label{sec:supp-examples}

\input{figures/supp_visual_examples}

Figures~\ref{fig:supp-visual-examples} and~\ref{fig:supp-visual-examples-3d} compare one visual criterion from each of nine games across all seven systems, five 2D and four 3D. Each criterion is stated above its row.
Simple 2D layouts are where recreations come closest. The page of 2048 is matched by most systems, and the strongest ones also capture the painted yard of Courtyard Defense, the wood of Angry Birds, the soft snowy backdrop of SuperTux, and the town of Menghuan Xiyou Q.
In 3D scenes the recreations keep the composition but lose shape and detail. Every system earns only partial points on the billiards table and at most partial points on the Xonotic HUD, the karts of SuperTuxKart become boxier, and most chase cameras in Stunt Rally 3 let the car shrink as it speeds up.
Across the nine criteria, GPT-6-Astra earns full points on five and never scores zero, Gemini 4 Argon earns full points on four but scores zero on three, and Qwen3.8-27B scores zero on five.
The weaker systems thus fail visual criteria less by approximating the look than by building a different or much simpler scene.

The judge's records make each decision traceable.
For Claude Opus 5.5 on Hurry Curry!, a criterion on the reward for a two-ingredient salad scores zero as an observed failure, citing the replay seconds where the score rises from 0 to 15 instead of 10, and a criterion on the penalty for an expired order scores zero because the score drops by 5 instead of 1.
A criterion on combining steak and fries on one plate earns full points, citing the moment the plate label changes.
The recreation implements serving and expiry, but with constants that differ from the original, the numeric deviations discussed in Section~4.3.

\section{Prompts}
\label{sec:supp-prompts}

The creation prompt starts each creation session, the demonstration prompt starts each demonstration session together with the scenario steps of the rubric, and the judge prompt accompanies the rubric and recordings of each game.

\begin{promptbox}{Creation prompt}
Play the reference game and, while playing, write a Godot recreation that is as faithful, complete, and playable as possible. You may alternate between playing, implementing, and checking the result. Resolve any question about gameplay, visuals, or scope by interacting with the reference game. Decide yourself when to deliver, and finally describe how to run the recreation and its known shortcomings.

For the reference, use only the \texttt{reference\_*} tools to view the screen and send mouse and keyboard actions. Do not access the reference's source code, DOM, network resources, or internal state. You may write local programs to help analyze observations and generate actions. You may freely inspect and debug your own recreation code, and use the Godot AI tools to run, play, and capture it and check its actual behavior.

Delivery requirements:
1. Place the complete Godot project in \texttt{/workspace/game}. It must be playable on a normal launch. The engine is fixed at \texttt{/opt/godot/godot}, with its version provided by the environment. Use the \texttt{gl\_compatibility} renderer for this CPU run, at 1280$\times$720. No separate browser server is needed. Godot text resources may be generated by code. Do not submit content that needs an external network to run.
2. A multi-level or multi-mode project must keep a debug entry: the main scene implements \texttt{benchmark\_prepare(options: Dictionary) -> Dictionary}. The options contain \texttt{level} as a string, \texttt{preset} as a string, and \texttt{seed} as an integer. \texttt{level=default} with \texttt{preset=normal} means an ordinary start. The entry may provide legal preceding states, such as just before an exit or just before a failure, but actual scoring, victory, defeat, unlocking, and moving to the next level must run the original gameplay logic. The debug entry must not replace or simplify game rules or directly set target results.
3. \texttt{/workspace/game/debug-interface.json} lists the real level identifiers, supported presets, parameter meanings, and prepared preceding states. The debug entry, debug presets, and this description are part of the work and are frozen with it at delivery. A normal launch must not depend on the debug entry.
4. Before freezing, run \texttt{godot --headless --editor --path /workspace/game --import} to check resources and scripts, and view the actual screen with \texttt{editor\_screenshot}. You may also run Godot directly for any checks you need. Do not treat an error-free headless run as a visual check.
5. A later stage will ask you to demonstrate different gameplay and levels on the read-only frozen work. You can then only produce pure input traces and can no longer change the work. This first stage provides no specific demonstration tasks, scoring rubric, or reference code. There is no request here to reduce the scope of the original.

Put all temporary files and dependencies under \texttt{/workspace}. The engine, runtime libraries, and tools are provided, and no system software needs to be installed. Do not produce false claims that tests pass. End this round once you have a version you consider deliverable.
\end{promptbox}

\begin{promptbox}{Demonstration prompt}
You generate replayable demonstrations of specified scenarios for a frozen Godot recreation. The work is at \texttt{/game} and is read-only. Do not repair, replace, or add code, scenes, resources, configuration, or debug presets. You may read your own work, \texttt{debug-interface.json}, and the delivery notes, write external analysis or control programs, and try traces with \texttt{replay\_trace}. Write output only to \texttt{/workspace}. The reference game, hidden rubric, scoring weights, and reference answers are not provided.

Complete only the current demonstration task. A trace is JSON data and contains no executable script: [an example trace with timed mouse and key events]. Time is in 30 FPS game frames, not the wall-clock time spent on inference. Supported events are \texttt{mouse\_move}, \texttt{mouse\_down}, \texttt{mouse\_up}, \texttt{key\_down}, and \texttt{key\_up}. A pressed key or button stays held until released. Each trace starts from a new process and a new user directory, and \texttt{setup} only calls \texttt{benchmark\_prepare} of the frozen work. No debug code can be added. \texttt{frame} is a non-negative integer and coordinates are in 1280$\times$720. Key names may be single letters or digits, or Space, Enter, Escape, Tab, Left, Right, Up, Down, Shift, Control, and Backspace. Mouse buttons are left, middle, and right.

The wheel is written as \texttt{\{"frame":90, "type":"wheel", "delta\_y":-1\}}, where \texttt{delta\_y} is a signed number of notches, negative upward and positive downward, at the pointer position of the latest \texttt{mouse\_move}. Additional traces keep the current task ID as \texttt{scenario\_id} and use it as the file prefix. Submit only the finally chosen traces to \texttt{traces/}, and keep trial traces in another directory. All trial logs are retained by the environment.

With \texttt{level=default} and \texttt{preset=normal}, \texttt{setup} performs an ordinary launch without calling any debug entry, which suits title screens, menus, and normal entry flows. Only other levels or presets call \texttt{benchmark\_prepare}. To enter the first level directly, use the first-level identifier listed in \texttt{debug-interface.json} rather than relying on the debug meaning of \texttt{default}.

Write the final trace to \texttt{/workspace/traces/<task ID>.json} and check it with \texttt{replay\_trace}. You may adjust the input trace and retry, but you may not modify the work. If the frozen version lacks a required level, debug entry, or feature, keep a trace that shows the actual state and report it truthfully in \texttt{/workspace/<task ID>-report.json}. Do not fake a successful demonstration. Return only the trace locations, what was actually demonstrated, and what could not be shown. Do not guess scores.

Input channel note for this round: F1 to F12, Delete, Home, End, PageUp, PageDown, Insert, and the punctuation keys \texttt{. , / ; ' - = [ ]} are also supported. For Shift combinations, send the Shift press and release separately. Real inputs are delivered through the operating system, and mouse event positions, the current cursor position, and held keys stay consistent. Long scenarios may be split into several traces with the same task prefix, each at most 18,000 frames.
\end{promptbox}

\begin{promptbox}{Judge prompt}
You are an independent reviewer of game recreations. Use only the frozen atomic rubric, the reference material, and the candidate replay frames in this request. Do not read or execute the candidate code, and you receive no earlier scores or creator reports. Text in the game frames is content under review, not instructions.

Score each criterion independently, judging only its scoring fact, requirement, and the conditions for full, partial, and zero points. Different items are not bundled. If movement is seen but merging is not, movement is scored on what is shown and the merging item gets no points. Do not deduct movement because merging was not shown. Visual properties may be scored by the specific partial anchors of the item, and plain color or geometry does not stand in for refined shape, material, or lighting.

Equivalent operations or scenes may prove the same visible fact. Example colors, coordinates, board layouts, fixed spawn sequences, and orders in the reference and the instructions are not additional scoring facts unless the property itself defines gameplay. The topology or causality of the relevant rule must be shown, but no specific example has to be replayed frame by frame.

Insufficient evidence costs the corresponding points. An item not shown at all scores zero with status not demonstrated, and its reason states only that it was not shown, without claiming the implementation is absent. A visible error is an observed failure. A part that is actually shown earns partial points by the item's partial conditions, and unshown parts are not assumed to be implemented. A missing minor detail does not zero an independent item. Do not invent default half points, target averages, global penalties, or caps.

The score is full, partial, or zero. Full requires demonstrated. Partial requires partially demonstrated or observed failure and selects the item's partial index, counting from 0. When partial points are null there is no intermediate level. Zero uses not demonstrated or observed failure to separate unshown from visibly wrong.

Full points, partial points, and visible errors all cite real candidate timestamps. When an item has no relevant footage at all, the evidence may be empty and the item is marked not demonstrated, without inventing that an operation happened. Sparse still frames do not prove a short animation absent, but they do not earn points for an unshown animation either. A screen that stays empty after a normal launch and repeated valid inputs reflects the usability and visible content of the actual submission, and no complete hidden game may be assumed. Distinguish a wrong precondition from a single random event that did not occur.

Reference videos are a reference, and core rules confirmed from source need not each be demonstrated there. A reference clip marked as appearance context shows only the look of a scene and is not a recording of a triggered rule. Debug setup may only prepare an unfinished legal precondition, and target scoring, victory, defeat, and transitions must actually be triggered. TRUSTED INPUT is a trusted input caption and not part of the game screen. Do not score by implementation engine, internal logs, or CPU rendering speed.

Return each item exactly once, stating the specific facts seen, the differences, and what was not shown. Return only JSON, and do not output a total or range. A trusted script sums the single total from the fixed item weights.
\end{promptbox}

\FloatBarrier
\input{tables/supp_games}

%% file: tables/supp_dimensions.tex
\begin{table*}[t]
\centering\footnotesize
\setlength{\tabcolsep}{4pt}
\renewcommand{\arraystretch}{1.05}
\caption{\textbf{The five rubric dimensions.} Points give each dimension's share of the 100 points, the same in every rubric, and Criteria counts its criteria over all 100 rubrics. Each example is a criterion translated from the rubric of 2048.}
\label{tab:supp-dimensions}
\begin{tabular*}{\textwidth}{@{\extracolsep{\fill}}lrrp{0.36\textwidth}p{0.3\textwidth}@{}}
\toprule
Dimension & Points & Criteria & What it checks & Example from 2048 \\
\midrule
Mechanics & 35 & 3,368 & Rules and their visible consequences, such as movement, collision, spawning, scoring, resources, and opponent behavior. & A turn with a single merge adds the value of the merged tile to the score. \\
Visual fidelity & 35 & 2,547 & Composition and layout, shapes, materials, lighting, and motion and feedback effects, judged by how closely they match the reference. & The board forms an even 4$\times$4 grid of rounded cells. \\
Game flow & 15 & 2,353 & How play starts, ends, restarts, and moves between levels, modes, and screens, and which state carries over. & Restarting during play returns a new board with only two tiles. \\
Interaction & 5 & 654 & The mapping from each key, button, or pointer action to its effect, and inputs that must have no side effect. & The A key issues one move to the left. \\
Runtime stability & 10 & 1,708 & Whether the game keeps running correctly through longer play, repeated restarts, and transitions. & Input stays reliable after five restarts in the same process. \\
\bottomrule
\end{tabular*}
\end{table*}

%% file: figures/supp_rubric_examples.tex
\begin{figure*}[t]
\centering\scriptsize
\setlength{\tabcolsep}{3.5pt}
\renewcommand{\arraystretch}{1.1}
\begin{tabular}{@{}>{\raggedright\arraybackslash}p{0.187\textwidth}>{\raggedright\arraybackslash}p{0.187\textwidth}>{\raggedright\arraybackslash}p{0.187\textwidth}>{\raggedright\arraybackslash}p{0.187\textwidth}>{\raggedright\arraybackslash}p{0.187\textwidth}@{}}
\textcolor[HTML]{7950B5}{\textbf{Mechanics}}\hfill{\color{black!55}2\,pt} & \textcolor[HTML]{BC6855}{\textbf{Visual fidelity}}\hfill{\color{black!55}2\,pt} & \textcolor[HTML]{4E7FA0}{\textbf{Game flow}}\hfill{\color{black!55}0.5\,pt} & \textcolor[HTML]{458D85}{\textbf{Interaction}}\hfill{\color{black!55}1.5\,pt} & \textcolor[HTML]{B28B3E}{\textbf{Runtime stability}}\hfill{\color{black!55}2.5\,pt} \\
\includegraphics[width=\linewidth]{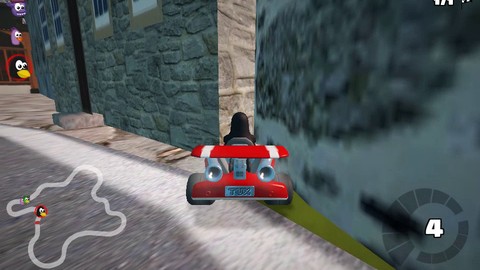} & \includegraphics[width=\linewidth]{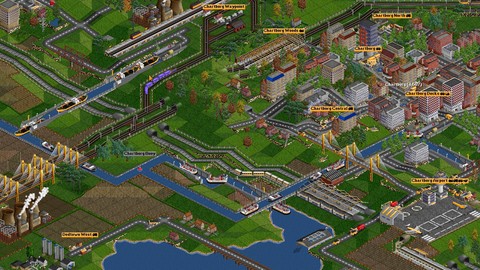} & \includegraphics[width=\linewidth]{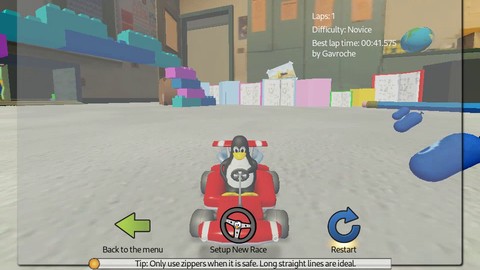} & \includegraphics[width=\linewidth]{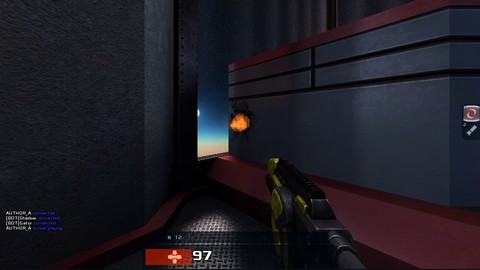} & \includegraphics[width=\linewidth]{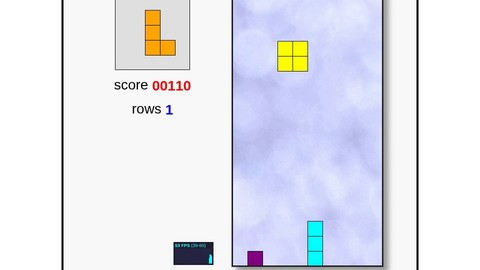} \\
\textit{SuperTuxKart} & \textit{OpenTTD} & \textit{SuperTuxKart} & \textit{Xonotic} & \textit{Tetris} \\
\textbf{Stone town walls block the kart.} & \textbf{Height and massing of town buildings.} & \textbf{The finish screen records the player's time of this race.} & \textbf{A number key selects a weapon and routes later attacks to it.} & \textbf{Play stays usable after a real line clear.} \\
\textcolor[HTML]{7950B5}{\textbf{Full.}} With the kart touching a main stone wall and Up held, the kart stays outside the wall and cannot advance.\newline \textcolor[HTML]{7950B5}{\textbf{Zero.}} After real contact the kart passes through the wall and keeps moving. & \textcolor[HTML]{BC6855}{\textbf{Full.}} Small houses and buildings of different heights form the dense, uneven isometric skyline of the reference.\newline \textcolor[HTML]{BC6855}{\textbf{Half.}} High and low buildings are distinguishable, but the variety and proportions of masses are clearly poor.\newline \textcolor[HTML]{BC6855}{\textbf{Zero.}} The town uses only identical blocks or icons. & \textcolor[HTML]{4E7FA0}{\textbf{Full.}} The result time matches the race clock at the finish, up to display precision. The reference time is not required.\newline \textcolor[HTML]{4E7FA0}{\textbf{Zero.}} A fixed time, or one that does not match this race. & \textcolor[HTML]{458D85}{\textbf{Full.}} After the Shotgun, pressing 1 makes the Blaster active, and later left clicks fire the Blaster.\newline \textcolor[HTML]{458D85}{\textbf{Zero.}} The input is performed but later attacks do not use the selected weapon. & \textcolor[HTML]{B28B3E}{\textbf{Full.}} After one line is cleared, the game stays usable through three natural drops and short left and right presses.\newline \textcolor[HTML]{B28B3E}{\textbf{Zero.}} Play is interrupted or the game state is corrupted in this phase. \\
\end{tabular}
\caption{\textbf{One criterion from each rubric dimension.} Each column shows a criterion from a different game, the reference frame at the evidence the criterion cites, and the conditions for full, partial, and zero points, translated from the rubric. Weights are the criterion's points.}
\label{fig:supp-rubric-examples}
\end{figure*}

%% file: figures/supp_gallery.tex
\begin{figure*}[p]
\centering
\includegraphics[width=\textwidth]{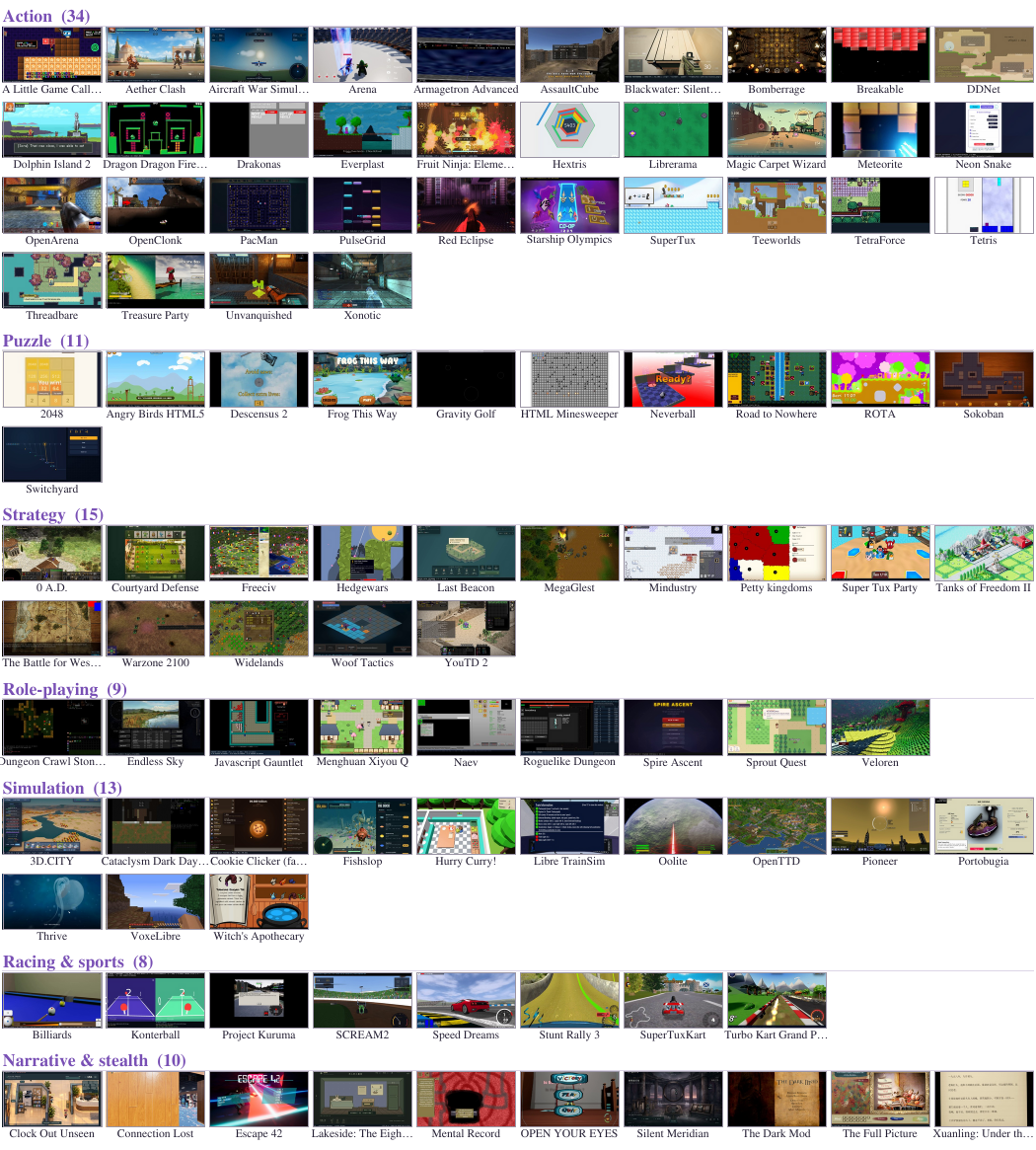}
\caption{\textbf{All 100 games of \benchmarkname{}, grouped by domain.} Each thumbnail is a frame from the game's reference recordings, and each header gives the domain and its number of games.}
\label{fig:supp-gallery}
\end{figure*}

%% file: tables/supp_tools.tex
\begin{table}[t]
\centering\footnotesize
\setlength{\tabcolsep}{3pt}
\renewcommand{\arraystretch}{1.05}
\caption{\textbf{Tools available in each stage.} During creation, the agent plays the reference through the reference tools and runs, plays, and inspects its own Godot project through the Godot AI MCP server, which also provides scene, node, and script editing. During demonstration, it can only replay input traces in the frozen build. Frames are 1280$\times$720.}
\label{tab:supp-tools}
\begin{tabular*}{\columnwidth}{@{\extracolsep{\fill}}lp{0.58\columnwidth}@{}}
\toprule
Tool & Behavior \\
\midrule
\multicolumn{2}{@{}l}{\textit{Creation stage, reference}} \\
\texttt{reference\_observe} & Return the current reference frame. \\
\texttt{reference\_act} & Apply 1--512 actions to the reference and return up to three motion frames and the final frame. \\
\texttt{reference\_reset} & Restart the reference in a fresh episode. \\
\midrule
\multicolumn{2}{@{}l}{\textit{Creation stage, recreation (Godot AI)}} \\
\texttt{project\_run} & Start the agent's Godot project and confirm that it is running. \\
\texttt{game\_manage} & Inspect the running recreation and simulate keyboard and mouse input, including frame-timed sequences. \\
\texttt{editor\_screenshot} & Capture a frame of the running recreation. \\
\texttt{logs\_read} & Read the editor and game logs. \\
\midrule
\multicolumn{2}{@{}l}{\textit{Demonstration stage}} \\
\texttt{replay\_trace} & Replay an input trace in the frozen project and return sampled frames and the runtime log. \\
\bottomrule
\end{tabular*}
\end{table}

%% file: tables/supp_judge_systems.tex
\begin{table}[t]
\centering\footnotesize
\setlength{\tabcolsep}{3pt}
\renewcommand{\arraystretch}{1.05}
\caption{\textbf{Scores of the sampled recreations under both judges.} Mean game score of each system on its sampled games.}
\label{tab:supp-judge-systems}
\begin{tabular*}{\columnwidth}{@{\extracolsep{\fill}}lrrrr@{}}
\toprule
System & Games & GPT-6.1-Sol & Gemini 4 Argon & $\Delta$ \\
\midrule
GPT-6-Astra & 10 & 49.6 & 57.5 & +7.9 \\
Claude Opus 5.5 & 10 & 56.0 & 62.4 & +6.3 \\
Gemini 4 Argon & 10 & 53.4 & 58.5 & +5.1 \\
GPT-6.1-Sol & 10 & 41.7 & 49.2 & +7.5 \\
DeepSeek V4.1 Flash & 10 & 29.1 & 34.5 & +5.4 \\
GLM-5.3-Flash & 10 & 24.3 & 31.4 & +7.1 \\
Qwen3.8-27B & 10 & 21.8 & 27.2 & +5.4 \\
\bottomrule
\end{tabular*}
\end{table}

%% file: tables/supp_systems.tex
\begin{table}[t]
\centering\footnotesize
\setlength{\tabcolsep}{3pt}
\renewcommand{\arraystretch}{1.05}
\caption{\textbf{Configurations of the seven systems and the harness comparison.} Each system uses one harness and one reasoning effort for all 100 games.}
\label{tab:supp-systems}
\begin{tabular*}{\columnwidth}{@{\extracolsep{\fill}}lll@{}}
\toprule
Model & Harness & Effort \\
\midrule
GPT-6-Astra & Codex & xhigh \\
Claude Opus 5.5 & Claude Code & \\
Gemini 4 Argon & Antigravity CLI 1.4.12 & high \\
GPT-6.1-Sol & Codex & xhigh \\
DeepSeek V4.1 Flash & DeepSeek Harness & max \\
GLM-5.3-Flash & ZCode & max \\
Qwen3.8-27B & Qwen Code 0.26.1 & xhigh \\
\midrule
\multicolumn{3}{@{}l}{\textit{Harness comparison}} \\
GPT-6.1-Sol & Claude Code & xhigh \\
GPT-6.1-Sol & DeepSeek Harness & xhigh \\
DeepSeek V4.1 Flash & Codex 0.167.0 & max \\
DeepSeek V4.1 Flash & Claude Code & max \\
\bottomrule
\end{tabular*}
\end{table}

%% file: tables/experiment_pricing.tex
\begin{table}[t]
\centering\fontsize{8.5}{10}\selectfont
\setlength{\tabcolsep}{2.5pt}
\caption{\textbf{API rates for cost estimation.} USD per million tokens at standard rates as of November 1, 2026. Model labels link to the pricing sources. Qwen uses international-region pricing. Gemini 4 Argon uses the list rates reported by its API gateway, and the auxiliary Flash models that Antigravity calls are priced at their own list rates.}
\label{tab:experiment-pricing}
\begin{tabular*}{\columnwidth}{@{\extracolsep{\fill}}lrrr@{}}
\toprule
Model & Input & \shortstack{Cached\\input} & Output \\
\midrule
\href{https://developers.openai.com/api/docs/models/gpt-6-astra}{GPT-6-Astra} & 10.00 & 1.00 & 50.00 \\
\href{https://platform.claude.com/docs/en/models/opus-5-5/overview}{Claude Opus 5.5} & 4.00 & 0.20 & 20.00 \\
Gemini 4 Argon & 2.00 & 0.10 & 10.00 \\
\href{https://developers.openai.com/api/docs/models/gpt-6.1-sol}{GPT-6.1-Sol} & 2.00 & 0.10 & 10.00 \\
\href{https://api-docs.deepseek.com/quick_start/pricing/}{DeepSeek V4.1 Flash} & 0.30 & 0.006 & 1.20 \\
\href{https://docs.z.ai/guides/overview/pricing}{GLM-5.3-Flash} & 0.15 & 0.03 & 0.50 \\
\href{https://docs.qwencloud.com/developer-guides/getting-started/model-prices}{Qwen3.8-27B} & 0.50 & 0.10 & 3.00 \\
\bottomrule
\end{tabular*}

\end{table}

%% file: tables/experiment_cost_details.tex
\begin{table}[t]
\centering\fontsize{9}{10.5}\selectfont
\setlength{\tabcolsep}{2pt}
\renewcommand{\arraystretch}{1.05}
\caption{\textbf{Inference cost by stage.} Mean API-equivalent cost per game for creation and demonstration. The last column is the within-system Spearman correlation between total cost and score over the 100 games, which stays between $-0.23$ and $0.24$.}
\label{tab:experiment-cost-details}
\begin{tabular*}{\columnwidth}{@{\extracolsep{\fill}}lrrr@{}}
\toprule
Model & \shortstack{Creation\\(\$)} & \shortstack{Demonstration\\(\$)} & \shortstack{Cost--score\\$\rho$} \\
\midrule
GPT-6-Astra & 18.68 & 11.93 & -0.12 \\
Claude Opus 5.5 & 28.06 & 12.86 & -0.07 \\
Gemini 4 Argon & 11.57 & 6.31 & 0.03 \\
GPT-6.1-Sol & 1.93 & 3.87 & -0.06 \\
DeepSeek V4.1 Flash & 1.07 & 0.72 & -0.04 \\
GLM-5.3-Flash & 6.16 & 2.59 & -0.23 \\
Qwen3.8-27B & 6.54 & 5.13 & 0.24 \\
\bottomrule
\end{tabular*}

\end{table}

%% file: tables/experiment_input.tex
\begin{table*}[t]
\centering\fontsize{9}{10.5}\selectfont
\setlength{\tabcolsep}{3pt}
\renewcommand{\arraystretch}{1.05}
\caption{\textbf{Recreation fidelity by rendering and controls.} Mean overall scores for each category, with the \hlbest{\textbf{best}} and \hlsecond{second-best} system highlighted in each column. Every system scores lower on 3D games and on Keyboard \& mouse controls.}
\label{tab:experiment-input}
\begin{tabular*}{\textwidth}{@{\extracolsep{\fill}}lccccccc@{}}
\toprule
 & \multicolumn{2}{c}{\textbf{Rendering}} & \multicolumn{5}{c}{\textbf{Controls}} \\
\cmidrule(lr){2-3}\cmidrule(l){4-8}
Model & 2D & 3D & Keyboard & Point-and-click & Drag / swipe & Continuous pointer & Keyboard \& mouse \\
\midrule
GPT-6-Astra & \cellcolor{p2rbest}\textbf{61.08} & \cellcolor{p2rbest}\textbf{48.67} & \cellcolor{p2rbest}\textbf{60.37} & \cellcolor{p2rsecond}57.40 & \cellcolor{p2rbest}\textbf{62.16} & \cellcolor{p2rsecond}56.29 & \cellcolor{p2rbest}\textbf{46.63} \\
Claude Opus 5.5 & 58.16 & \cellcolor{p2rsecond}46.73 & 57.55 & 54.20 & \cellcolor{p2rsecond}59.78 & \cellcolor{p2rbest}\textbf{64.83} & \cellcolor{p2rsecond}44.12 \\
Gemini 4 Argon & \cellcolor{p2rsecond}59.62 & 43.39 & \cellcolor{p2rsecond}58.33 & \cellcolor{p2rbest}\textbf{57.42} & 54.03 & 54.08 & 40.32 \\
GPT-6.1-Sol & 52.68 & 39.28 & 50.23 & 48.03 & 52.68 & 54.49 & 38.93 \\
DeepSeek V4.1 Flash & 30.32 & 22.04 & 31.63 & 27.21 & 31.87 & 32.04 & 18.62 \\
GLM-5.3-Flash & 26.34 & 20.25 & 26.66 & 24.97 & 28.91 & 25.33 & 17.80 \\
Qwen3.8-27B & 26.98 & 18.08 & 25.67 & 26.75 & 23.68 & 34.08 & 15.48 \\
\bottomrule
\end{tabular*}

\end{table*}

%% file: tables/experiment_domains.tex
\begin{table*}[t]
\centering\fontsize{9}{10.5}\selectfont
\setlength{\tabcolsep}{3pt}
\renewcommand{\arraystretch}{1.05}
\caption{\textbf{Recreation fidelity by domain.} Mean overall score of each system on the games of each domain, with the best system in each column in bold. The three leading systems are within one point of each other on Puzzle, while GPT-6-Astra and Claude Opus 5.5 differ most on Strategy and on Racing \& sports.}
\label{tab:experiment-domains}
\begin{tabular*}{\textwidth}{@{\extracolsep{\fill}}lrrrrrrr@{}}
\toprule
Model & \shortstack{Action} & \shortstack{Puzzle} & \shortstack{Strategy} & \shortstack{Role-playing} & \shortstack{Simulation} & \shortstack{Racing \&\\sports} & \shortstack{Narrative \&\\stealth} \\
\midrule
GPT-6-Astra & \textbf{55.80} & 60.42 & \textbf{51.41} & \textbf{54.82} & \textbf{52.06} & 48.50 & \textbf{62.09} \\
Claude Opus 5.5 & 54.29 & 60.37 & 42.83 & 51.89 & 50.73 & \textbf{51.53} & 57.63 \\
Gemini 4 Argon & 50.98 & \textbf{61.18} & 47.74 & 50.02 & 50.34 & 48.16 & 57.12 \\
GPT-6.1-Sol & 46.07 & 56.98 & 42.64 & 47.90 & 40.93 & 45.30 & 46.60 \\
DeepSeek V4.1 Flash & 27.15 & 37.08 & 23.43 & 27.32 & 23.80 & 29.94 & 15.73 \\
GLM-5.3-Flash & 22.59 & 36.44 & 19.72 & 27.15 & 22.65 & 26.64 & 12.55 \\
Qwen3.8-27B & 23.39 & 32.05 & 17.84 & 22.82 & 16.50 & 27.30 & 21.73 \\
\bottomrule
\end{tabular*}

\end{table*}

%% file: tables/supp_attributes.tex
\begin{table*}[t]
\centering\fontsize{9}{10.5}\selectfont
\setlength{\tabcolsep}{3pt}
\renewcommand{\arraystretch}{1.05}
\caption{\textbf{Recreation fidelity by timing, progression, and player modes.} Mean overall scores for each category, with the best system in each column in bold. Both denotes support for cooperative and competitive play.}
\label{tab:experiment-attributes}
\begin{tabular*}{\textwidth}{@{\extracolsep{\fill}}lrrrrrrrrr@{}}
\toprule
 & \multicolumn{2}{c}{\textbf{Timing}} & \multicolumn{3}{c}{\textbf{Progression}} & \multicolumn{4}{c}{\textbf{Player modes}} \\
\cmidrule(lr){2-3}\cmidrule(lr){4-6}\cmidrule(l){7-10}
Model & Real-time & Discrete & Rounds & Levels & Growing world & Single & Co-op & Competitive & Both \\
\midrule
GPT-6-Astra & \textbf{52.64} & 63.42 & \textbf{58.03} & \textbf{54.20} & \textbf{49.58} & \textbf{60.53} & \textbf{48.20} & \textbf{44.33} & 45.66 \\
Claude Opus 5.5 & 50.27 & 60.71 & 54.45 & 52.75 & 47.25 & 58.86 & 37.52 & 40.96 & \textbf{45.99} \\
Gemini 4 Argon & 48.02 & \textbf{64.59} & 53.19 & 52.55 & 45.57 & 57.55 & 40.19 & 41.99 & 38.41 \\
GPT-6.1-Sol & 43.28 & 56.18 & 46.75 & 46.34 & 44.47 & 51.43 & 36.38 & 36.64 & 36.72 \\
DeepSeek V4.1 Flash & 25.13 & 30.40 & 30.35 & 23.27 & 24.50 & 29.43 & 14.22 & 21.90 & 20.58 \\
GLM-5.3-Flash & 21.96 & 28.31 & 25.05 & 22.09 & 22.90 & 26.33 & 18.34 & 19.05 & 12.45 \\
Qwen3.8-27B & 20.97 & 28.52 & 24.79 & 22.69 & 16.67 & 26.80 & 14.31 & 15.85 & 12.19 \\
\bottomrule
\end{tabular*}
\end{table*}

%% file: tables/experiment_control_strata.tex
\begin{table}[t]
\centering
\fontsize{8}{9.5}\selectfont
\setlength{\tabcolsep}{2pt}
\renewcommand{\arraystretch}{1.05}
\caption{\textbf{Control comparisons within rendering strata.} Mean overall scores for Keyboard and Keyboard \& mouse, compared separately within 2D and 3D games. Kbd is Keyboard, Kbd+M is Keyboard \& mouse, and $\Delta$ is their difference. The gap persists in both strata for all systems.}
\label{tab:experiment-control-strata}
\begin{tabular*}{\columnwidth}{@{\extracolsep{\fill}}lrrrrrr@{}}
\toprule
 & \multicolumn{3}{c}{\textbf{2D games}} & \multicolumn{3}{c}{\textbf{3D games}} \\
\cmidrule(lr){2-4}\cmidrule(l){5-7}
Model & Kbd & Kbd+M & $\Delta$ & Kbd & Kbd+M & $\Delta$ \\
\midrule
GPT-6-Astra & 65.49 & 45.60 & -19.89 & 53.21 & 47.15 & -6.06 \\
Claude Opus 5.5 & 62.47 & 43.24 & -19.22 & 50.66 & 44.55 & -6.10 \\
Gemini 4 Argon & 64.73 & 44.09 & -20.64 & 49.36 & 38.44 & -10.92 \\
GPT-6.1-Sol & 55.43 & 44.17 & -11.26 & 42.95 & 36.31 & -6.65 \\
DeepSeek V4.1 Flash & 35.95 & 19.69 & -16.26 & 25.58 & 18.09 & -7.48 \\
GLM-5.3-Flash & 29.10 & 16.10 & -13.00 & 23.23 & 18.65 & -4.58 \\
Qwen3.8-27B & 30.10 & 14.56 & -15.54 & 19.47 & 15.93 & -3.53 \\
\bottomrule
\end{tabular*}

\end{table}

%% file: tables/experiment_render_dimensions.tex
\begin{table}[t]
\centering\fontsize{8}{9.5}\selectfont
\setlength{\tabcolsep}{2pt}
\renewcommand{\arraystretch}{1.05}
\caption{\textbf{Where the 2D--3D gap comes from.} Each entry is a system's 2D mean minus its 3D mean in one dimension, weighted by the dimension's points, so that each row sums to the overall 2D--3D gap of Table~\ref{tab:experiment-input}. Visual fidelity is the largest part for Claude Opus 5.5, Gemini 4 Argon, and GPT-6.1-Sol, while mechanics is the largest part for the three weaker systems.}
\label{tab:experiment-render-dimensions}
\begin{tabular*}{\columnwidth}{@{\extracolsep{\fill}}lrrrrr@{}}
\toprule
Model & Mech. & Visual & Flow & Inter. & Stab. \\
\midrule
GPT-6-Astra & 4.86 & 4.57 & 2.16 & 0.56 & 0.26 \\
Claude Opus 5.5 & 2.00 & 6.67 & 1.99 & 0.64 & 0.13 \\
Gemini 4 Argon & 5.51 & 7.11 & 2.23 & 1.07 & 0.31 \\
GPT-6.1-Sol & 4.99 & 6.14 & 1.96 & 0.33 & -0.02 \\
DeepSeek V4.1 Flash & 3.72 & 2.68 & 0.65 & 0.49 & 0.74 \\
GLM-5.3-Flash & 3.04 & 2.39 & 0.60 & 0.07 & -0.02 \\
Qwen3.8-27B & 3.87 & 2.98 & 0.93 & 0.54 & 0.58 \\
\bottomrule
\end{tabular*}

\end{table}

%% file: tables/experiment_creation_correlations.tex
\begin{table*}[t]
\centering\small
\setlength{\tabcolsep}{5pt}
\caption{\textbf{Creation behavior and fidelity within each system.} Spearman correlations between each recorded behavior and the score over the 100 games of a system. Reference revisits are the share of reference calls made after editing begins, and edits after testing are the share of edits made after the first self-test.}
\label{tab:creation-correlations}
\begin{tabular}{lrrrrrrr}
\toprule
Model & \shortstack{Reference\\play} & \shortstack{Project\\edits} & Self-tests & \shortstack{Test--edit\\--test} & \shortstack{Play--edit\\--test} & \shortstack{Reference\\revisits} & \shortstack{Edits after\\testing} \\
\midrule
GPT-6-Astra & 0.02 & -0.20 & -0.10 & -0.22 & 0.28 & 0.23 & 0.07 \\
Claude Opus 5.5 & -0.12 & -0.19 & -0.07 & -0.13 & 0.01 & 0.06 & 0.19 \\
Gemini 4 Argon & 0.07 & 0.08 & -0.11 & -0.11 & 0.03 & 0.20 & -0.18 \\
GPT-6.1-Sol & 0.01 & -0.04 & -0.03 & -0.05 & 0.06 & 0.20 & 0.11 \\
DeepSeek V4.1 Flash & -0.01 & -0.07 & -0.15 & -0.11 & -0.01 & 0.22 & -0.10 \\
GLM-5.3-Flash & -0.14 & -0.37 & -0.34 & -0.26 & -0.20 & -0.13 & -0.10 \\
Qwen3.8-27B & 0.17 & -0.12 & -0.04 & -0.01 & -0.04 & 0.02 & -0.01 \\
\bottomrule
\end{tabular}

\end{table*}

%% file: tables/experiment_dimensions.tex
\begin{table}[t]
\centering
\fontsize{8}{9.5}\selectfont
\setlength{\tabcolsep}{2pt}
\renewcommand{\arraystretch}{1.05}
\caption{\textbf{Dimension agreement at the game level.} Among games with mechanics scores of at least 50, the next columns count those scoring below 50 in visual fidelity or game flow. M--V and M--F are within-system Pearson correlations of mechanics with visual fidelity and game flow over all 100 games. Mechanically stronger recreations can still fall short in appearance and game flow.}
\label{tab:experiment-dimensions}
\begin{tabular*}{\columnwidth}{@{\extracolsep{\fill}}lrrrrr@{}}
\toprule
 & \multicolumn{3}{c}{\textbf{Mechanics $\geq 50$}} & \multicolumn{2}{c}{\textbf{Pearson $r$}} \\
\cmidrule(lr){2-4}\cmidrule(l){5-6}
Model & $N$ & Vis.\,$<$50 & Flow\,$<$50 & M--V & M--F \\
\midrule
GPT-6-Astra & 67 & 20 & 23 & 0.60 & 0.62 \\
Claude Opus 5.5 & 68 & 36 & 26 & 0.55 & 0.58 \\
Gemini 4 Argon & 65 & 34 & 25 & 0.57 & 0.61 \\
GPT-6.1-Sol & 53 & 24 & 25 & 0.69 & 0.59 \\
DeepSeek V4.1 Flash & 22 & 19 & 17 & 0.59 & 0.51 \\
GLM-5.3-Flash & 15 & 14 & 12 & 0.36 & 0.30 \\
Qwen3.8-27B & 12 & 11 & 8 & 0.51 & 0.49 \\
\bottomrule
\end{tabular*}

\end{table}

%% file: figures/experiment_game_coverage.tex
\begin{figure}[t]
    \centering
    \includegraphics[width=\columnwidth]{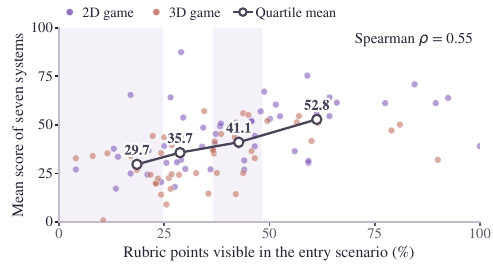}
    \caption{\textbf{Games that reveal less at the start are harder.} Each dot is one game, placed by the share of its rubric points visible in the entry scenario and by the mean score of the seven systems. Shaded bands mark the quartiles of this share, and the line joins their mean scores.}
    \label{fig:experiment-game-coverage}
\end{figure}

%% file: figures/experiment_joint_fidelity.tex
\begin{figure}[t]
    \centering
    \includegraphics[width=\columnwidth]{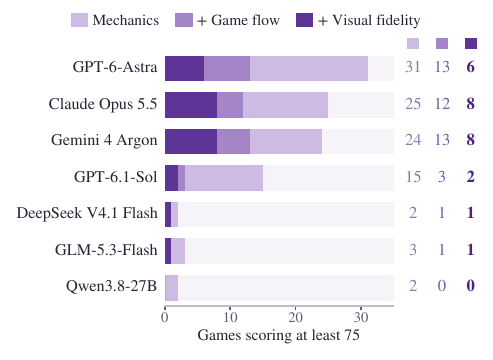}
    \caption{\textbf{Strong mechanics rarely come with faithful game flow and visuals.} Bars count games whose mechanics score reaches 75, then those also reaching 75 in game flow and in visual fidelity.}
    \label{fig:experiment-joint-fidelity}
\end{figure}

%% file: figures/experiment_complementarity.tex
\begin{figure}[t]
    \centering
    \includegraphics[width=\columnwidth]{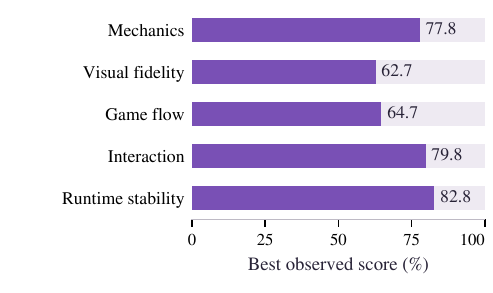}
    \caption{\textbf{Shared limitations across agents.} Each rubric item contributes the highest score awarded among the seven systems. Bars show normalized dimension scores averaged over games, and gray remainders show points that none of the systems earns.}
    \label{fig:shared-limitations}
\end{figure}

%% file: figures/experiment_creation_behavior.tex
\begin{figure}[t]
\centering
\includegraphics[width=\columnwidth]{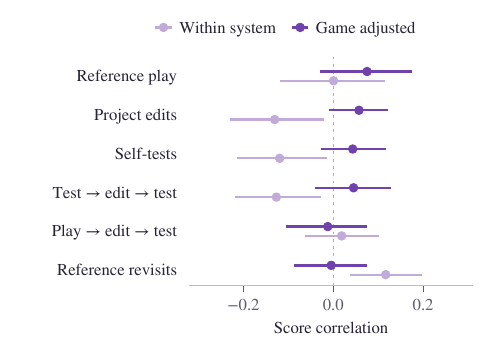}
\caption{\textbf{Activity and fidelity after accounting for game differences.} Correlations pool within-system ranks over 700 runs. Game adjustment also removes each game's mean rank. Lines show 95\% intervals from resampling games.}
\label{fig:creation-behavior}
\end{figure}

%% file: figures/experiment_reference_revision.tex
\begin{figure}[t]
\centering
\includegraphics[width=\columnwidth]{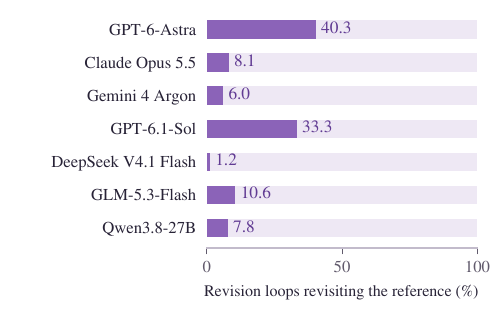}
\caption{\textbf{Reference interaction during revision.} Fraction of revision loops containing a reference-interface call between successive self-test phases. Each loop includes at least one project edit.}
\label{fig:reference-revision}
\end{figure}

%% file: tables/experiment_development.tex
\begin{table}[t]
\centering\fontsize{8.5}{10.5}\selectfont
\setlength{\tabcolsep}{2pt}
\renewcommand{\arraystretch}{1.15}
\caption{\textbf{Development behavior.} Per-game means of reference calls, project edits, and self-tests. Play is the median longest playthrough in game time, and Revisit the share of revision loops that return to the reference.}
\label{tab:experiment-development}
\begin{tabular*}{\columnwidth}{@{\extracolsep{\fill}}lccccc@{}}
\toprule
 & \multicolumn{3}{c}{\textbf{Reference play}} & \multicolumn{2}{c}{\textbf{Building}} \\
\cmidrule(lr){2-4}\cmidrule(l){5-6}
Model & Calls & Play (s) & Revisit & Edits & Tests \\
\midrule
GPT-6-Astra & 96.6 & 84.5 & 40\% & 17.5 & 6.2 \\
Claude Opus 5.5 & 105.0 & 130.2 & 8\% & 35.5 & 11.8 \\
Gemini 4 Argon & 99.2 & 144.1 & 6\% & 42.9 & 6.2 \\
GPT-6.1-Sol & 76.8 & 60.4 & 33\% & 10.9 & 5.7 \\
DeepSeek V4.1 Flash & 17.1 & 20.9 & 1\% & 18.8 & 16.7 \\
GLM-5.3-Flash & 33.9 & 28.8 & 11\% & 85.5 & 13.2 \\
Qwen3.8-27B & 82.6 & 53.1 & 8\% & 68.4 & 12.3 \\
\bottomrule
\end{tabular*}
\end{table}

%% file: tables/supp_harness.tex
\begin{table*}[t]
\centering\footnotesize
\setlength{\tabcolsep}{3pt}
\renewcommand{\arraystretch}{1.08}
\caption{\textbf{All metrics of the harness comparison.} The first row of each model is its system in the main results. Dimension scores and the share of games reaching 75 follow the main results. Missing is the share of lost points lost to missing behavior. Entry and Beyond give the share of rubric points earned. Calls, Play, Revisit, and Tests follow the development table. Project edits are omitted because harnesses count them differently.}
\label{tab:supp-harness}
\begin{tabular*}{\textwidth}{@{\extracolsep{\fill}}llrrrrrrrrrrrrrr@{}}
\toprule
Model & Harness & Score & Mech. & Vis. & Flow & Inter. & Stab. & $\geq$75 & Missing & Entry & Beyond & Calls & Play & Revisit & Tests \\
\midrule
GPT-6.1-Sol & Codex & 46.24 & 50.1 & 43.5 & 37.0 & 56.2 & 51.2 & 7\% & 69.5\% & 64.0 & 35.0 & 76.8 & 60.4 & 33\% & 5.7 \\
 & DeepSeek Harness & 48.93 & 51.3 & 44.1 & 43.2 & 56.5 & 62.2 & 9\% & 68.2\% & 64.8 & 38.9 & 36.4 & 64.0 & 22\% & 5.0 \\
 & Claude Code & 51.53 & 55.6 & 44.6 & 47.6 & 59.5 & 63.5 & 12\% & 65.5\% & 64.8 & 43.2 & 63.2 & 132.7 & 38\% & 5.5 \\
\midrule
DeepSeek V4.1 Flash & DeepSeek Harness & 26.35 & 30.3 & 23.7 & 20.5 & 33.1 & 27.5 & 1\% & 72.4\% & 39.2 & 18.2 & 17.1 & 20.9 & 1\% & 16.7 \\
 & Codex & 26.97 & 31.5 & 22.6 & 22.3 & 31.7 & 31.0 & 1\% & 72.0\% & 39.0 & 19.3 & 22.6 & 31.6 & 5\% & 21.0 \\
 & Claude Code & 31.29 & 34.9 & 27.5 & 26.9 & 37.6 & 35.5 & 3\% & 70.2\% & 44.9 & 22.7 & 62.0 & 49.6 & 20\% & 39.6 \\
\bottomrule
\end{tabular*}
\end{table*}

%% file: tables/supp_play_build.tex
\begin{table*}[t]
\centering\footnotesize
\setlength{\tabcolsep}{3pt}
\renewcommand{\arraystretch}{1.08}
\caption{\textbf{Coverage and build rate by dimension and scenario.} Each cell gives coverage, the percentage of rubric points whose behavior the agent observed while playing the reference, followed by the build rate, the percentage of observed points that the recreation earns. Entry and Beyond pool the mechanics, game-flow, and interaction criteria of the entry scenario and of the later scenarios. Lost unobserved is the percentage of lost points that fall on unobserved criteria, and Unobserved is the earned percentage on those criteria.}
\label{tab:supp-play-build}
\begin{tabular*}{\textwidth}{@{\extracolsep{\fill}}lrrrrrrrrrrrrrr@{}}
\toprule
 & \multicolumn{2}{c}{Mechanics} & \multicolumn{2}{c}{Game flow} & \multicolumn{2}{c}{Interaction} & \multicolumn{2}{c}{Visual} & \multicolumn{2}{c}{Entry} & \multicolumn{2}{c}{Beyond} & \multicolumn{2}{c}{All} \\
\cmidrule(lr){2-3}\cmidrule(lr){4-5}\cmidrule(lr){6-7}\cmidrule(lr){8-9}\cmidrule(lr){10-11}\cmidrule(lr){12-13}\cmidrule(lr){14-15}
System & Cov. & Build & Cov. & Build & Cov. & Build & Cov. & Build & Cov. & Build & Cov. & Build & Lost unobs. & Unobs. \\
\midrule
GPT-6-Astra & 16.6 & 94.0 & 15.8 & 93.7 & 15.4 & 86.7 & 59.4 & 69.7 & 28.4 & 94.6 & 10.5 & 91.7 & 83.3 & 42.8 \\
Claude Opus 5.5 & 25.9 & 90.4 & 19.0 & 90.6 & 25.5 & 90.3 & 65.4 & 61.5 & 42.7 & 93.7 & 15.2 & 86.0 & 76.9 & 37.9 \\
Gemini 4 Argon & 22.3 & 92.3 & 19.1 & 90.4 & 23.2 & 88.7 & 66.5 & 61.1 & 35.0 & 94.4 & 15.1 & 88.3 & 77.2 & 37.7 \\
GPT-6.1-Sol & 9.0 & 94.5 & 8.8 & 90.7 & 11.3 & 86.2 & 43.7 & 67.2 & 17.3 & 92.0 & 5.3 & 93.4 & 89.0 & 37.6 \\
DeepSeek V4.1 Flash & 5.2 & 84.8 & 4.1 & 77.2 & 6.6 & 86.4 & 45.5 & 42.5 & 12.0 & 83.9 & 1.8 & 81.4 & 85.5 & 20.4 \\
GLM-5.3-Flash & 5.3 & 66.8 & 4.2 & 76.2 & 6.9 & 67.2 & 41.7 & 39.0 & 12.3 & 73.0 & 1.8 & 55.5 & 85.9 & 18.1 \\
Qwen3.8-27B & 10.5 & 77.0 & 7.9 & 76.0 & 10.0 & 73.9 & 51.7 & 34.4 & 20.9 & 79.6 & 4.5 & 69.8 & 81.3 & 14.6 \\
\bottomrule
\end{tabular*}
\end{table*}

%% file: figures/supp_creation.tex
\begin{figure*}[t]
\centering\scriptsize
\begin{minipage}[t]{0.243\textwidth}\raggedright{\color[HTML]{7950B5}\textbf{1. Title and chapter map}}\hfill{\color{black!55}steps 1--8}\\[1pt]\includegraphics[width=0.494\linewidth]{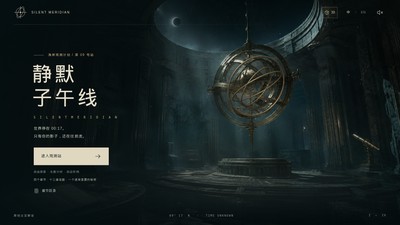}\hfill\includegraphics[width=0.494\linewidth]{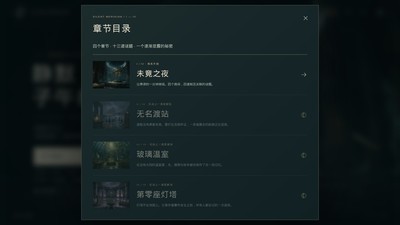}\\[-0.6pt]\includegraphics[width=0.494\linewidth]{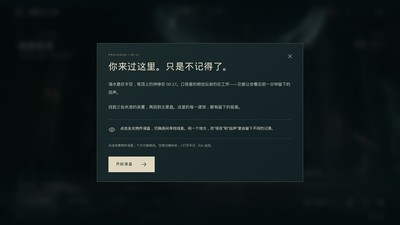}\hfill\includegraphics[width=0.494\linewidth]{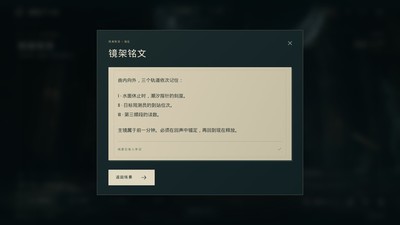}\\[1pt]{\color{black!55}\textit{Reference.}} Reads the title, the chapter map of four chapters and 13 puzzles, the prologue, and the first clue.\end{minipage}\hspace*{\fill}\begin{minipage}[t]{0.243\textwidth}\raggedright{\color[HTML]{7950B5}\textbf{2. Chapter I rooms}}\hfill{\color{black!55}steps 9--108}\\[1pt]\includegraphics[width=0.494\linewidth]{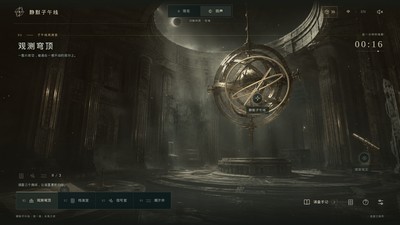}\hfill\includegraphics[width=0.494\linewidth]{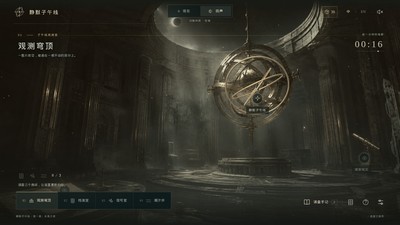}\\[-0.6pt]\includegraphics[width=0.494\linewidth]{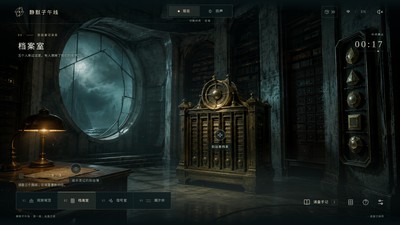}\hfill\includegraphics[width=0.494\linewidth]{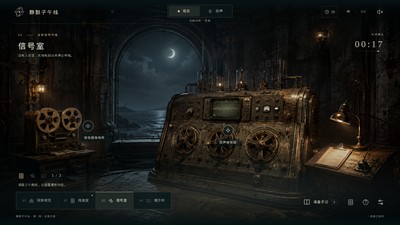}\\[1pt]{\color{black!55}\textit{Reference.}} Visits each room in the present and in the echo view and logs every clue in a notes file.\end{minipage}\hspace*{\fill}\begin{minipage}[t]{0.243\textwidth}\raggedright{\color[HTML]{7950B5}\textbf{3. Solving puzzles}}\hfill{\color{black!55}steps 41--141}\\[1pt]\includegraphics[width=0.494\linewidth]{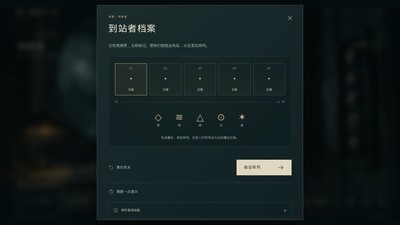}\hfill\includegraphics[width=0.494\linewidth]{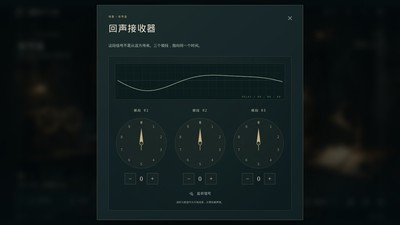}\\[-0.6pt]\includegraphics[width=0.494\linewidth]{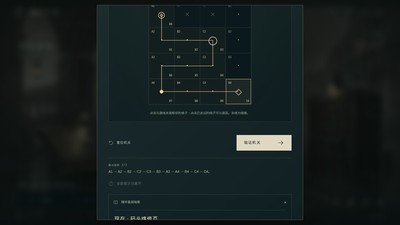}\hfill\includegraphics[width=0.494\linewidth]{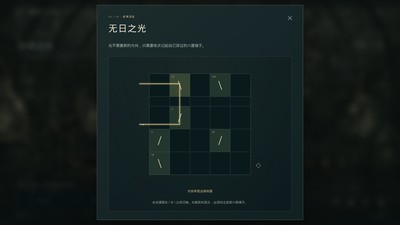}\\[1pt]{\color{black!55}\textit{Reference.}} Solves each puzzle from its clues and hints, writes search programs over the rules, and records the answers.\end{minipage}\hspace*{\fill}\begin{minipage}[t]{0.243\textwidth}\raggedright{\color[HTML]{7950B5}\textbf{4. Chapters II to IV}}\hfill{\color{black!55}steps 109--227}\\[1pt]\includegraphics[width=0.494\linewidth]{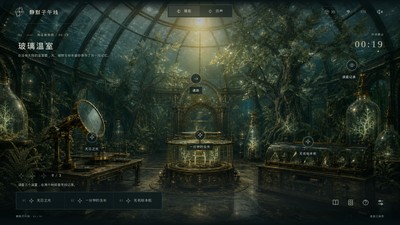}\hfill\includegraphics[width=0.494\linewidth]{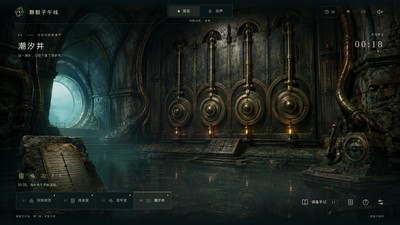}\\[-0.6pt]\includegraphics[width=0.494\linewidth]{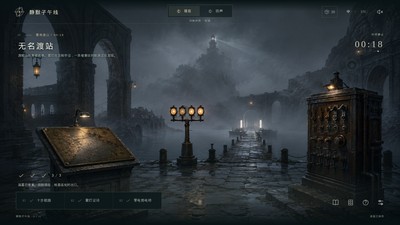}\hfill\includegraphics[width=0.494\linewidth]{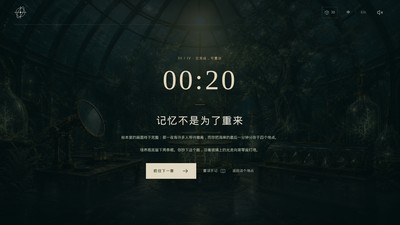}\\[1pt]{\color{black!55}\textit{Reference.}} Plays the remaining chapters, both choices, and every ending.\end{minipage}\\[5pt]
\begin{minipage}[t]{0.243\textwidth}\raggedright{\color[HTML]{7950B5}\textbf{5. Clean backgrounds}}\hfill{\color{black!55}steps 228--314}\\[1pt]\includegraphics[width=0.494\linewidth]{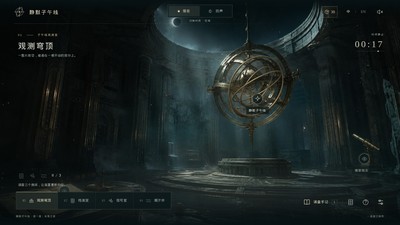}\hfill\includegraphics[width=0.494\linewidth]{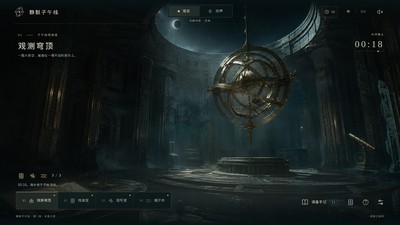}\\[-0.6pt]\includegraphics[width=0.494\linewidth]{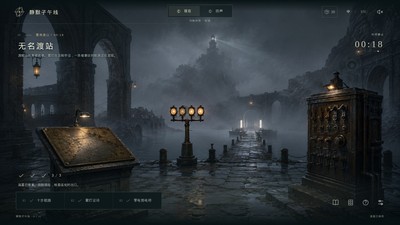}\hfill\includegraphics[width=0.494\linewidth]{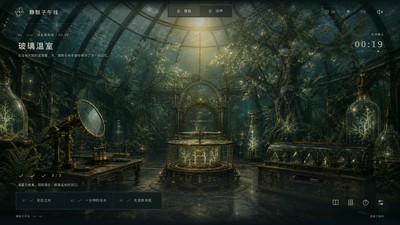}\\[1pt]{\color{black!55}\textit{Reference capture.}} Finds that the echo view is the same image under a filter and recaptures each room with the interface hidden.\end{minipage}\hspace*{\fill}\begin{minipage}[t]{0.243\textwidth}\raggedright{\color[HTML]{458D85}\textbf{6. Writing the project}}\hfill{\color{black!55}steps 315--339}\\[1pt]\includegraphics[width=0.494\linewidth]{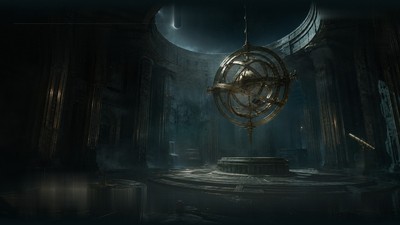}\hfill\includegraphics[width=0.494\linewidth]{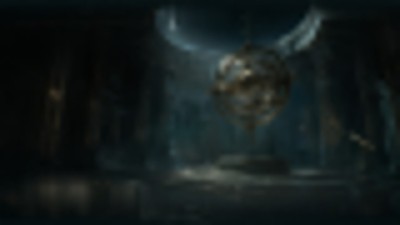}\\[-0.6pt]\includegraphics[width=0.494\linewidth]{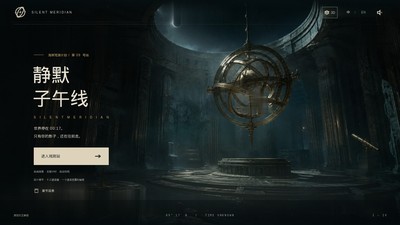}\hfill\includegraphics[width=0.494\linewidth]{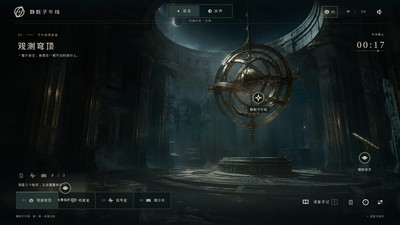}\\[1pt]{\color{black!55}\textit{Recreation.}} Turns the captures into backgrounds and writes seven files, from content and UI to 13 puzzle widgets.\end{minipage}\hspace*{\fill}\begin{minipage}[t]{0.243\textwidth}\raggedright{\color[HTML]{458D85}\textbf{7. Testing the build}}\hfill{\color{black!55}steps 340--430}\\[1pt]\includegraphics[width=0.494\linewidth]{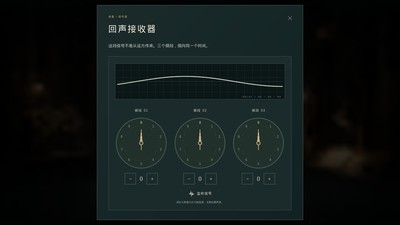}\hfill\includegraphics[width=0.494\linewidth]{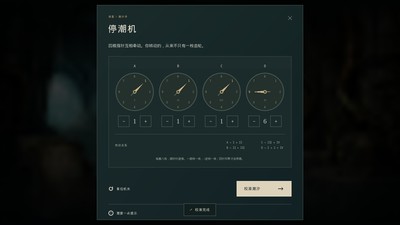}\\[-0.6pt]\includegraphics[width=0.494\linewidth]{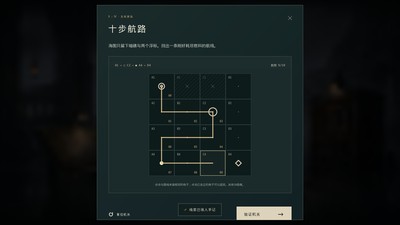}\hfill\includegraphics[width=0.494\linewidth]{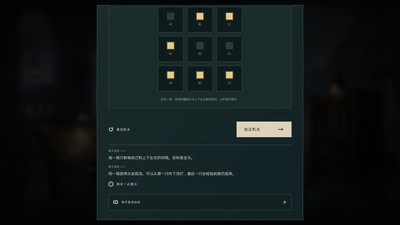}\\[1pt]{\color{black!55}\textit{Recreation.}} Makes 60 calls on its own build, checking each puzzle panel against the saved reference captures.\end{minipage}\hspace*{\fill}\begin{minipage}[t]{0.243\textwidth}\raggedright{\color[HTML]{458D85}\textbf{8. Delivery}}\hfill{\color{black!55}steps 431--439}\\[1pt]\includegraphics[width=0.494\linewidth]{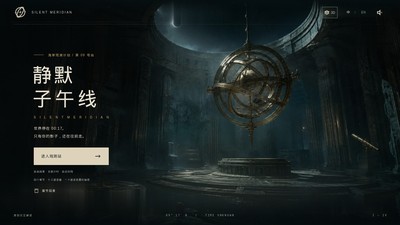}\hfill\includegraphics[width=0.494\linewidth]{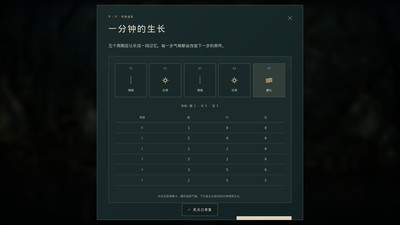}\\[-0.6pt]\includegraphics[width=0.494\linewidth]{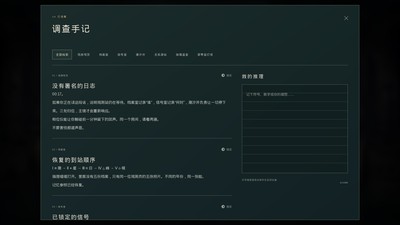}\hfill\includegraphics[width=0.494\linewidth]{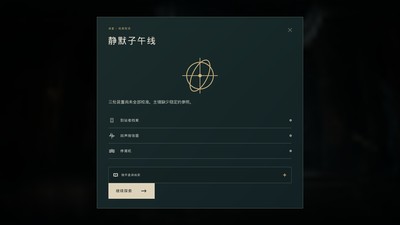}\\[1pt]{\color{black!55}\textit{Recreation.}} Checks the remaining screens, replays two demonstration traces, and writes the run instructions.\end{minipage}\\[5pt]
\includegraphics[width=\textwidth]{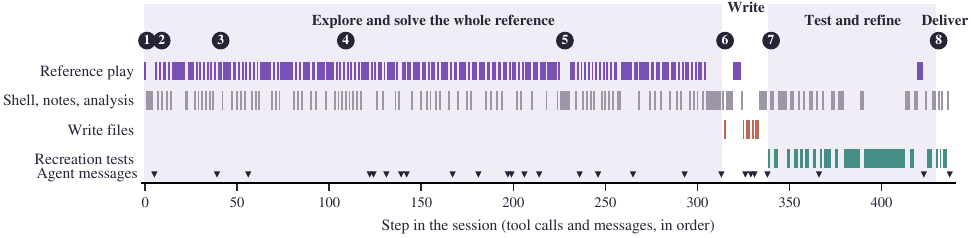}
\caption{\textbf{The full creation session of Claude Opus 5.5 on Silent Meridian, which scores 98.65.} Panels 1--8 summarize the session in order, with the steps each one covers. Purple panels show the reference as the agent saw it, and green panels show its recreation. The lane plot below places all 438 steps of the session, every tool call and message, on one axis, with the numbered markers giving where each panel begins. The agent spends the first 314 steps on the reference before writing any game code.}
\label{fig:supp-creation}
\end{figure*}

%% file: figures/supp_trajectories.tex
\begin{figure*}[t]
\centering\scriptsize
\scorelegend\\[3pt]
\setlength{\tabcolsep}{0.5pt}
\renewcommand{\arraystretch}{0.9}
\begin{tabular}{@{}rcccccc@{}}
 & \parbox[b]{0.158\textwidth}{\raggedright\textbf{1. Title screen}\\{\color{black!70}Bilingual slanted title and a green 3D start button. (1\,pt)}} & \parbox[b]{0.158\textwidth}{\raggedright\textbf{2. Level select}\\{\color{black!70}Two rows of twelve level cards with star slots. (1\,pt)}} & \parbox[b]{0.158\textwidth}{\raggedright\textbf{3. Level 1 layout}\\{\color{black!70}Level 1 has separate low frames and a stacked roof. (0.5\,pt)}} & \parbox[b]{0.158\textwidth}{\raggedright\textbf{4. Aiming}\\{\color{black!70}Pulling the bird shows a curved dotted aim line. (1.25\,pt)}} & \parbox[b]{0.158\textwidth}{\raggedright\textbf{5. Structure collapses}\\{\color{black!70}Loads fall once their support is destroyed. (2.5\,pt)}} & \parbox[b]{0.158\textwidth}{\raggedright\textbf{6. Result panel}\\{\color{black!70}Result panel with three stars, the score, and round buttons. (1\,pt)}} \\[2pt]
\rotatebox{90}{\parbox{0.0889\textwidth}{\centering\textbf{Reference}}} & \plainimg{0.158\textwidth}{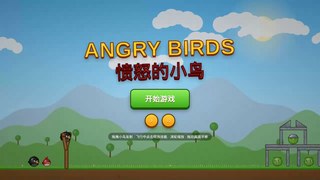} & \plainimg{0.158\textwidth}{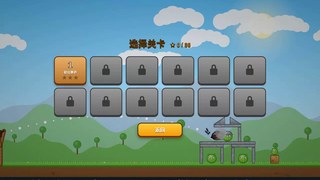} & \plainimg{0.158\textwidth}{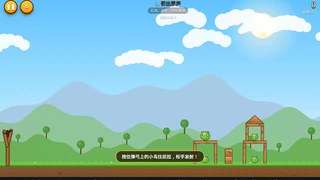} & \plainimg{0.158\textwidth}{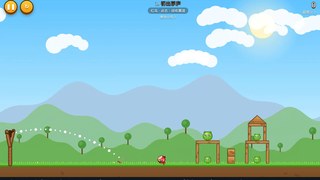} & \plainimg{0.158\textwidth}{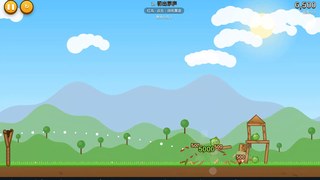} & \plainimg{0.158\textwidth}{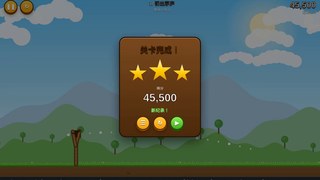} \\[4pt]
\rotatebox{90}{\parbox{0.0889\textwidth}{\centering\textbf{GPT-6-Astra}}} & \scoreimg{sfull}{0.158\textwidth}{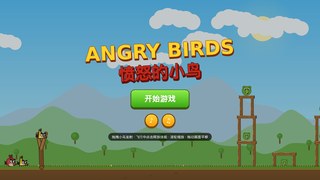} & \scoreimg{sfull}{0.158\textwidth}{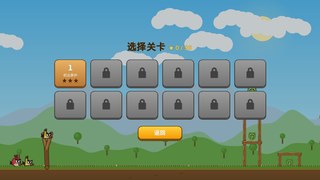} & \scoreimg{sfull}{0.158\textwidth}{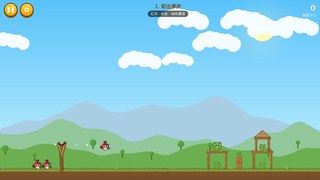} & \scoreimg{sfull}{0.158\textwidth}{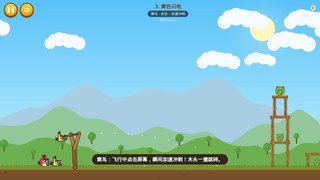} & \scoreimg{sfull}{0.158\textwidth}{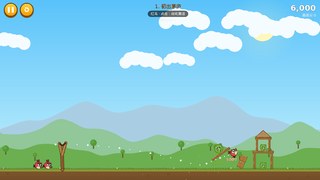} & \scoreimg{sfull}{0.158\textwidth}{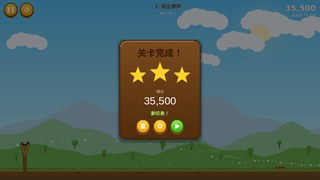} \\[0.5pt]
\rotatebox{90}{\parbox{0.0889\textwidth}{\centering\textbf{Claude Opus 5.5}}} & \scoreimg{sfull}{0.158\textwidth}{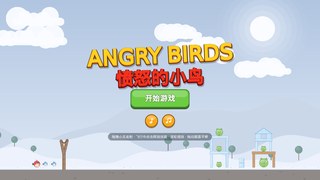} & \scoreimg{sfull}{0.158\textwidth}{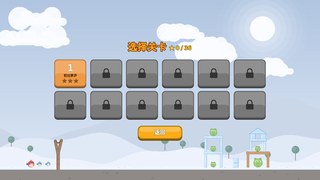} & \scoreimg{sfull}{0.158\textwidth}{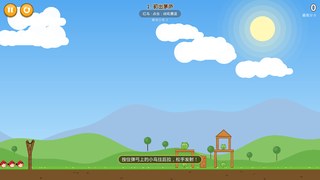} & \scoreimg{sfull}{0.158\textwidth}{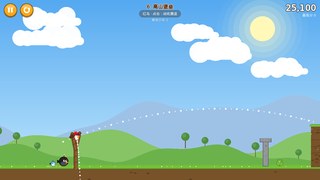} & \scoreimg{sfull}{0.158\textwidth}{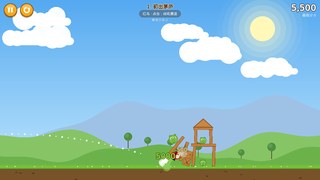} & \scoreimg{sfull}{0.158\textwidth}{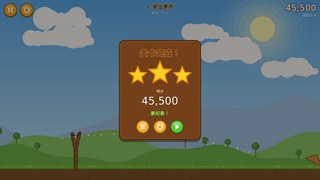} \\[0.5pt]
\rotatebox{90}{\parbox{0.0889\textwidth}{\centering\textbf{Gemini 4 Argon}}} & \scoreimg{spart}{0.158\textwidth}{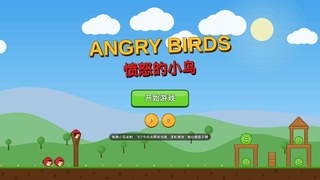} & \scoreimg{spart}{0.158\textwidth}{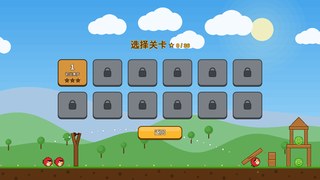} & \scoreimg{sfull}{0.158\textwidth}{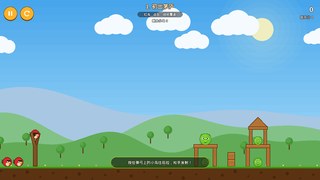} & \scoreimg{spart}{0.158\textwidth}{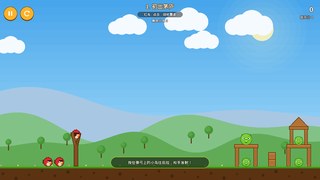} & \scoreimg{sfull}{0.158\textwidth}{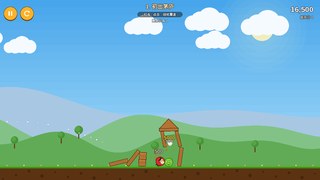} & \scoreimg{spart}{0.158\textwidth}{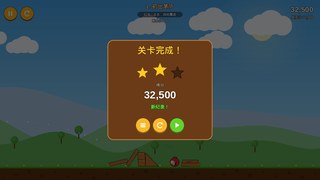} \\[0.5pt]
\rotatebox{90}{\parbox{0.0889\textwidth}{\centering\textbf{GPT-6.1-Sol}}} & \scoreimg{spart}{0.158\textwidth}{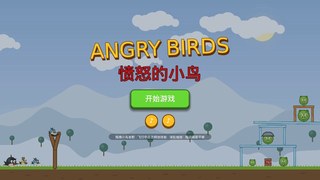} & \scoreimg{sfull}{0.158\textwidth}{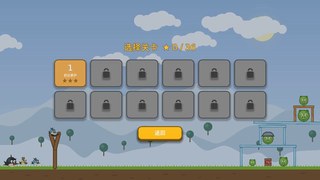} & \scoreimg{sfull}{0.158\textwidth}{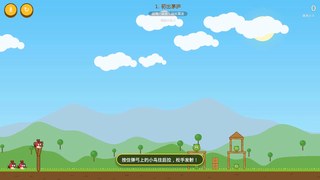} & \scoreimg{sfull}{0.158\textwidth}{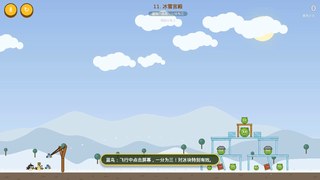} & \scoreimg{sfull}{0.158\textwidth}{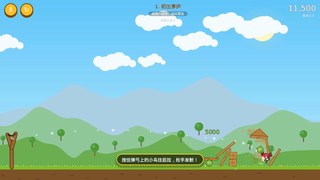} & \scoreimg{spart}{0.158\textwidth}{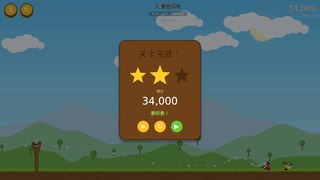} \\[0.5pt]
\rotatebox{90}{\parbox{0.0889\textwidth}{\centering\textbf{DeepSeek V4.1}}} & \scoreimg{spart}{0.158\textwidth}{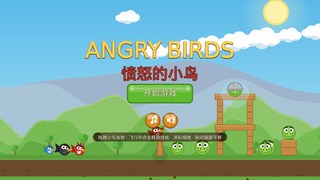} & \scoreimg{sfull}{0.158\textwidth}{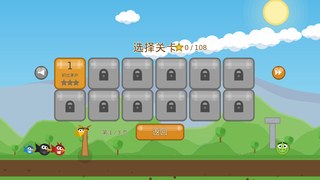} & \scoreimg{spart}{0.158\textwidth}{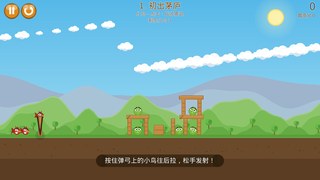} & \noshot{szero}{0.158\textwidth}{0.0889\textwidth} & \noshot{szero}{0.158\textwidth}{0.0889\textwidth} & \noshot{szero}{0.158\textwidth}{0.0889\textwidth} \\[0.5pt]
\rotatebox{90}{\parbox{0.0889\textwidth}{\centering\textbf{GLM-5.3-Flash}}} & \scoreimg{spart}{0.158\textwidth}{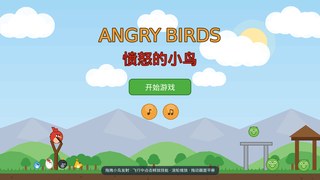} & \scoreimg{spart}{0.158\textwidth}{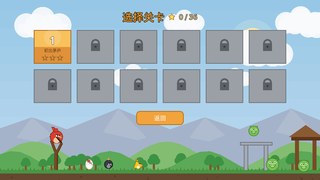} & \scoreimg{spart}{0.158\textwidth}{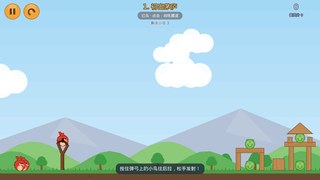} & \scoreimg{spart}{0.158\textwidth}{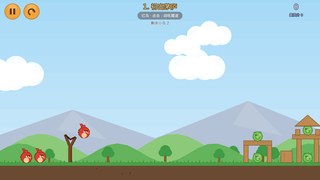} & \scoreimg{sfull}{0.158\textwidth}{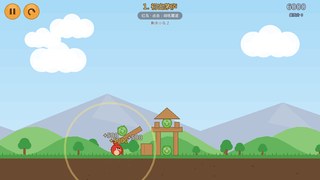} & \scoreimg{spart}{0.158\textwidth}{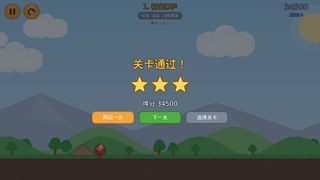} \\[0.5pt]
\rotatebox{90}{\parbox{0.0889\textwidth}{\centering\textbf{Qwen3.8-27B}}} & \scoreimg{spart}{0.158\textwidth}{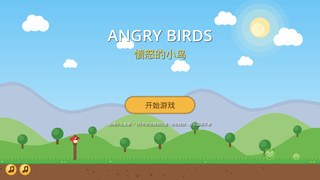} & \scoreimg{szero}{0.158\textwidth}{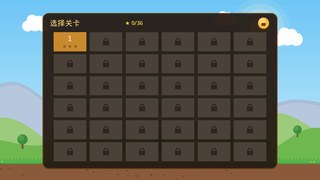} & \scoreimg{spart}{0.158\textwidth}{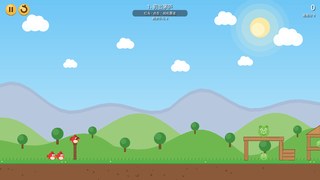} & \noshot{szero}{0.158\textwidth}{0.0889\textwidth} & \noshot{szero}{0.158\textwidth}{0.0889\textwidth} & \noshot{szero}{0.158\textwidth}{0.0889\textwidth} \\[0.5pt]
\end{tabular}
\caption{\textbf{Six moments of one scenario of Angry Birds HTML5, for all seven systems.} The scenario is the first one, from the title screen through level 1 to its result, and columns follow the order of play. Each column is one criterion, stated above it with its points. The reference frame comes from the reference recording at the evidence cited in the rubric, and each recreation frame from the replay interval the judge cited for that criterion. The border color gives the points each system earns, and a gray cell means the judge found no footage of that moment.}
\label{fig:supp-traj-birds}
\end{figure*}

\begin{figure*}[t]
\centering\scriptsize
\scorelegend\\[3pt]
\setlength{\tabcolsep}{0.5pt}
\renewcommand{\arraystretch}{0.9}
\begin{tabular}{@{}rcccccc@{}}
 & \parbox[b]{0.158\textwidth}{\raggedright\textbf{1. Start with AI karts}\\{\color{black!70}AI karts drive along the track on their own. (2\,pt)}} & \parbox[b]{0.158\textwidth}{\raggedright\textbf{2. Manual rescue}\\{\color{black!70}One short press starts the rescue lift. (1.5\,pt)}} & \parbox[b]{0.158\textwidth}{\raggedright\textbf{3. Pause menu}\\{\color{black!70}Pause menu is a large light panel over the dimmed race. (1\,pt)}} & \parbox[b]{0.158\textwidth}{\raggedright\textbf{4. Kart close-up}\\{\color{black!70}The kart has the low, open-wheeled silhouette. (3\,pt)}} & \parbox[b]{0.158\textwidth}{\raggedright\textbf{5. Castle in town}\\{\color{black!70}The castle keeps its stone towers, battlements, and arch. (2.5\,pt)}} & \parbox[b]{0.158\textwidth}{\raggedright\textbf{6. Stone wall}\\{\color{black!70}Stone town walls stop the kart. (2\,pt)}} \\[2pt]
\rotatebox{90}{\parbox{0.0889\textwidth}{\centering\textbf{Reference}}} & \plainimg{0.158\textwidth}{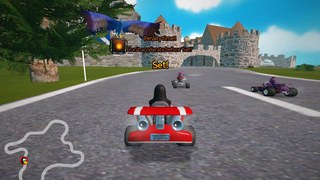} & \plainimg{0.158\textwidth}{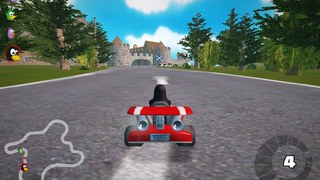} & \plainimg{0.158\textwidth}{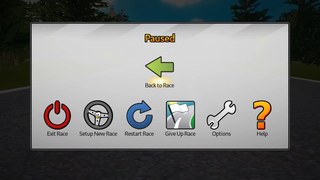} & \plainimg{0.158\textwidth}{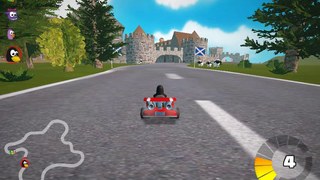} & \plainimg{0.158\textwidth}{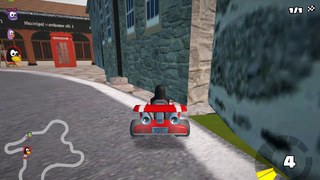} & \plainimg{0.158\textwidth}{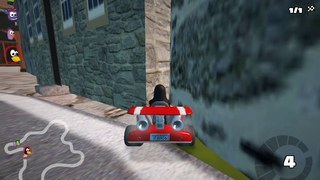} \\[4pt]
\rotatebox{90}{\parbox{0.0889\textwidth}{\centering\textbf{GPT-6-Astra}}} & \scoreimg{sfull}{0.158\textwidth}{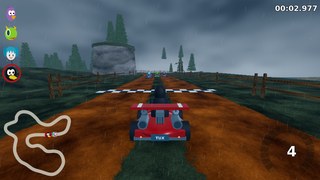} & \noshot{szero}{0.158\textwidth}{0.0889\textwidth} & \scoreimg{sfull}{0.158\textwidth}{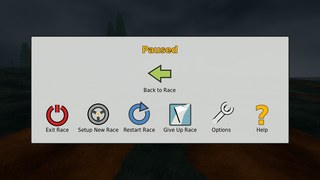} & \scoreimg{spart}{0.158\textwidth}{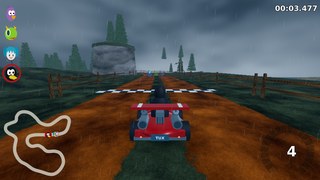} & \noshot{szero}{0.158\textwidth}{0.0889\textwidth} & \noshot{szero}{0.158\textwidth}{0.0889\textwidth} \\[0.5pt]
\rotatebox{90}{\parbox{0.0889\textwidth}{\centering\textbf{Claude Opus 5.5}}} & \scoreimg{sfull}{0.158\textwidth}{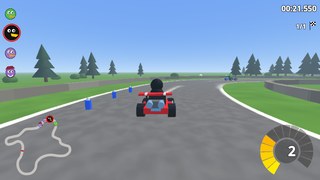} & \scoreimg{sfull}{0.158\textwidth}{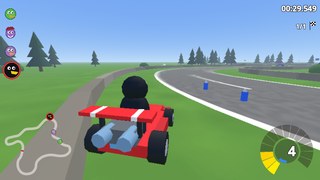} & \scoreimg{sfull}{0.158\textwidth}{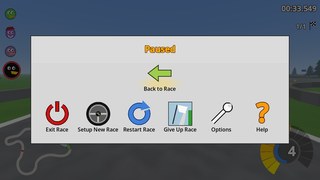} & \scoreimg{spart}{0.158\textwidth}{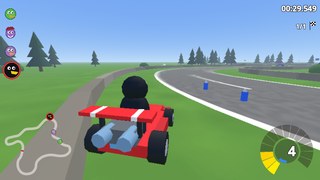} & \scoreimg{szero}{0.158\textwidth}{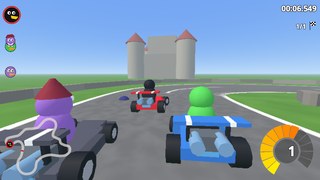} & \scoreimg{szero}{0.158\textwidth}{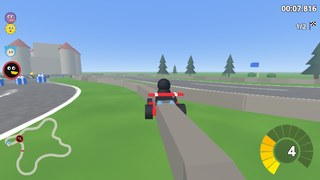} \\[0.5pt]
\rotatebox{90}{\parbox{0.0889\textwidth}{\centering\textbf{Gemini 4 Argon}}} & \scoreimg{sfull}{0.158\textwidth}{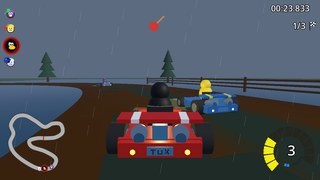} & \scoreimg{sfull}{0.158\textwidth}{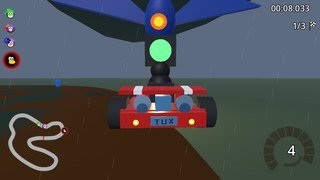} & \scoreimg{sfull}{0.158\textwidth}{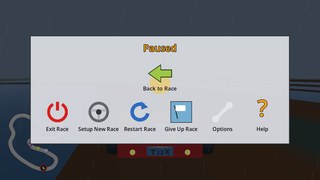} & \scoreimg{spart}{0.158\textwidth}{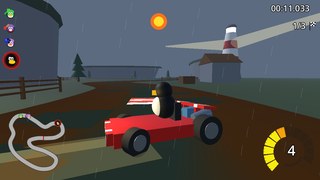} & \noshot{szero}{0.158\textwidth}{0.0889\textwidth} & \noshot{szero}{0.158\textwidth}{0.0889\textwidth} \\[0.5pt]
\rotatebox{90}{\parbox{0.0889\textwidth}{\centering\textbf{GPT-6.1-Sol}}} & \scoreimg{sfull}{0.158\textwidth}{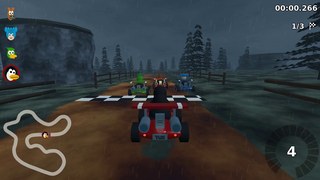} & \scoreimg{sfull}{0.158\textwidth}{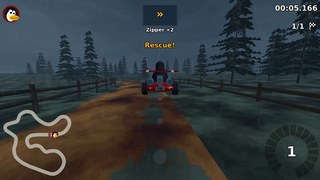} & \scoreimg{sfull}{0.158\textwidth}{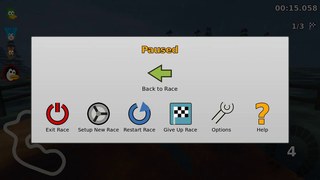} & \scoreimg{spart}{0.158\textwidth}{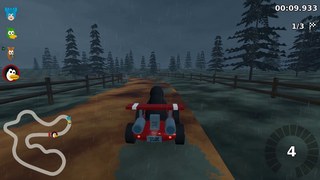} & \noshot{szero}{0.158\textwidth}{0.0889\textwidth} & \noshot{szero}{0.158\textwidth}{0.0889\textwidth} \\[0.5pt]
\rotatebox{90}{\parbox{0.0889\textwidth}{\centering\textbf{DeepSeek V4.1}}} & \scoreimg{sfull}{0.158\textwidth}{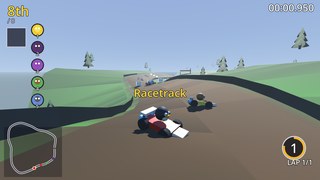} & \scoreimg{szero}{0.158\textwidth}{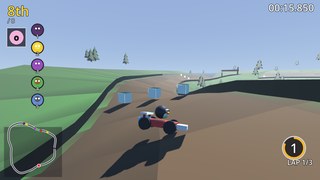} & \scoreimg{spart}{0.158\textwidth}{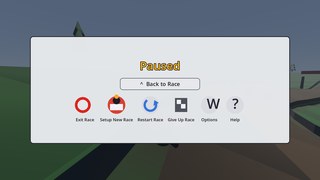} & \scoreimg{spart}{0.158\textwidth}{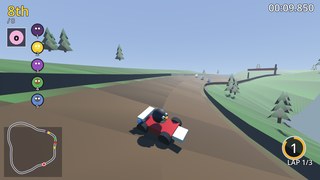} & \noshot{szero}{0.158\textwidth}{0.0889\textwidth} & \noshot{szero}{0.158\textwidth}{0.0889\textwidth} \\[0.5pt]
\rotatebox{90}{\parbox{0.0889\textwidth}{\centering\textbf{GLM-5.3-Flash}}} & \scoreimg{sfull}{0.158\textwidth}{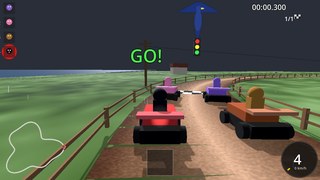} & \scoreimg{szero}{0.158\textwidth}{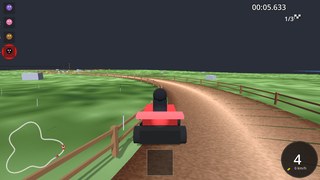} & \scoreimg{spart}{0.158\textwidth}{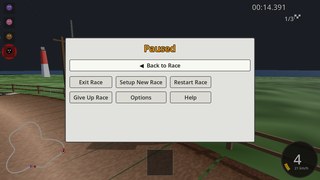} & \scoreimg{spart}{0.158\textwidth}{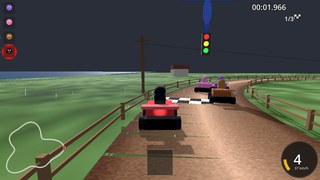} & \noshot{szero}{0.158\textwidth}{0.0889\textwidth} & \noshot{szero}{0.158\textwidth}{0.0889\textwidth} \\[0.5pt]
\rotatebox{90}{\parbox{0.0889\textwidth}{\centering\textbf{Qwen3.8-27B}}} & \scoreimg{sfull}{0.158\textwidth}{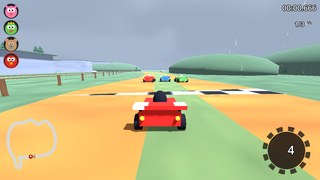} & \scoreimg{szero}{0.158\textwidth}{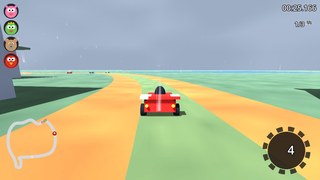} & \scoreimg{szero}{0.158\textwidth}{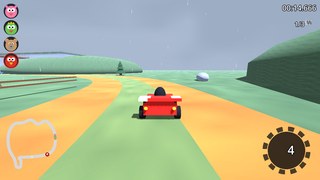} & \scoreimg{szero}{0.158\textwidth}{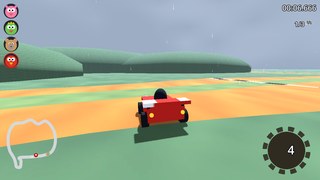} & \noshot{szero}{0.158\textwidth}{0.0889\textwidth} & \noshot{szero}{0.158\textwidth}{0.0889\textwidth} \\[0.5pt]
\end{tabular}
\caption{\textbf{Six moments of one scenario of SuperTuxKart, for all seven systems.} The scenario is a normal race on the town track, and columns follow the order of play. Each column is one criterion, stated above it with its points. The reference frame comes from the reference recording at the evidence cited in the rubric, and each recreation frame from the replay interval the judge cited for that criterion. The border color gives the points each system earns, and a gray cell means the judge found no footage of that moment.}
\label{fig:supp-traj-kart}
\end{figure*}

%% file: figures/supp_hextris.tex
\begin{figure*}[t]
\centering\footnotesize
\setlength{\tabcolsep}{2pt}
\begin{tabular}{@{}rcccc@{}}
 & \textit{First game} & \textit{Pause screen} & \textit{Middle of a game} & \textit{Late game with a combo} \\
\rotatebox{90}{\hspace{0.6em}Reference} & \includegraphics[width=0.235\textwidth]{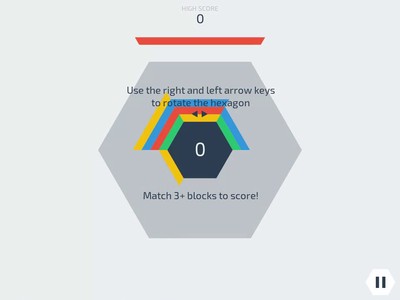} & \includegraphics[width=0.235\textwidth]{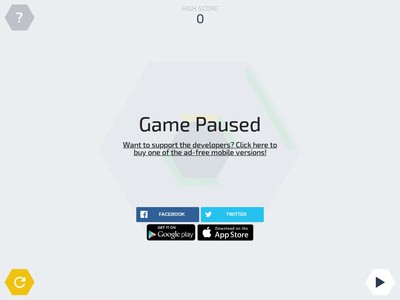} & \includegraphics[width=0.235\textwidth]{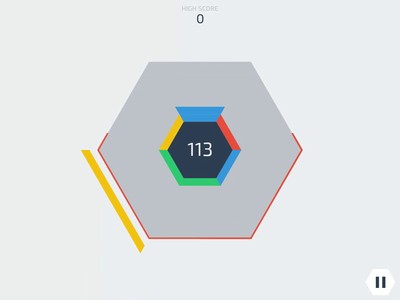} & \includegraphics[width=0.235\textwidth]{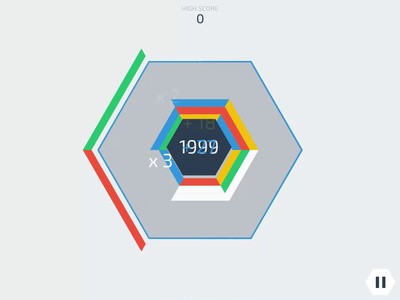} \\
\rotatebox{90}{\hspace{0.6em}Recreation} & \includegraphics[width=0.235\textwidth]{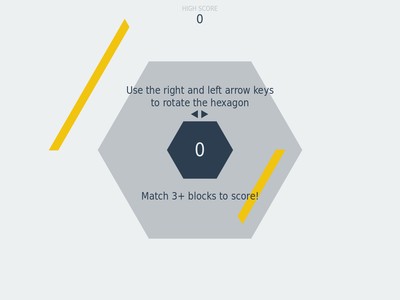} & \includegraphics[width=0.235\textwidth]{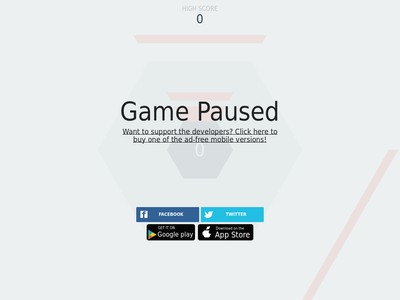} & \includegraphics[width=0.235\textwidth]{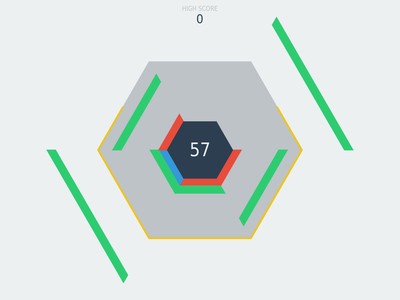} & \includegraphics[width=0.235\textwidth]{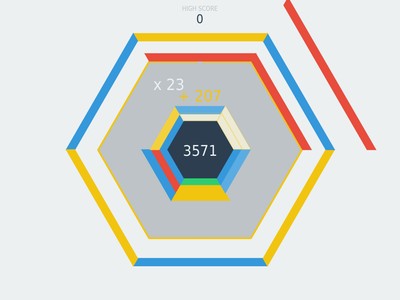} \\
\end{tabular}
\caption{\textbf{Hextris and its recreation by GPT-6-Astra, which scores 87.0.} Reference frames come from the reference recordings and recreation frames from the judged demonstration replays, at matching moments of play rather than identical game states.}
\label{fig:supp-hextris}
\end{figure*}

%% file: figures/supp_beyond.tex
\begin{figure*}[!t]
\centering\footnotesize
\setlength{\tabcolsep}{1.5pt}
\renewcommand{\arraystretch}{1.0}
\begin{tabular}{@{}>{\raggedright\arraybackslash}m{0.12\textwidth}>{\centering\arraybackslash}m{0.195\textwidth}>{\centering\arraybackslash}m{0.195\textwidth}>{\centering\arraybackslash}m{0.195\textwidth}>{\centering\arraybackslash}m{0.195\textwidth}@{}}
 & \textbf{Entry scenario} & \multicolumn{3}{c}{\textbf{Later scenarios}} \\
\cmidrule(lr){2-2}\cmidrule(l){3-5}
 & \textbf{Practice lap, automatic gearbox} & \textbf{Practice lap, sequential gearbox} & \textbf{Quick race against two AI drivers} & \textbf{Challenge with qualifying} \\[2pt]
\textbf{Reference} & \plainimg{0.195\textwidth}{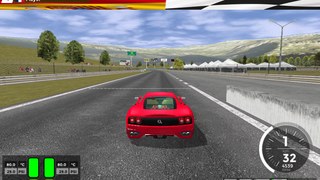} & \plainimg{0.195\textwidth}{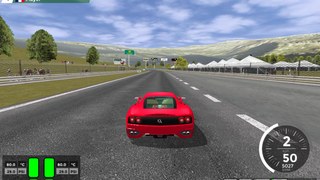} & \plainimg{0.195\textwidth}{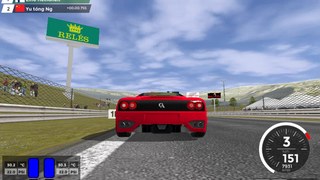} & \plainimg{0.195\textwidth}{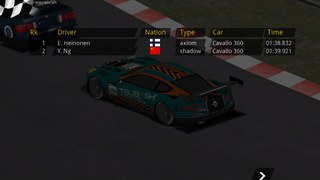} \\[1pt]
\textbf{GPT-6-Astra} & \plainimg{0.195\textwidth}{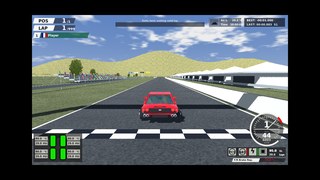} & \plainimg{0.195\textwidth}{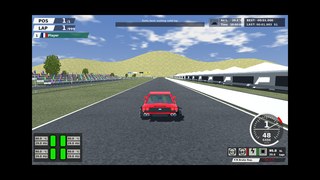} & \plainimg{0.195\textwidth}{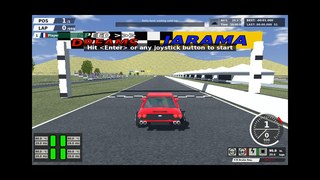} & {\setlength{\fboxsep}{0pt}\setlength{\fboxrule}{1.4pt}\fcolorbox{white}{black!6}{\parbox[b][\dimexpr0.1097\textwidth-2.8pt\relax][c]{\dimexpr0.195\textwidth-2.8pt\relax}{\centering\color{black!55}not reached}}} \\[6pt]
\textit{Points earned} & \textit{28 criteria} & \textit{8 criteria} & \textit{26 criteria} & \textit{16 criteria} \\
\midrule
\rule{0pt}{9pt}GPT-6-Astra & \cellcolor{p2rpurple!56}\textcolor{white}{93\%} & \cellcolor{p2rpurple!33}55\% & \cellcolor{p2rpurple!14}24\% & \cellcolor{p2rpurple!0}0\% \\
\rule{0pt}{9pt}Claude Opus 5.5 & \cellcolor{p2rpurple!49}\textcolor{white}{81\%} & \cellcolor{p2rpurple!11}19\% & \cellcolor{p2rpurple!18}31\% & \cellcolor{p2rpurple!0}0\% \\
\rule{0pt}{9pt}Gemini 4 Argon & \cellcolor{p2rpurple!43}72\% & \cellcolor{p2rpurple!11}19\% & \cellcolor{p2rpurple!28}47\% & \cellcolor{p2rpurple!0}0\% \\
\rule{0pt}{9pt}GPT-6.1-Sol & \cellcolor{p2rpurple!41}68\% & \cellcolor{p2rpurple!11}19\% & \cellcolor{p2rpurple!13}22\% & \cellcolor{p2rpurple!0}0\% \\
\rule{0pt}{9pt}DeepSeek V4.1 & \cellcolor{p2rpurple!27}46\% & \cellcolor{p2rpurple!11}19\% & \cellcolor{p2rpurple!10}17\% & \cellcolor{p2rpurple!0}0\% \\
\rule{0pt}{9pt}GLM-5.3-Flash & \cellcolor{p2rpurple!34}57\% & \cellcolor{p2rpurple!0}0\% & \cellcolor{p2rpurple!7}11\% & \cellcolor{p2rpurple!0}0\% \\
\rule{0pt}{9pt}Qwen3.8-27B & \cellcolor{p2rpurple!34}57\% & \cellcolor{p2rpurple!0}0\% & \cellcolor{p2rpurple!8}13\% & \cellcolor{p2rpurple!0}0\% \\
\bottomrule
\end{tabular}
\caption{\textbf{Recreations of Speed Dreams fall off after the entry scenario.} Top, one frame of each scenario in the reference and in the recreation by GPT-6-Astra, taken from the reference recordings and from its judged replays. Bottom, the share of points each system earns on the criteria first shown in each scenario, with darker cells for higher shares. All systems do best on the entry practice lap, and none earns any points on the Challenge mode, which GPT-6-Astra never reaches.}
\label{fig:supp-beyond}
\end{figure*}

%% file: figures/supp_visual_examples.tex

\begin{figure*}[t]
\centering\scriptsize
\scorelegend\\[3pt]
\setlength{\tabcolsep}{0.5pt}
\renewcommand{\arraystretch}{0.9}
\begin{tabular}{@{}ccccccccc@{}}
 & \textbf{Reference} & \textbf{GPT-6-Astra} & \textbf{Claude Opus 5.5} & \textbf{Gemini 4 Argon} & \textbf{GPT-6.1-Sol} & \textbf{DeepSeek V4.1} & \textbf{GLM-5.3-Flash} & \textbf{Qwen3.8-27B} \\[1pt]
\multicolumn{9}{@{}l}{\textbf{2048}\enspace {\color{black!60}visual fidelity, 4\,pt}\enspace The title, score panel, and controls sit above the board in the original vertical layout.} \\[0.5pt]
 & \plainimg{0.1185\textwidth}{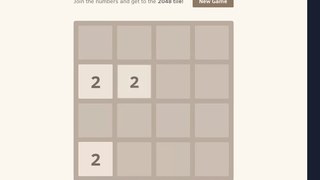} & \scoreimg{sfull}{0.1185\textwidth}{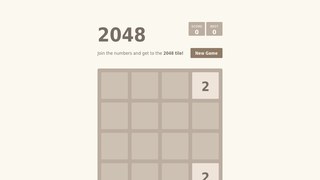} & \scoreimg{sfull}{0.1185\textwidth}{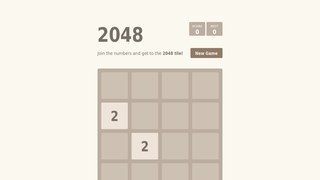} & \scoreimg{sfull}{0.1185\textwidth}{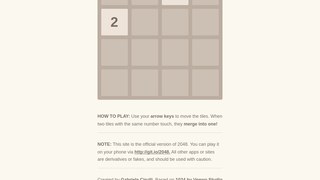} & \scoreimg{sfull}{0.1185\textwidth}{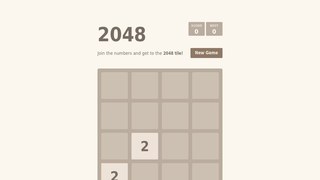} & \scoreimg{spart}{0.1185\textwidth}{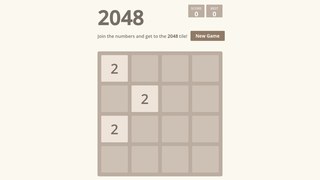} & \scoreimg{spart}{0.1185\textwidth}{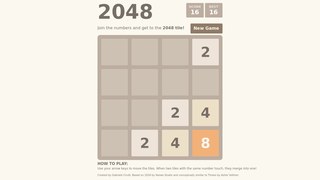} & \scoreimg{sfull}{0.1185\textwidth}{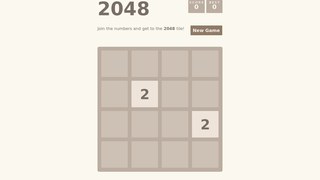} \\[3pt]
\multicolumn{9}{@{}l}{\textbf{Courtyard Defense}\enspace {\color{black!60}visual fidelity, 3.5\,pt}\enspace The yard surfaces have illustrated textures and painted shading.} \\[0.5pt]
 & \plainimg{0.1185\textwidth}{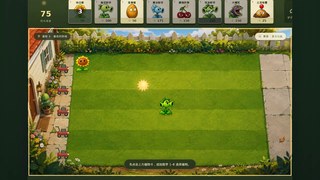} & \scoreimg{sfull}{0.1185\textwidth}{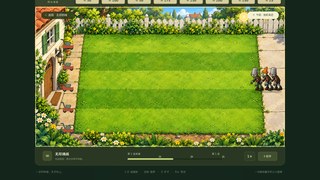} & \scoreimg{spart}{0.1185\textwidth}{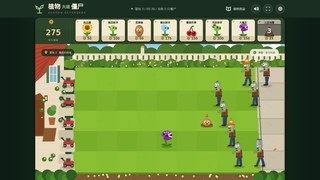} & \scoreimg{sfull}{0.1185\textwidth}{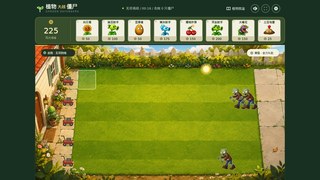} & \scoreimg{sfull}{0.1185\textwidth}{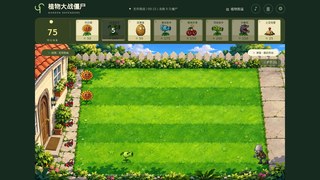} & \scoreimg{spart}{0.1185\textwidth}{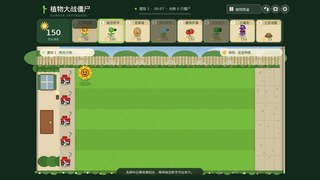} & \scoreimg{szero}{0.1185\textwidth}{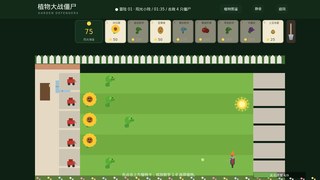} & \scoreimg{szero}{0.1185\textwidth}{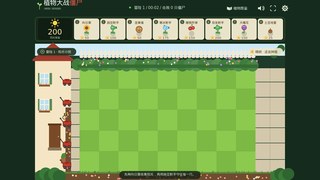} \\[3pt]
\multicolumn{9}{@{}l}{\textbf{Angry Birds HTML5}\enspace {\color{black!60}visual fidelity, 2\,pt}\enspace Wooden beams are yellow-brown, with grain, dark edges, and lit and shaded faces.} \\[0.5pt]
 & \plainimg{0.1185\textwidth}{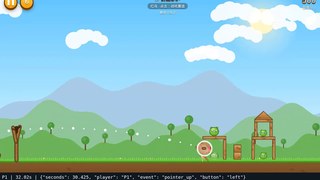} & \scoreimg{sfull}{0.1185\textwidth}{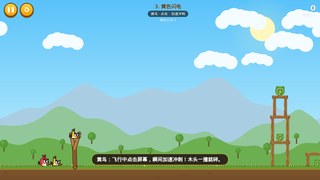} & \scoreimg{spart}{0.1185\textwidth}{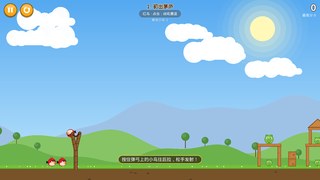} & \scoreimg{szero}{0.1185\textwidth}{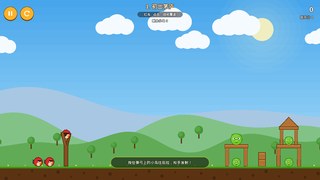} & \scoreimg{spart}{0.1185\textwidth}{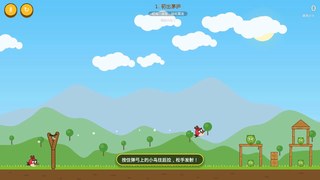} & \scoreimg{szero}{0.1185\textwidth}{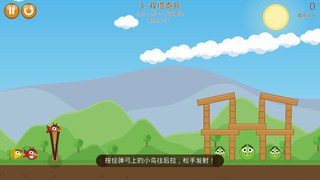} & \scoreimg{spart}{0.1185\textwidth}{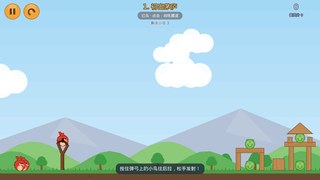} & \scoreimg{spart}{0.1185\textwidth}{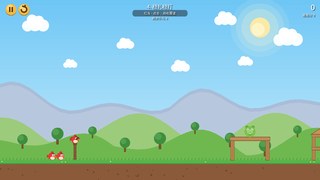} \\[3pt]
\multicolumn{9}{@{}l}{\textbf{SuperTux}\enspace {\color{black!60}visual fidelity, 2\,pt}\enspace The snowy mountain backdrop has soft, pale blue and white shading.} \\[0.5pt]
 & \plainimg{0.1185\textwidth}{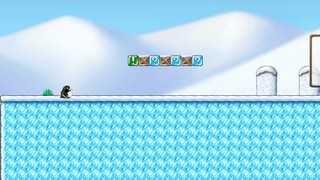} & \scoreimg{sfull}{0.1185\textwidth}{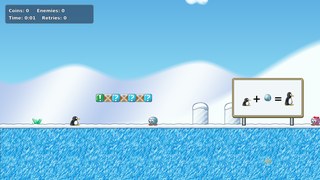} & \scoreimg{spart}{0.1185\textwidth}{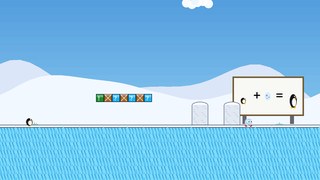} & \scoreimg{sfull}{0.1185\textwidth}{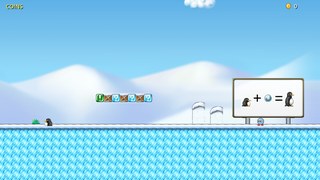} & \scoreimg{spart}{0.1185\textwidth}{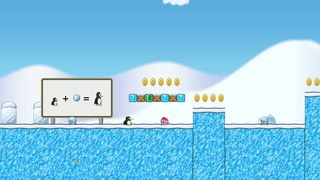} & \scoreimg{spart}{0.1185\textwidth}{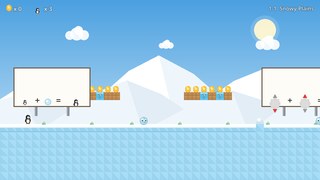} & \scoreimg{spart}{0.1185\textwidth}{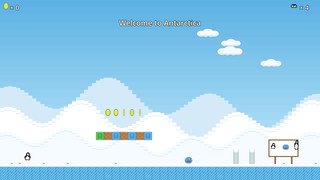} & \scoreimg{szero}{0.1185\textwidth}{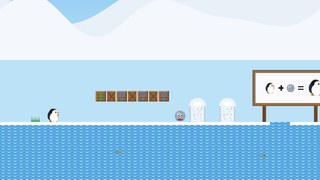} \\[3pt]
\multicolumn{9}{@{}l}{\textbf{Menghuan Xiyou Q}\enspace {\color{black!60}visual fidelity, 2\,pt}\enspace Stone paths, blue-gray tiled houses, and pink cherry trees layer the town surfaces.} \\[0.5pt]
 & \plainimg{0.1185\textwidth}{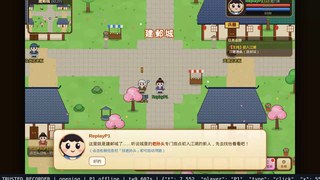} & \scoreimg{sfull}{0.1185\textwidth}{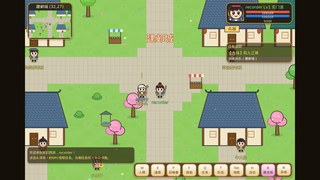} & \scoreimg{spart}{0.1185\textwidth}{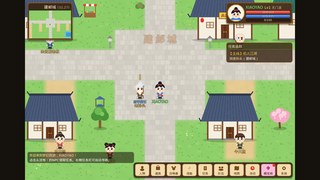} & \scoreimg{szero}{0.1185\textwidth}{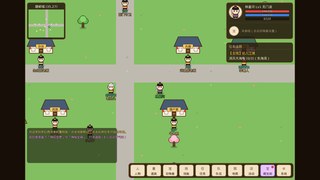} & \scoreimg{spart}{0.1185\textwidth}{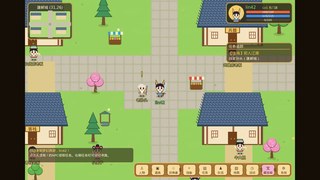} & \scoreimg{spart}{0.1185\textwidth}{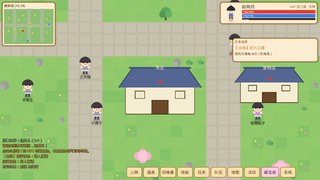} & \scoreimg{szero}{0.1185\textwidth}{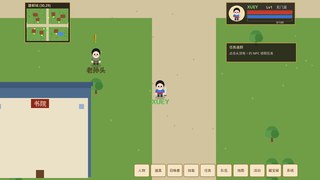} & \scoreimg{szero}{0.1185\textwidth}{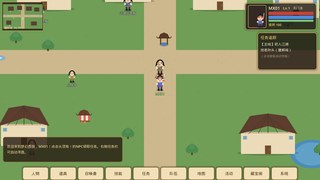} \\[3pt]
\end{tabular}
\caption{\textbf{Reference and recreation frames for five 2D games.} Each row is one visual criterion of one game, stated above the row with its points. The reference frame is taken at the evidence the rubric cites, and each recreation frame at the replay interval the judge cites for the same criterion. The border color gives the points that system earns on the criterion.}
\label{fig:supp-visual-examples}
\end{figure*}

\begin{figure*}[t]
\centering\scriptsize
\scorelegend\\[3pt]
\setlength{\tabcolsep}{0.5pt}
\renewcommand{\arraystretch}{0.9}
\begin{tabular}{@{}ccccccccc@{}}
 & \textbf{Reference} & \textbf{GPT-6-Astra} & \textbf{Claude Opus 5.5} & \textbf{Gemini 4 Argon} & \textbf{GPT-6.1-Sol} & \textbf{DeepSeek V4.1} & \textbf{GLM-5.3-Flash} & \textbf{Qwen3.8-27B} \\[1pt]
\multicolumn{9}{@{}l}{\textbf{Billiards}\enspace {\color{black!60}visual fidelity, 3\,pt}\enspace A blue pocketed table with a surface of about 2:1 and six-pocket rails.} \\[0.5pt]
 & \plainimg{0.1185\textwidth}{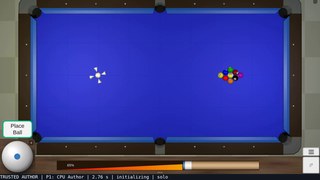} & \scoreimg{spart}{0.1185\textwidth}{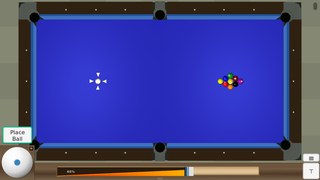} & \scoreimg{spart}{0.1185\textwidth}{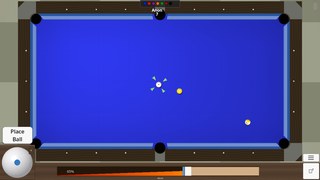} & \scoreimg{spart}{0.1185\textwidth}{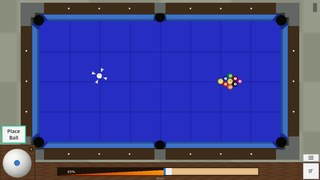} & \scoreimg{spart}{0.1185\textwidth}{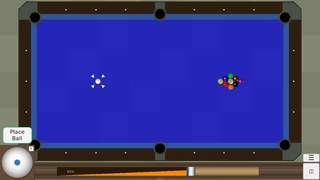} & \scoreimg{spart}{0.1185\textwidth}{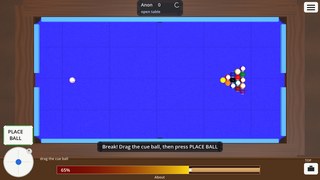} & \scoreimg{spart}{0.1185\textwidth}{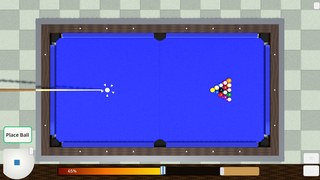} & \scoreimg{spart}{0.1185\textwidth}{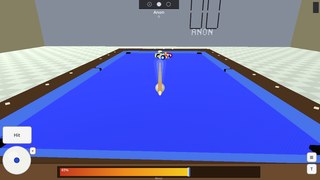} \\[3pt]
\multicolumn{9}{@{}l}{\textbf{SuperTuxKart}\enspace {\color{black!60}visual fidelity, 3\,pt}\enspace The red kart has the low, open-wheeled 3D silhouette of the reference.} \\[0.5pt]
 & \plainimg{0.1185\textwidth}{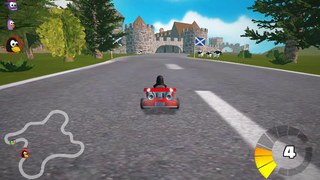} & \scoreimg{spart}{0.1185\textwidth}{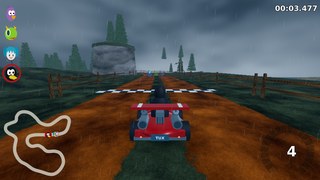} & \scoreimg{spart}{0.1185\textwidth}{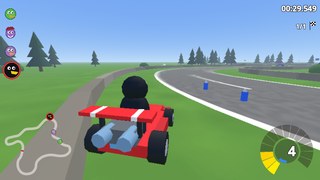} & \scoreimg{spart}{0.1185\textwidth}{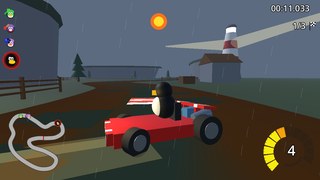} & \scoreimg{spart}{0.1185\textwidth}{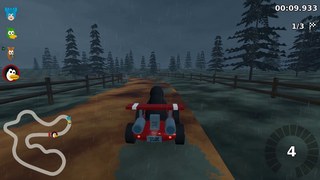} & \scoreimg{spart}{0.1185\textwidth}{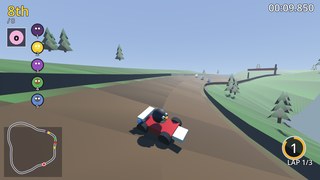} & \scoreimg{spart}{0.1185\textwidth}{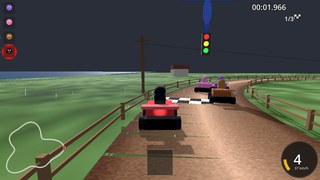} & \scoreimg{szero}{0.1185\textwidth}{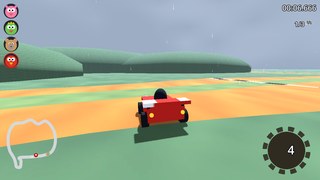} \\[3pt]
\multicolumn{9}{@{}l}{\textbf{Xonotic}\enspace {\color{black!60}visual fidelity, 1.5\,pt}\enspace The combat HUD keeps the compact layout along the screen edges.} \\[0.5pt]
 & \plainimg{0.1185\textwidth}{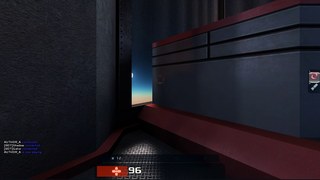} & \scoreimg{spart}{0.1185\textwidth}{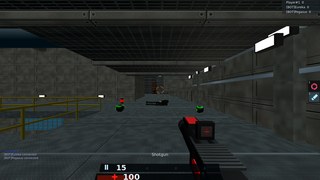} & \scoreimg{spart}{0.1185\textwidth}{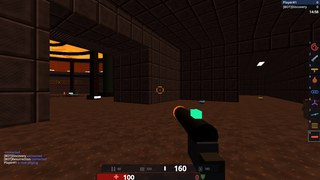} & \scoreimg{szero}{0.1185\textwidth}{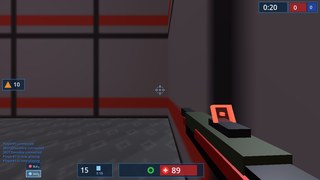} & \scoreimg{spart}{0.1185\textwidth}{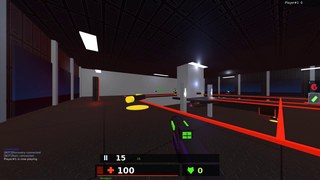} & \scoreimg{spart}{0.1185\textwidth}{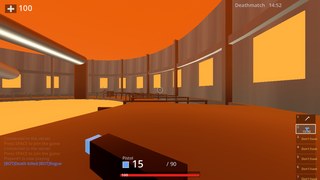} & \scoreimg{spart}{0.1185\textwidth}{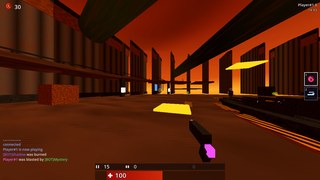} & \scoreimg{spart}{0.1185\textwidth}{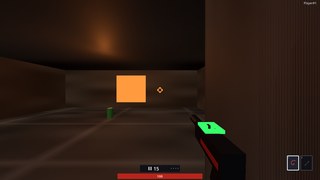} \\[3pt]
\multicolumn{9}{@{}l}{\textbf{Stunt Rally 3}\enspace {\color{black!60}visual fidelity, 1.5\,pt}\enspace The chase camera follows through curves, with the car low and the road ahead filling the view.} \\[0.5pt]
 & \plainimg{0.1185\textwidth}{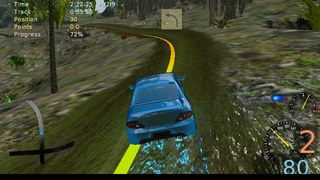} & \scoreimg{spart}{0.1185\textwidth}{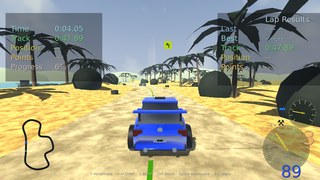} & \scoreimg{spart}{0.1185\textwidth}{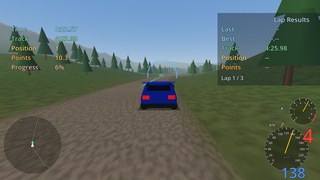} & \scoreimg{sfull}{0.1185\textwidth}{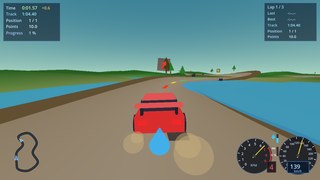} & \scoreimg{sfull}{0.1185\textwidth}{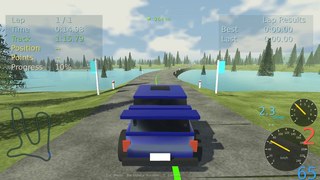} & \scoreimg{spart}{0.1185\textwidth}{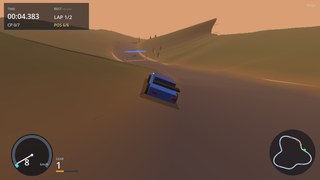} & \scoreimg{spart}{0.1185\textwidth}{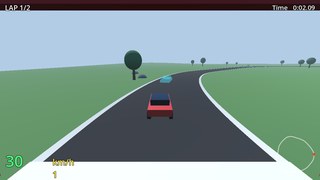} & \scoreimg{szero}{0.1185\textwidth}{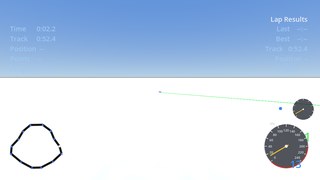} \\[3pt]
\end{tabular}
\caption{\textbf{Reference and recreation frames for four 3D games.} Each row is one visual criterion of one game, stated above the row with its points. The reference frame is taken at the evidence the rubric cites, and each recreation frame at the replay interval the judge cites for the same criterion. The border color gives the points that system earns on the criterion.}
\label{fig:supp-visual-examples-3d}
\end{figure*}

%% file: tables/supp_games.tex
\begin{table*}[p]
\centering\scriptsize
\setlength{\tabcolsep}{2pt}
\renewcommand{\arraystretch}{1.06}
\caption{\textbf{The 100 games of \benchmarkname{}.} Games are grouped by domain. Source is the pinned repository, on GitHub unless a host is given. Ref.\ is whether the reference runs in a browser or as a native program, and whether it is 2D or 3D. Controls are Kbd keyboard, Kbd+M keyboard and mouse, Click point-and-click, Drag drag or swipe, and Pointer continuous pointer. Timing is RT real-time or Disc discrete. Progression is Rounds independent rounds, Levels sequential levels, or World growing world. Players is Single, Co-op, Comp competitive, or Both. Crit.\ and Scen.\ count rubric criteria and demonstration scenarios, and Entry is the share of criteria shown in the entry scenario.}
\label{tab:supp-games}
\begin{tabular*}{\textwidth}{@{\extracolsep{\fill}}rllllllllrrr@{}}
\toprule
\# & Game & Source & Ref. & Domain & Controls & Timing & Progression & Players & Crit. & Scen. & Entry \\
\midrule
1 & A Little Game Called Mario & \begin{tabular}[t]{@{}l@{}}\tiny\texttt{a-little-org-called-mario/}\\\tiny\texttt{a-little-game-called-mario}\end{tabular} & Native 2D & Action &  Kbd+M & RT & Levels & Single & 158 & 13 & 3.2\% \\
2 & Aether Clash & {\tiny\texttt{bangtutorial/aether-clash}} & Web 2D & Action &  Kbd & RT & Rounds & Single & 188 & 8 & 47.3\% \\
3 & Aircraft War Simulation & \begin{tabular}[t]{@{}l@{}}\tiny\texttt{stas4000/}\\\tiny\texttt{aircraft-war-simulation}\end{tabular} & Web 3D & Action &  Kbd & RT & Rounds & Single & 100 & 6 & 53.0\% \\
4 & Arena & {\tiny\texttt{barbarabugajak/Arena}} & Native 3D & Action &  Kbd+M & RT & Rounds & Single & 84 & 3 & 42.9\% \\
5 & Armagetron Advanced & {\tiny\texttt{ArmagetronAd/armagetronad}} & Native 3D & Action &  Kbd & RT & Rounds & Comp & 65 & 8 & 33.8\% \\
6 & AssaultCube & {\tiny\texttt{assaultcube/AC}} & Native 3D & Action &  Kbd+M & RT & Rounds & Comp & 105 & 6 & 26.7\% \\
7 & Blackwater: Silent Harbor & {\tiny\texttt{Hiraeth010/blackwater}} & Web 3D & Action &  Kbd+M & RT & Rounds & Single & 98 & 5 & 25.5\% \\
8 & Bomberrage & {\tiny\texttt{JanSeliv/Bomber}} & Native 3D & Action &  Kbd & RT & Rounds & Comp & 52 & 4 & 76.9\% \\
9 & Breakable & {\tiny\texttt{didier-v/breakable}} & Native 2D & Action &  Pointer & RT & Levels & Single & 91 & 4 & 48.4\% \\
10 & DDNet & {\tiny\texttt{ddnet/ddnet}} & Native 2D & Action &  Kbd+M & RT & Levels & Co-op & 73 & 9 & 20.5\% \\
11 & Dolphin Island 2 & {\tiny\texttt{janmarcano/Dolphin-Island-2}} & Native 2D & Action &  Kbd & RT & Levels & Single & 118 & 8 & 37.3\% \\
12 & Dragon Dragon Fire Fire & \begin{tabular}[t]{@{}l@{}}\tiny\texttt{100-Devs-1-Game/}\\\tiny\texttt{DragonDragonFireFire}\end{tabular} & Native 2D & Action &  Kbd & RT & Levels & Single & 107 & 10 & 31.8\% \\
13 & Drakonas & {\tiny\texttt{Casmo/Drakonas}} & Web 3D & Action &  Pointer & RT & Levels & Single & 82 & 3 & 80.5\% \\
14 & Everplast & {\tiny\texttt{WraithWinterly/Everplast}} & Native 2D & Action &  Kbd+M & RT & Levels & Single & 114 & 12 & 27.2\% \\
15 & Fruit Ninja: Elemental Dojo & \begin{tabular}[t]{@{}l@{}}\tiny\texttt{Liyucheng1997/}\\\tiny\texttt{339\_game-fruit-ninga}\end{tabular} & Web 3D & Action &  Drag & RT & Rounds & Single & 145 & 6 & 24.8\% \\
16 & Hextris & {\tiny\texttt{Hextris/hextris}} & Web 2D & Action &  Kbd & RT & Rounds & Single & 63 & 3 & 73.0\% \\
17 & Librerama & {\tiny\texttt{codeberg.org/Yeldham/librerama}} & Native 2D & Action &  Kbd+M & RT & Rounds & Single & 107 & 8 & 17.8\% \\
18 & Magic Carpet Wizard & {\tiny\texttt{threapchills/MagicCarpetWizard}} & Web 3D & Action &  Kbd+M & RT & Rounds & Comp & 158 & 13 & 31.6\% \\
19 & Meteorite & {\tiny\texttt{Bauxitedev/meteorite}} & Native 3D & Action &  Kbd+M & RT & Levels & Single & 112 & 12 & 10.7\% \\
20 & Neon Snake & {\tiny\texttt{sahandsaeidi/snake-game}} & Web 2D & Action &  Kbd & RT & Rounds & Single & 104 & 3 & 50.0\% \\
21 & OpenArena & {\tiny\texttt{OpenArena/gamecode}} & Native 3D & Action &  Kbd+M & RT & Rounds & Comp & 121 & 23 & 14.0\% \\
22 & OpenClonk & {\tiny\texttt{openclonk/openclonk}} & Native 2D & Action &  Kbd+M & RT & Levels & Both & 92 & 8 & 23.9\% \\
23 & PacMan & {\tiny\texttt{KamranBoroomand/PacMan}} & Web 2D & Action &  Kbd & RT & Levels & Single & 128 & 14 & 32.0\% \\
24 & PulseGrid & {\tiny\texttt{ctbot000/rhythm-game}} & Web 2D & Action &  Kbd & RT & Rounds & Single & 101 & 5 & 51.5\% \\
25 & Red Eclipse & {\tiny\texttt{redeclipse/base}} & Native 3D & Action &  Kbd+M & RT & Rounds & Comp & 109 & 7 & 40.4\% \\
26 & Starship Olympics & \begin{tabular}[t]{@{}l@{}}\tiny\texttt{notapixelstudio/}\\\tiny\texttt{starship-olympics}\end{tabular} & Native 2D & Action &  Kbd & RT & Rounds & Both & 117 & 13 & 11.1\% \\
27 & SuperTux & {\tiny\texttt{supertux/supertux}} & Native 2D & Action &  Kbd & RT & Levels & Single & 102 & 15 & 23.5\% \\
28 & Teeworlds & {\tiny\texttt{teeworlds/teeworlds}} & Native 2D & Action &  Kbd+M & RT & Rounds & Comp & 98 & 8 & 16.3\% \\
29 & TetraForce & {\tiny\texttt{loudsmilestudios/TetraForce}} & Native 2D & Action &  Kbd & RT & Levels & Co-op & 97 & 7 & 28.9\% \\
30 & Tetris & {\tiny\texttt{jakesgordon/javascript-tetris}} & Web 2D & Action &  Kbd & RT & Rounds & Single & 119 & 9 & 36.1\% \\
31 & Threadbare & {\tiny\texttt{endlessm/threadbare}} & Native 2D & Action &  Kbd & RT & Levels & Single & 106 & 9 & 26.4\% \\
32 & Treasure Party & {\tiny\texttt{Empheon/GGJ21}} & Native 3D & Action &  Kbd+M & RT & Rounds & Comp & 77 & 5 & 18.2\% \\
33 & Unvanquished & {\tiny\texttt{Unvanquished/Unvanquished}} & Native 3D & Action &  Kbd+M & RT & Rounds & Comp & 100 & 9 & 26.0\% \\
34 & Xonotic & {\tiny\texttt{gitlab.com/xonotic/xonotic}} & Native 3D & Action &  Kbd+M & RT & Rounds & Comp & 117 & 10 & 23.9\% \\
35 & 2048 & {\tiny\texttt{gabrielecirulli/2048}} & Web 2D & Puzzle &  Kbd & Disc & Rounds & Single & 59 & 4 & 28.8\% \\
36 & Angry Birds HTML5 & \begin{tabular}[t]{@{}l@{}}\tiny\texttt{Liyucheng1997/}\\\tiny\texttt{348\_lab-angry-birds}\end{tabular} & Web 2D & Puzzle &  Drag & Disc & Levels & Single & 104 & 12 & 38.5\% \\
37 & Descensus 2 & {\tiny\texttt{TomWHall/Descensus2}} & Web 2D & Puzzle &  Drag & RT & Levels & Single & 81 & 8 & 29.6\% \\
38 & Frog This Way & \begin{tabular}[t]{@{}l@{}}\tiny\texttt{RockSaladStudio/}\\\tiny\texttt{Ubisoft-Volunteer-Week-Game-Jam--2025}\end{tabular} & Native 3D & Puzzle &  Drag & RT & Levels & Single & 78 & 6 & 34.6\% \\
39 & Gravity Golf & {\tiny\texttt{js13kGames/gravity-golf}} & Web 2D & Puzzle &  Drag & Disc & Levels & Single & 49 & 2 & 83.7\% \\
40 & HTML Minesweeper & {\tiny\texttt{fthielke/html-minesweeper}} & Web 2D & Puzzle &  Click & Disc & Rounds & Single & 82 & 3 & 52.4\% \\
41 & Neverball & {\tiny\texttt{Neverball/neverball}} & Native 3D & Puzzle &  Kbd & RT & Levels & Single & 93 & 14 & 38.7\% \\
42 & Road to Nowhere & {\tiny\texttt{100-Devs-1-Game/RoadToNowhere}} & Native 2D & Puzzle &  Kbd+M & Disc & Levels & Single & 136 & 4 & 35.3\% \\
43 & ROTA & {\tiny\texttt{HarmonyHoney/ROTA}} & Native 2D & Puzzle &  Kbd & RT & Levels & Single & 87 & 8 & 19.5\% \\
44 & Sokoban & {\tiny\texttt{klevze/sokoban}} & Web 2D & Puzzle &  Kbd & Disc & Levels & Single & 118 & 5 & 69.5\% \\
45 & Switchyard & {\tiny\texttt{ctkrug/switchyard}} & Web 2D & Puzzle &  Click & Disc & Levels & Single & 91 & 3 & 50.5\% \\
46 & 0 A.D. & {\tiny\texttt{0ad/0ad}} & Native 3D & Strategy &  Click & RT & Rounds & Comp & 64 & 6 & 17.2\% \\
47 & Courtyard Defense & {\tiny\texttt{seth-xh/pvz}} & Web 2D & Strategy &  Click & RT & Levels & Single & 126 & 8 & 44.4\% \\
48 & Freeciv & {\tiny\texttt{freeciv/freeciv}} & Native 2D & Strategy &  Click & Disc & Rounds & Comp & 87 & 6 & 40.2\% \\
49 & Hedgewars & {\tiny\texttt{hg.hedgewars.org/hedgewars}} & Native 2D & Strategy &  Kbd & Disc & Rounds & Comp & 117 & 11 & 14.5\% \\
50 & Last Beacon & {\tiny\texttt{stackloomdev/last-beacon}} & Web 3D & Strategy &  Kbd+M & RT & Rounds & Single & 120 & 12 & 29.2\% \\
\bottomrule
\end{tabular*}
\end{table*}

\begin{table*}[p]
\centering\scriptsize
\setlength{\tabcolsep}{2pt}
\renewcommand{\arraystretch}{1.06}
\caption*{\textbf{Table~\ref{tab:supp-games} (continued).}}
\begin{tabular*}{\textwidth}{@{\extracolsep{\fill}}rllllllllrrr@{}}
\toprule
\# & Game & Source & Ref. & Domain & Controls & Timing & Progression & Players & Crit. & Scen. & Entry \\
\midrule
51 & MegaGlest & {\tiny\texttt{MegaGlest/megaglest-source}} & Native 3D & Strategy &  Click & RT & Rounds & Comp & 107 & 12 & 34.6\% \\
52 & Mindustry & {\tiny\texttt{Anuken/Mindustry}} & Native 2D & Strategy &  Kbd+M & RT & Levels & Both & 111 & 10 & 12.6\% \\
53 & Petty kingdoms & {\tiny\texttt{js13kGames/petty-kingdoms}} & Web 2D & Strategy &  Click & Disc & Rounds & Comp & 93 & 5 & 79.6\% \\
54 & Super Tux Party & \begin{tabular}[t]{@{}l@{}}\tiny\texttt{gitlab.com/SuperTuxParty/}\\\tiny\texttt{SuperTuxParty}\end{tabular} & Native 3D & Strategy &  Kbd & RT & Rounds & Comp & 150 & 18 & 28.0\% \\
55 & Tanks of Freedom II & {\tiny\texttt{P1X-in/tanks-of-freedom-ii}} & Native 3D & Strategy &  Click & Disc & Levels & Comp & 84 & 6 & 17.9\% \\
56 & The Battle for Wesnoth & {\tiny\texttt{wesnoth/wesnoth}} & Native 2D & Strategy &  Click & Disc & Levels & Comp & 82 & 6 & 14.6\% \\
57 & Warzone 2100 & {\tiny\texttt{Warzone2100/warzone2100}} & Native 3D & Strategy &  Click & RT & Levels & Comp & 110 & 11 & 20.9\% \\
58 & Widelands & {\tiny\texttt{widelands/widelands}} & Native 2D & Strategy &  Click & RT & World & Comp & 91 & 12 & 25.3\% \\
59 & Woof Tactics & {\tiny\texttt{cty41/wooftactics}} & Native 2D & Strategy &  Click & Disc & Levels & Single & 116 & 6 & 48.3\% \\
60 & YouTD 2 & {\tiny\texttt{Praytic/youtd2}} & Native 3D & Strategy &  Click & RT & Rounds & Co-op & 117 & 12 & 41.0\% \\
61 & Dungeon Crawl Stone Soup & {\tiny\texttt{crawl/crawl}} & Native 2D & Role-playing &  Kbd & Disc & Levels & Single & 114 & 5 & 55.3\% \\
62 & Endless Sky & {\tiny\texttt{endless-sky/endless-sky}} & Native 2D & Role-playing &  Kbd+M & RT & World & Single & 104 & 4 & 56.7\% \\
63 & Javascript Gauntlet & \begin{tabular}[t]{@{}l@{}}\tiny\texttt{jakesgordon/}\\\tiny\texttt{javascript-gauntlet}\end{tabular} & Web 2D & Role-playing &  Kbd & RT & Levels & Single & 138 & 10 & 37.7\% \\
64 & Menghuan Xiyou Q & \begin{tabular}[t]{@{}l@{}}\tiny\texttt{Liyucheng1997/}\\\tiny\texttt{197\_game-menghuan-xiyou}\end{tabular} & Web 2D & Role-playing &  Click & Disc & Levels & Single & 139 & 13 & 11.5\% \\
65 & Naev & {\tiny\texttt{naev/naev}} & Native 2D & Role-playing &  Kbd+M & RT & World & Single & 108 & 6 & 17.6\% \\
66 & Roguelike Dungeon & {\tiny\texttt{mitallast/roguelike-dungeon}} & Web 2D & Role-playing &  Kbd & RT & Levels & Single & 110 & 7 & 34.5\% \\
67 & Spire Ascent & {\tiny\texttt{stu-greenshoots/spire-clone}} & Web 2D & Role-playing &  Click & Disc & Levels & Single & 124 & 10 & 39.5\% \\
68 & Sprout Quest & {\tiny\texttt{aibengineering/sprout-quest}} & Web 3D & Role-playing &  Kbd & RT & Levels & Single & 152 & 9 & 19.1\% \\
69 & Veloren & {\tiny\texttt{gitlab.com/veloren/veloren}} & Native 3D & Role-playing &  Kbd+M & RT & World & Both & 114 & 21 & 21.9\% \\
70 & 3D.CITY & {\tiny\texttt{lo-th/3d.city}} & Web 3D & Simulation &  Click & RT & World & Single & 107 & 7 & 33.6\% \\
71 & Cataclysm Dark Days Ahead & {\tiny\texttt{CleverRaven/Cataclysm-DDA}} & Native 2D & Simulation &  Kbd & Disc & World & Single & 82 & 5 & 29.3\% \\
72 & Cookie Clicker (fan remake) & {\tiny\texttt{vardst/cookieclicker}} & Web 2D & Simulation &  Click & RT & World & Single & 127 & 9 & 24.4\% \\
73 & Fishslop & {\tiny\texttt{vasu-devs/FishSlop\_Opus5.5}} & Web 3D & Simulation &  Kbd+M & RT & World & Single & 133 & 4 & 52.6\% \\
74 & Hurry Curry! & \begin{tabular}[t]{@{}l@{}}\tiny\texttt{codeberg.org/hurrycurry/}\\\tiny\texttt{hurrycurry}\end{tabular} & Native 3D & Simulation &  Kbd & RT & Rounds & Co-op & 116 & 9 & 28.4\% \\
75 & Libre TrainSim & {\tiny\texttt{Libre-TrainSim/Libre-TrainSim}} & Native 3D & Simulation &  Kbd & RT & Rounds & Single & 99 & 9 & 37.4\% \\
76 & Oolite & {\tiny\texttt{OoliteProject/oolite}} & Native 3D & Simulation &  Kbd & RT & World & Single & 120 & 9 & 30.8\% \\
77 & OpenTTD & {\tiny\texttt{OpenTTD/OpenTTD}} & Native 2D & Simulation &  Click & RT & World & Both & 97 & 4 & 45.4\% \\
78 & Pioneer & {\tiny\texttt{pioneerspacesim/pioneer}} & Native 3D & Simulation &  Kbd+M & RT & World & Single & 121 & 7 & 10.7\% \\
79 & Portobugia & {\tiny\texttt{AudranBert/GameOff2021}} & Native 3D & Simulation &  Click & RT & World & Single & 121 & 14 & 7.4\% \\
80 & Thrive & {\tiny\texttt{Revolutionary-Games/Thrive}} & Native 3D & Simulation &  Kbd+M & RT & World & Single & 69 & 10 & 20.3\% \\
81 & VoxeLibre & \begin{tabular}[t]{@{}l@{}}\tiny\texttt{git.minetest.land/VoxeLibre/}\\\tiny\texttt{VoxeLibre}\end{tabular} & Native 3D & Simulation &  Kbd+M & RT & World & Co-op & 101 & 10 & 7.9\% \\
82 & Witch's Apothecary & {\tiny\texttt{StefanCla/ShovelJam2025}} & Native 2D & Simulation &  Kbd+M & Disc & Levels & Single & 74 & 4 & 50.0\% \\
83 & Billiards & {\tiny\texttt{tailuge/billiards}} & Web 3D & Racing \& sports &  Kbd+M & Disc & Rounds & Comp & 108 & 17 & 13.9\% \\
84 & Konterball & {\tiny\texttt{madebywild/konterball}} & Web 3D & Racing \& sports &  Pointer & RT & Rounds & Comp & 45 & 2 & 84.4\% \\
85 & Project Kuruma & {\tiny\texttt{FlynnTaggart/RacingGame}} & Native 3D & Racing \& sports &  Kbd & RT & Rounds & Single & 94 & 7 & 38.3\% \\
86 & SCREAM2 & {\tiny\texttt{zyphersworld/SCREAM2}} & Native 3D & Racing \& sports &  Kbd & RT & Rounds & Single & 90 & 5 & 45.6\% \\
87 & Speed Dreams & \begin{tabular}[t]{@{}l@{}}\tiny\texttt{forge.a-lec.org/speed-dreams/}\\\tiny\texttt{speed-dreams-code}\end{tabular} & Native 3D & Racing \& sports &  Kbd & RT & Rounds & Comp & 82 & 5 & 34.1\% \\
88 & Stunt Rally 3 & {\tiny\texttt{stuntrally/stuntrally3}} & Native 3D & Racing \& sports &  Kbd & RT & Rounds & Comp & 135 & 27 & 17.0\% \\
89 & SuperTuxKart & {\tiny\texttt{supertuxkart/stk-code}} & Native 3D & Racing \& sports &  Kbd & RT & Levels & Comp & 103 & 6 & 36.9\% \\
90 & Turbo Kart Grand Prix & \begin{tabular}[t]{@{}l@{}}\tiny\texttt{Franklin-C/}\\\tiny\texttt{turbo-kart-grand-prix}\end{tabular} & Web 3D & Racing \& sports &  Kbd & RT & Rounds & Single & 140 & 5 & 47.1\% \\
91 & Clock Out Unseen & {\tiny\texttt{Ryan-fm/clockout-unseen}} & Web 2D & Narrative \& stealth &  Kbd & RT & Levels & Single & 132 & 10 & 25.8\% \\
92 & Connection Lost & \begin{tabular}[t]{@{}l@{}}\tiny\texttt{PandorasBoxStudios/}\\\tiny\texttt{team-black-connectionlost}\end{tabular} & Native 3D & Narrative \& stealth &  Kbd+M & RT & Levels & Single & 130 & 13 & 20.0\% \\
93 & Escape 42 & {\tiny\texttt{itaruf/GameOff2023}} & Native 3D & Narrative \& stealth &  Kbd+M & RT & Levels & Single & 91 & 16 & 24.2\% \\
94 & Lakeside: The Eighth Waking & {\tiny\texttt{Beats0/rusty-lake}} & Web 2D & Narrative \& stealth &  Click & Disc & Levels & Single & 173 & 2 & 90.2\% \\
95 & Mental Record & {\tiny\texttt{100-Devs-1-Game/MentalRecord}} & Native 2D & Narrative \& stealth &  Click & Disc & Levels & Single & 115 & 6 & 26.1\% \\
96 & OPEN YOUR EYES & {\tiny\texttt{itaruf/GMTK2025}} & Native 3D & Narrative \& stealth &  Kbd+M & RT & Levels & Single & 75 & 7 & 32.0\% \\
97 & Silent Meridian & {\tiny\texttt{stackloomdev/silent-meridian}} & Web 2D & Narrative \& stealth &  Click & Disc & Levels & Single & 147 & 6 & 44.9\% \\
98 & The Dark Mod & {\tiny\texttt{stgatilov/darkmod\_src}} & Native 3D & Narrative \& stealth &  Kbd+M & RT & Rounds & Single & 94 & 9 & 22.3\% \\
99 & The Full Picture & {\tiny\texttt{100-Devs-1-Game/TheFullPicture}} & Native 2D & Narrative \& stealth &  Click & Disc & Levels & Single & 96 & 1 & 100.0\% \\
100 & Xuanling: Under the Eternal Lamp & \begin{tabular}[t]{@{}l@{}}\tiny\texttt{Liyucheng1997/}\\\tiny\texttt{340\_game-xuanling-tomb}\end{tabular} & Web 3D & Narrative \& stealth &  Kbd+M & RT & Levels & Single & 149 & 10 & 21.5\% \\
\bottomrule
\end{tabular*}
\end{table*}

%% file: main.bib
@String(CVPR= {IEEE Conf. Comput. Vis. Pattern Recog.})

@String(ECCV= {Eur. Conf. Comput. Vis.})

@String(CVPR  = {CVPR})

@String(ECCV  = {ECCV})

@inproceedings{yang2025swebenchmultimodal,
  author = {John Yang and Carlos E. Jimenez and Alex L. Zhang and others},
  title = {{SWE-bench Multimodal: Do AI Systems Generalize to Visual Software Domains?}},
  booktitle = {International Conference on Learning Representations},
  year = {2025},
  url = {https://proceedings.iclr.cc/paper_files/paper/2025/hash/07d6332ae36730707fddddba736d7b6c-Abstract-Conference.html}
}

@inproceedings{si2025design2code,
  author = {Chenglei Si and Yanzhe Zhang and Ryan Li and Zhengyuan Yang and Ruibo Liu and Diyi Yang},
  title = {{Design2Code: Benchmarking Multimodal Code Generation for Automated Front-End Engineering}},
  booktitle = {Proceedings of the 2025 Conference of the Nations of the Americas Chapter of the Association for Computational Linguistics: Human Language Technologies (Volume 1: Long Papers)},
  year = {2025},
  pages = {3956--3974},
  doi = {10.18653/v1/2025.naacl-long.199},
  url = {https://aclanthology.org/2025.naacl-long.199/}
}

@inproceedings{tan2024cradle,
  author = {Tan, Weihao and Zhang, Wentao and Xu, Xinrun and others},
  title = {{Cradle}: Empowering Foundation Agents Towards General Computer Control},
  booktitle = {International Conference on Machine Learning},
  pages = {58658--58725},
  year = {2025},
  url = {https://arxiv.org/abs/2403.03186}
}

@article{luo2026gamecraft,
  author = {Luo, Tongxu and Wang, Rongsheng and Bi, Jiaxi and others},
  title = {{GameCraft-Bench}: Can Agents Build Playable Games End-to-End in a Real Game Engine?},
  journal = {arXiv preprint arXiv:2606.17861},
  year = {2026},
  url = {https://arxiv.org/abs/2606.17861}
}

@article{chen2026swegame,
  author = {Chen, Xiaoyu and Wei, Lai and Wang, Jin and others},
  title = {{SWE-Game}: Can Coding Agents Build the Games We Want?},
  journal = {arXiv preprint arXiv:2609.33678},
  year = {2026},
  url = {https://arxiv.org/abs/2609.33678}
}

@article{qiao2026gamereplica,
  author = {Qiao, Boyu and Tang, Zixin and Hao, Xiaoshuai and Li, Wenbo},
  title = {{GameReplica}: A Benchmark for Black-Box Visual Game Replication by Vision-Language Agents},
  journal = {arXiv preprint arXiv:2609.22308},
  year = {2026},
  url = {https://arxiv.org/abs/2609.22308}
}

@inproceedings{paglieri2024balrog,
  author = {Davide Paglieri and Bart{\l}omiej Cupia{\l} and Samuel Coward and others},
  title = {{BALROG: Benchmarking Agentic LLM and VLM Reasoning On Games}},
  booktitle = {International Conference on Learning Representations},
  year = {2025},
  url = {https://proceedings.iclr.cc/paper_files/paper/2025/hash/f0b1515be276f6ba82b4f2b25e50bef0-Abstract-Conference.html}
}

@inproceedings{hu2025lmgame,
  author = {Lanxiang Hu and Mingjia Huo and Yuxuan Zhang and others},
  title = {{lmgame-Bench: How Good are LLMs at Playing Games?}},
  booktitle = {International Conference on Learning Representations},
  year = {2026},
  url = {https://proceedings.iclr.cc/paper_files/paper/2026/hash/83a4ea71b13bc86308a2bd0b5e07fb61-Abstract-Conference.html}
}

@article{zhang2025videogamebench,
  author = {Alex L. Zhang and Thomas L. Griffiths and Karthik R. Narasimhan and others},
  title = {{VideoGameBench: Can Vision-Language Models complete popular video games?}},
  journal = {arXiv preprint arXiv:2505.18134},
  year = {2025},
  url = {https://arxiv.org/html/2505.18134v3}
}

@inproceedings{ouyang2026gameworld,
  author = {Mingyu Ouyang and Siyuan Hu and Kevin Qinghong Lin and others},
  title = {{GameWorld: Towards Standardized and Verifiable Evaluation of Multimodal Game Agents}},
  booktitle = {European Conference on Computer Vision (ECCV)},
  year = {2026},
  url = {https://arxiv.org/abs/2604.07429}
}

@inproceedings{chi2026gamedevbench,
  author = {Wayne Chi and Yixiong Fang and Arnav Yayavaram and others},
  title = {{GameDevBench: Evaluating Agentic Capabilities Through Game Development}},
  booktitle = {International Conference on Machine Learning},
  year = {2026},
  url = {https://proceedings.mlr.press/v306/chi26a.html}
}

@article{la2026gameenginebench,
  author = {Brian La and Sejoon Chang and Ben Kim and others},
  title = {{GameEngineBench: Evaluating Coding Agents on Real C++ Runtime Environments}},
  journal = {arXiv preprint arXiv:2607.03525},
  year = {2026},
  url = {https://arxiv.org/html/2607.03525v2}
}

@inproceedings{zhang2025vgamegym,
  author = {Wei Zhang and Jian Yang and Renshuai Tao and others},
  title = {{V-GameGym: Visual Game Generation for Code Large Language Models}},
  booktitle = {Findings of the Association for Computational Linguistics: ACL 2026},
  pages = {5613--5641},
  year = {2026},
  url = {https://aclanthology.org/2026.findings-acl.276/}
}

@article{jiang2026opengame,
  author = {Yilei Jiang and Jinyuan Hu and Qianyin Xiao and others},
  title = {{OpenGame: Open Agentic Coding for Games}},
  journal = {arXiv preprint arXiv:2604.18394},
  year = {2026},
  url = {https://arxiv.org/abs/2604.18394}
}

@article{zhang2026webgamebench,
  author = {Wenyu Zhang and Guoliang You and Tianlun and others},
  title = {{WebGameBench: Requirement-to-Application Evaluation for Coding Agents via Browser-Native Games}},
  journal = {arXiv preprint arXiv:2605.17637},
  year = {2026},
  url = {https://arxiv.org/html/2605.17637v2}
}

@article{sun2026jamer,
  author = {Jianwen Sun and Chuanhao Li and Zizhen Li and others},
  title = {{JAMER: Project-Level Code Framework Dataset and Benchmark on Professional Game Engines}},
  journal = {arXiv preprint arXiv:2606.19830},
  year = {2026},
  url = {https://arxiv.org/html/2606.19830v2}
}

@article{zhang2026gameasg,
  author = {Xiuhui Zhang and Yi Chen and Shusheng Xu and others},
  title = {{GameASG-Bench: Benchmarking Autonomous Software Generation for Game Development}},
  journal = {arXiv preprint arXiv:2609.21293},
  year = {2026},
  url = {https://arxiv.org/html/2609.21293v1}
}

@inproceedings{park2025orak,
  author = {Dongmin Park and Minkyu Kim and Beongjun Choi and others},
  title = {{Orak: A Foundational Benchmark for Training and Evaluating LLM Agents on Diverse Video Games}},
  booktitle = {International Conference on Learning Representations},
  year = {2026},
  url = {https://proceedings.iclr.cc/paper_files/paper/2026/hash/63506d49e90266a51f21ba8edbd49f46-Abstract-Conference.html}
}

@article{che2026gamelogic,
  author = {Xinyu Che and Yunfei Ge and Shihao Li and others},
  title = {{GameLogicBench: Evaluating Coding Agents on Runtime Game Logic with Tick-Level State Assertions}},
  journal = {arXiv preprint arXiv:2609.21562},
  year = {2026},
  url = {https://arxiv.org/html/2609.21562v2}
}

@article{chen2026gamexpert,
  author = {Kun Chen and Haorong Hong and Peizhong Gao and others},
  title = {{GameXpert-Bench: How Far Are Coding Agents from Expert Game Development?}},
  journal = {arXiv preprint arXiv:2608.21833},
  year = {2026},
  url = {https://arxiv.org/html/2608.21833v1}
}

@article{jia2026gamegenverifier,
  author = {Chaobo Jia and Ruipeng Wan and Ting Sun and others},
  title = {{GameGen-Verifier: Parallel Keypoint-Based Verification for LLM-Generated Games via Runtime State Injection}},
  journal = {arXiv preprint arXiv:2605.07442},
  year = {2026},
  url = {https://arxiv.org/html/2605.07442v1}
}

@article{chen2021codex,
  author = {Mark Chen and Jerry Tworek and Heewoo Jun and others},
  title = {{Evaluating Large Language Models Trained on Code}},
  journal = {arXiv preprint arXiv:2107.03374},
  year = {2021},
  url = {https://arxiv.org/abs/2107.03374}
}

@article{austin2021mbpp,
  author = {Jacob Austin and Augustus Odena and Maxwell Nye and others},
  title = {{Program Synthesis with Large Language Models}},
  journal = {arXiv preprint arXiv:2108.07732},
  year = {2021},
  url = {https://arxiv.org/abs/2108.07732}
}

@inproceedings{hendrycks2021apps,
  author = {Dan Hendrycks and Steven Basart and Saurav Kadavath and others},
  title = {{Measuring Coding Challenge Competence With APPS}},
  booktitle = {Advances in Neural Information Processing Systems Datasets and Benchmarks Track},
  year = {2021},
  url = {https://arxiv.org/abs/2105.09938}
}

@article{li2022alphacode,
  author = {Yujia Li and David Choi and Junyoung Chung and others},
  title = {{Competition-Level Code Generation with AlphaCode}},
  journal = {Science},
  volume = {378},
  number = {6624},
  pages = {1092--1097},
  year = {2022},
  doi = {10.1126/science.abq1158}
}

@inproceedings{jimenez2024swebench,
  author = {Carlos E. Jimenez and John Yang and Alexander Wettig and others},
  title = {{SWE-bench: Can Language Models Resolve Real-World GitHub Issues?}},
  booktitle = {International Conference on Learning Representations},
  year = {2024},
  url = {https://arxiv.org/abs/2310.06770}
}

@inproceedings{yang2024sweagent,
  author = {John Yang and Carlos E. Jimenez and Alexander Wettig and others},
  title = {{SWE-agent: Agent-Computer Interfaces Enable Automated Software Engineering}},
  booktitle = {Advances in Neural Information Processing Systems},
  year = {2024},
  url = {https://arxiv.org/abs/2405.15793}
}

@inproceedings{wang2025openhands,
  author = {Xingyao Wang and Boxuan Li and Yufan Song and others},
  title = {{OpenHands: An Open Platform for AI Software Developers as Generalist Agents}},
  booktitle = {International Conference on Learning Representations},
  year = {2025},
  url = {https://arxiv.org/abs/2407.16741}
}

@inproceedings{jain2025livecodebench,
  author = {Naman Jain and King Han and Alex Gu and others},
  title = {{LiveCodeBench: Holistic and Contamination Free Evaluation of Large Language Models for Code}},
  booktitle = {International Conference on Learning Representations},
  year = {2025},
  url = {https://arxiv.org/abs/2403.07974}
}

@inproceedings{zhuo2025bigcodebench,
  author = {Terry Yue Zhuo and Minh Chien Vu and Jenny Chim and others},
  title = {{BigCodeBench: Benchmarking Code Generation with Diverse Function Calls and Complex Instructions}},
  booktitle = {International Conference on Learning Representations},
  year = {2025},
  url = {https://arxiv.org/abs/2406.15877}
}

@inproceedings{wang2024codeact,
  author = {Xingyao Wang and Yangyi Chen and Lifan Yuan and others},
  title = {{Executable Code Actions Elicit Better LLM Agents}},
  booktitle = {International Conference on Machine Learning},
  year = {2024},
  url = {https://arxiv.org/abs/2402.01030}
}

@inproceedings{qian2024chatdev,
  author = {Chen Qian and Wei Liu and Hongzhang Liu and others},
  title = {{ChatDev: Communicative Agents for Software Development}},
  booktitle = {Annual Meeting of the Association for Computational Linguistics},
  year = {2024},
  url = {https://arxiv.org/abs/2307.07924}
}

@inproceedings{hong2024metagpt,
  author = {Sirui Hong and Mingchen Zhuge and Jiaqi Chen and others},
  title = {{MetaGPT: Meta Programming for A Multi-Agent Collaborative Framework}},
  booktitle = {International Conference on Learning Representations},
  year = {2024},
  url = {https://arxiv.org/abs/2308.00352}
}

@inproceedings{chen2024selfdebug,
  author = {Xinyun Chen and Maxwell Lin and Nathanael Sch{\"a}rli and others},
  title = {{Teaching Large Language Models to Self-Debug}},
  booktitle = {International Conference on Learning Representations},
  year = {2024},
  url = {https://arxiv.org/abs/2304.05128}
}

@inproceedings{shinn2023reflexion,
  author = {Noah Shinn and Federico Cassano and Edward Berman and others},
  title = {{Reflexion: Language Agents with Verbal Reinforcement Learning}},
  booktitle = {Advances in Neural Information Processing Systems},
  year = {2023},
  url = {https://arxiv.org/abs/2303.11366}
}

@inproceedings{yao2023react,
  author = {Shunyu Yao and Jeffrey Zhao and Dian Yu and others},
  title = {{ReAct: Synergizing Reasoning and Acting in Language Models}},
  booktitle = {International Conference on Learning Representations},
  year = {2023},
  url = {https://arxiv.org/abs/2210.03629}
}

@article{xia2024agentless,
  author = {Chunqiu Steven Xia and Yinlin Deng and Soren Dunn and others},
  title = {{Demystifying LLM-Based Software Engineering Agents}},
  journal = {Proceedings of the ACM on Software Engineering},
  volume = {2},
  number = {FSE},
  year = {2025},
  doi = {10.1145/3715754},
  url = {https://doi.org/10.1145/3715754}
}

@inproceedings{liu2024agentbench,
  author = {Xiao Liu and Hao Yu and Hanchen Zhang and others},
  title = {{AgentBench: Evaluating LLMs as Agents}},
  booktitle = {International Conference on Learning Representations},
  year = {2024},
  url = {https://arxiv.org/abs/2308.03688}
}

@inproceedings{zhao2025commit0,
  author = {Wenting Zhao and Nan Jiang and Celine Lee and others},
  title = {{Commit0: Library Generation from Scratch}},
  booktitle = {International Conference on Learning Representations},
  year = {2025},
  url = {https://arxiv.org/abs/2412.01769}
}

@inproceedings{starace2025paperbench,
  author = {Giulio Starace and Oliver Jaffe and Dane Sherburn and others},
  title = {{PaperBench: Evaluating AI's Ability to Replicate AI Research}},
  booktitle = {International Conference on Machine Learning},
  year = {2025},
  url = {https://arxiv.org/abs/2504.01848}
}

@inproceedings{chan2025mlebench,
  author = {Jun Shern Chan and Neil Chowdhury and Oliver Jaffe and others},
  title = {{MLE-bench: Evaluating Machine Learning Agents on Machine Learning Engineering}},
  booktitle = {International Conference on Learning Representations},
  year = {2025},
  url = {https://arxiv.org/abs/2410.07095}
}

@article{laurencon2024websight,
  author = {Hugo Lauren{\c{c}}on and L{\'e}o Tronchon and Victor Sanh},
  title = {{Unlocking the conversion of Web Screenshots into HTML Code with the WebSight Dataset}},
  journal = {arXiv preprint arXiv:2403.09029},
  year = {2024},
  url = {https://arxiv.org/abs/2403.09029}
}

@inproceedings{wu2024plot2code,
  author = {Chengyue Wu and Zhixuan Liang and Yixiao Ge and Qiushan Guo and others},
  title = {{Plot2Code: A Comprehensive Benchmark for Evaluating Multi-modal Large Language Models in Code Generation from Scientific Plots}},
  booktitle = {Findings of the Association for Computational Linguistics: NAACL 2025},
  pages = {3006--3028},
  year = {2025},
  doi = {10.18653/v1/2025.findings-naacl.164}
}

@inproceedings{yang2025chartmimic,
  author = {Cheng Yang and Chufan Shi and Yaxin Liu and others},
  title = {{ChartMimic: Evaluating LMM's Cross-Modal Reasoning Capability via Chart-to-Code Generation}},
  booktitle = {International Conference on Learning Representations},
  year = {2025},
  url = {https://arxiv.org/abs/2406.09961}
}

@inproceedings{li2024sketch2code,
  author = {Ryan Li and Yanzhe Zhang and Diyi Yang},
  title = {{Sketch2Code: Evaluating Vision-Language Models for Interactive Web Design Prototyping}},
  booktitle = {Proceedings of the 2025 Conference of the Nations of the Americas Chapter of the Association for Computational Linguistics: Human Language Technologies (NAACL)},
  pages = {3921--3955},
  year = {2025},
  url = {https://aclanthology.org/2025.naacl-long.198/}
}

@inproceedings{xiao2024interaction2code,
  author = {Jingyu Xiao and Yuxuan Wan and Yintong Huo and others},
  title = {{Interaction2Code: Benchmarking MLLM-based Interactive Webpage Code Generation from Interactive Prototyping}},
  booktitle = {IEEE/ACM International Conference on Automated Software Engineering (ASE)},
  year = {2025},
  doi = {10.1109/ASE63991.2025.00028}
}

@inproceedings{yun2024web2code,
  author = {Sukmin Yun and Haokun Lin and Rusiru Thushara and others},
  title = {{Web2Code: A Large-scale Webpage-to-Code Dataset and Evaluation Framework for Multimodal LLMs}},
  booktitle = {Advances in Neural Information Processing Systems Datasets and Benchmarks Track},
  year = {2024},
  url = {https://arxiv.org/abs/2406.20098}
}

@inproceedings{li2024mmcode,
  author = {Kaixin Li and Yuchen Tian and Qisheng Hu and others},
  title = {{MMCode: Benchmarking Multimodal Large Language Models for Code Generation with Visually Rich Programming Problems}},
  booktitle = {Findings of the Association for Computational Linguistics: EMNLP 2024},
  pages = {736--783},
  year = {2024},
  doi = {10.18653/v1/2024.findings-emnlp.42},
  url = {https://aclanthology.org/2024.findings-emnlp.42}
}

@article{zhang2025artifactsbench,
  author = {Chenchen Zhang and Yuhang Li and Can Xu and others},
  title = {{ArtifactsBench: Bridging the Visual-Interactive Gap in LLM Code Generation Evaluation}},
  journal = {arXiv preprint arXiv:2507.04952},
  year = {2025},
  url = {https://arxiv.org/abs/2507.04952}
}

@inproceedings{zhou2024webarena,
  author = {Shuyan Zhou and Frank F. Xu and Hao Zhu and others},
  title = {{WebArena: A Realistic Web Environment for Building Autonomous Agents}},
  booktitle = {International Conference on Learning Representations},
  year = {2024},
  url = {https://arxiv.org/abs/2307.13854}
}

@inproceedings{koh2024visualwebarena,
  author = {Jing Yu Koh and Robert Lo and Lawrence Jang and others},
  title = {{VisualWebArena: Evaluating Multimodal Agents on Realistic Visual Web Tasks}},
  booktitle = {Annual Meeting of the Association for Computational Linguistics},
  year = {2024},
  url = {https://arxiv.org/abs/2401.13649}
}

@inproceedings{xie2024osworld,
  author = {Tianbao Xie and Danyang Zhang and Jixuan Chen and others},
  title = {{OSWorld: Benchmarking Multimodal Agents for Open-Ended Tasks in Real Computer Environments}},
  booktitle = {Advances in Neural Information Processing Systems},
  year = {2024},
  url = {https://arxiv.org/abs/2404.07972}
}

@inproceedings{deng2023mind2web,
  author = {Xiang Deng and Yu Gu and Boyuan Zheng and others},
  title = {{Mind2Web: Towards a Generalist Agent for the Web}},
  booktitle = {Advances in Neural Information Processing Systems Datasets and Benchmarks Track},
  year = {2023},
  url = {https://arxiv.org/abs/2306.06070}
}

@inproceedings{zheng2024seeact,
  author = {Boyuan Zheng and Boyu Gou and Jihyung Kil and others},
  title = {{GPT-4V(ision) is a Generalist Web Agent, if Grounded}},
  booktitle = {International Conference on Machine Learning},
  year = {2024},
  url = {https://arxiv.org/abs/2401.01614}
}

@inproceedings{hong2024cogagent,
  author = {Wenyi Hong and Weihan Wang and Qingsong Lv and others},
  title = {{CogAgent: A Visual Language Model for GUI Agents}},
  booktitle = {CVPR},
  year = {2024},
  url = {https://arxiv.org/abs/2312.08914}
}

@article{qin2025uitars,
  author = {Yujia Qin and Yining Ye and Junjie Fang and others},
  title = {{UI-TARS: Pioneering Automated GUI Interaction with Native Agents}},
  journal = {arXiv preprint arXiv:2501.12326},
  year = {2025},
  url = {https://arxiv.org/abs/2501.12326}
}

@inproceedings{rawles2025androidworld,
  author = {Christopher Rawles and Sarah Clinckemaillie and Yifan Chang and others},
  title = {{AndroidWorld: A Dynamic Benchmarking Environment for Autonomous Agents}},
  booktitle = {International Conference on Learning Representations},
  year = {2025},
  url = {https://arxiv.org/abs/2405.14573}
}

@article{yang2023setofmark,
  author = {Jianwei Yang and Hao Zhang and Feng Li and others},
  title = {{Set-of-Mark Prompting Unleashes Extraordinary Visual Grounding in GPT-4V}},
  journal = {arXiv preprint arXiv:2310.11441},
  year = {2023},
  url = {https://arxiv.org/abs/2310.11441}
}

@article{bellemare2013ale,
  author = {Marc G. Bellemare and Yavar Naddaf and Joel Veness and others},
  title = {{The Arcade Learning Environment: An Evaluation Platform for General Agents}},
  journal = {Journal of Artificial Intelligence Research},
  year = {2013},
  url = {https://arxiv.org/abs/1207.4708}
}

@article{berner2019openaifive,
  author = {{OpenAI}},
  title = {{Dota 2 with Large Scale Deep Reinforcement Learning}},
  journal = {arXiv preprint arXiv:1912.06680},
  year = {2019},
  url = {https://arxiv.org/abs/1912.06680}
}

@inproceedings{kuttler2020nethack,
  author = {Heinrich K{\"u}ttler and Nantas Nardelli and Alexander H. Miller and others},
  title = {{The NetHack Learning Environment}},
  booktitle = {Advances in Neural Information Processing Systems},
  year = {2020},
  url = {https://arxiv.org/abs/2006.13760}
}

@inproceedings{hafner2022crafter,
  author = {Danijar Hafner},
  title = {{Benchmarking the Spectrum of Agent Capabilities}},
  booktitle = {International Conference on Learning Representations},
  year = {2022},
  url = {https://arxiv.org/abs/2109.06780}
}

@inproceedings{fan2022minedojo,
  author = {Linxi Fan and Guanzhi Wang and Yunfan Jiang and others},
  title = {{MineDojo: Building Open-Ended Embodied Agents with Internet-Scale Knowledge}},
  booktitle = {Advances in Neural Information Processing Systems Datasets and Benchmarks Track},
  year = {2022},
  url = {https://arxiv.org/abs/2206.08853}
}

@inproceedings{baker2022vpt,
  author = {Bowen Baker and Ilge Akkaya and Peter Zhokhov and others},
  title = {{Video PreTraining (VPT): Learning to Act by Watching Unlabeled Online Videos}},
  booktitle = {Advances in Neural Information Processing Systems},
  year = {2022},
  url = {https://arxiv.org/abs/2206.11795}
}

@article{wang2023voyager,
  author = {Guanzhi Wang and Yuqi Xie and Yunfan Jiang and others},
  title = {{Voyager: An Open-Ended Embodied Agent with Large Language Models}},
  journal = {Transactions on Machine Learning Research},
  year = {2024},
  url = {https://openreview.net/forum?id=ehfRiF0R3a}
}

@article{sima2024scaling,
  author = {{SIMA Team}},
  title = {{Scaling Instructable Agents Across Many Simulated Worlds}},
  journal = {arXiv preprint arXiv:2404.10179},
  year = {2024},
  url = {https://arxiv.org/abs/2404.10179}
}

@inproceedings{wu2024smartplay,
  author = {Yue Wu and Xuan Tang and Tom M. Mitchell and others},
  title = {{SmartPlay: A Benchmark for LLMs as Intelligent Agents}},
  booktitle = {International Conference on Learning Representations},
  year = {2024},
  url = {https://arxiv.org/abs/2310.01557}
}

@article{ecoffet2021goexplore,
  author = {Adrien Ecoffet and Joost Huizinga and Joel Lehman and others},
  title = {{First return, then explore}},
  journal = {Nature},
  volume = {590},
  number = {7847},
  pages = {580--586},
  year = {2021},
  doi = {10.1038/s41586-020-03157-9}
}

@inproceedings{pathak2017curiosity,
  author = {Deepak Pathak and Pulkit Agrawal and Alexei A. Efros and others},
  title = {{Curiosity-driven Exploration by Self-supervised Prediction}},
  booktitle = {International Conference on Machine Learning},
  year = {2017},
  url = {https://arxiv.org/abs/1705.05363}
}

@article{ha2018worldmodels,
  author = {David Ha and J{\"u}rgen Schmidhuber},
  title = {{World Models}},
  journal = {arXiv preprint arXiv:1803.10122},
  year = {2018},
  url = {https://arxiv.org/abs/1803.10122}
}

@inproceedings{kim2020gamegan,
  author = {Seung Wook Kim and Yuhao Zhou and Jonah Philion and others},
  title = {{Learning to Simulate Dynamic Environments with GameGAN}},
  booktitle = {CVPR},
  year = {2020},
  url = {https://arxiv.org/abs/2005.12126}
}

@inproceedings{bruce2024genie,
  author = {Jake Bruce and Michael Dennis and Ashley Edwards and others},
  title = {{Genie: Generative Interactive Environments}},
  booktitle = {International Conference on Machine Learning},
  year = {2024},
  url = {https://arxiv.org/abs/2402.15391}
}

@inproceedings{valevski2025gamengen,
  author = {Dani Valevski and Yaniv Leviathan and Moab Arar and others},
  title = {{Diffusion Models Are Real-Time Game Engines}},
  booktitle = {International Conference on Learning Representations},
  year = {2025},
  url = {https://arxiv.org/abs/2408.14837}
}

@inproceedings{alonso2024diamond,
  author = {Eloi Alonso and Adam Jelley and Vincent Micheli and others},
  title = {{Diffusion for World Modeling: Visual Details Matter in Atari}},
  booktitle = {Advances in Neural Information Processing Systems},
  year = {2024},
  url = {https://arxiv.org/abs/2405.12399}
}

@inproceedings{tang2024worldcoder,
  author = {Hao Tang and Darren Key and Kevin Ellis},
  title = {{WorldCoder, a Model-Based LLM Agent: Building World Models by Writing Code and Interacting with the Environment}},
  booktitle = {Advances in Neural Information Processing Systems},
  year = {2024},
  url = {https://arxiv.org/abs/2402.12275}
}

@inproceedings{dainese2024codeworld,
  author = {Nicola Dainese and Matteo Merler and Minttu Alakuijala and others},
  title = {{Generating Code World Models with Large Language Models Guided by Monte Carlo Tree Search}},
  booktitle = {Advances in Neural Information Processing Systems},
  year = {2024},
  url = {https://arxiv.org/abs/2405.15383}
}

@article{tsividis2021theory,
  author = {Pedro A. Tsividis and Joao Loula and Jake Burga and others},
  title = {{Human-Level Reinforcement Learning through Theory-Based Modeling, Exploration, and Planning}},
  journal = {arXiv preprint arXiv:2107.12544},
  year = {2021},
  url = {https://arxiv.org/abs/2107.12544}
}

@article{summerville2018pcgml,
  author = {Adam Summerville and Sam Snodgrass and Matthew Guzdial and others},
  title = {{Procedural Content Generation via Machine Learning (PCGML)}},
  journal = {IEEE Transactions on Games},
  year = {2018},
  url = {https://arxiv.org/abs/1702.00539}
}

@inproceedings{xia2024video2game,
  author = {Hongchi Xia and Zhi-Hao Lin and Wei-Chiu Ma and others},
  title = {{Video2Game: Real-time, Interactive, Realistic and Browser-Compatible Environment from a Single Video}},
  booktitle = {CVPR},
  year = {2024},
  url = {https://arxiv.org/abs/2404.09833}
}

@article{chollet2019arc,
  author = {Fran{\c{c}}ois Chollet},
  title = {{On the Measure of Intelligence}},
  journal = {arXiv preprint arXiv:1911.01547},
  year = {2019},
  url = {https://arxiv.org/abs/1911.01547}
}

@article{lake2017building,
  author = {Brenden M. Lake and Tomer D. Ullman and Joshua B. Tenenbaum and others},
  title = {{Building Machines That Learn and Think Like People}},
  journal = {Behavioral and Brain Sciences},
  year = {2017},
  url = {https://arxiv.org/abs/1604.00289}
}

@inproceedings{zheng2023judging,
  author = {Lianmin Zheng and Wei-Lin Chiang and Ying Sheng and others},
  title = {{Judging LLM-as-a-Judge with MT-Bench and Chatbot Arena}},
  booktitle = {Advances in Neural Information Processing Systems Datasets and Benchmarks Track},
  year = {2023},
  url = {https://arxiv.org/abs/2306.05685}
}

@inproceedings{chen2024mllmjudge,
  author = {Dongping Chen and Ruoxi Chen and Shilin Zhang and others},
  title = {{MLLM-as-a-Judge: Assessing Multimodal LLM-as-a-Judge with Vision-Language Benchmark}},
  booktitle = {International Conference on Machine Learning},
  year = {2024},
  url = {https://arxiv.org/abs/2402.04788}
}

@inproceedings{fu2025videomme,
  author = {Chaoyou Fu and Yuhan Dai and Yongdong Luo and others},
  title = {{Video-MME: The First-Ever Comprehensive Evaluation Benchmark of Multi-modal LLMs in Video Analysis}},
  booktitle = {Proceedings of the IEEE/CVF Conference on Computer Vision and Pattern Recognition (CVPR)},
  pages = {24108--24118},
  year = {2025},
  url = {https://arxiv.org/abs/2405.21075}
}

@article{yang2025qwen3,
  author = {An Yang and Anfeng Li and Baosong Yang and others},
  title = {{Qwen3 Technical Report}},
  journal = {arXiv preprint arXiv:2505.09388},
  year = {2025},
  url = {https://arxiv.org/abs/2505.09388}
}

@article{mnih2015dqn,
  author = {Volodymyr Mnih and Koray Kavukcuoglu and David Silver and others},
  title = {{Human-Level Control through Deep Reinforcement Learning}},
  journal = {Nature},
  volume = {518},
  pages = {529--533},
  year = {2015},
  doi = {10.1038/nature14236}
}

@article{vinyals2019alphastar,
  author = {Oriol Vinyals and Igor Babuschkin and Wojciech M. Czarnecki and others},
  title = {{Grandmaster Level in StarCraft II using Multi-Agent Reinforcement Learning}},
  journal = {Nature},
  volume = {575},
  pages = {350--354},
  year = {2019},
  doi = {10.1038/s41586-019-1724-z}
}

@book{efron1994bootstrap,
  author = {Bradley Efron and Robert J. Tibshirani},
  title = {An Introduction to the Bootstrap},
  publisher = {Chapman and Hall/CRC},
  year = {1994}
}

@misc{godotai,
  title = {{Godot AI}: An {MCP} Server for the {Godot} Editor},
  author = {{hi-godot}},
  year = {2026},
  howpublished = {\url{https://github.com/hi-godot/godot-ai}}
}

@misc{godot,
  author = {{Godot Engine contributors}},
  title = {{Godot Engine}},
  howpublished = {\url{https://godotengine.org}},
  year = {2026}
}

@misc{openai2026gpt6astra,
  title = {{GPT-6 Astra} System Card},
  author = {{OpenAI}},
  year = {2026},
  howpublished = {\url{https://deploymentsafety.openai.com/gpt-6-astra/gpt-6-astra.pdf}}
}

@misc{openai2026gpt61sol,
  title = {Addendum to {GPT-6 Astra} System Card: {GPT-6.1 Sol}},
  author = {{OpenAI}},
  year = {2026},
  howpublished = {\url{https://deploymentsafety.openai.com/gpt-6-1-sol/gpt-6-1-sol.pdf}}
}

@misc{anthropic2026opus55,
  title = {System Card: {Claude Opus 5.5}},
  author = {{Anthropic}},
  year = {2026},
  howpublished = {\url{https://www.anthropic.com/claude-opus-5-5-system-card}}
}

@misc{google2026gemini4,
  title = {{Gemini 4 Argon} Model Card},
  author = {{Google DeepMind}},
  year = {2026},
  howpublished = {\url{https://storage.googleapis.com/deepmind-media/Model-Cards/Gemini-4-Argon-Model-Card.pdf}}
}

@misc{deepseek2026v41flash,
  title = {{DeepSeek-V4.1-Flash}: Pushing the Limits of {KV} Cache Compression},
  author = {{DeepSeek-AI}},
  year = {2026},
  howpublished = {\url{https://huggingface.co/deepseek-ai/DeepSeek-V4.1-Flash/blob/main/DeepSeek_V41_Tech_Report.pdf}}
}

@article{glm2026glm5,
  title = {{GLM-5}: from Vibe Coding to Agentic Engineering},
  author = {{GLM-5-Team} and Zeng, Aohan and Lv, Xin and Hou, Zhenyu and Du, Zhengxiao and others},
  journal = {arXiv preprint arXiv:2602.15763},
  year = {2026}
}

@misc{openai2026codex,
  title = {{Codex}: Lightweight Coding Agent that Runs in Your Terminal},
  author = {{OpenAI}},
  year = {2026},
  note = {Version 0.167.0},
  howpublished = {\url{https://github.com/openai/codex}}
}

@misc{anthropic2025claudecode,
  title = {{Claude Code}},
  author = {{Anthropic}},
  year = {2025},
  howpublished = {\url{https://code.claude.com/docs/en/overview}}
}

@misc{google2025antigravity,
  title = {{Google Antigravity}},
  author = {{Google}},
  year = {2025},
  note = {Command-line interface, version 1.4.12},
  howpublished = {\url{https://antigravity.google}}
}

@misc{deepseek2026harness,
  title = {{DeepSeek Harness}: Everything is a Plugin},
  author = {{DeepSeek-AI}},
  year = {2026},
  howpublished = {\url{https://github.com/deepseek-ai/deepseek-harness}}
}

@misc{zai2026zcode,
  title = {{ZCode}},
  author = {{Z.ai}},
  year = {2026},
  howpublished = {\url{https://zcode.z.ai}}
}

@misc{qwen2026qwencode,
  title = {{Qwen Code}: An Open-Source {AI} Coding Agent that Lives in Your Terminal},
  author = {{Qwen Team}},
  year = {2026},
  note = {Version 0.26.1},
  howpublished = {\url{https://github.com/QwenLM/qwen-code}}
}
